\documentclass{article}

    \PassOptionsToPackage{numbers, compress}{natbib}
 \usepackage[preprint]{neurips_2026}

\usepackage[utf8]{inputenc} 
\usepackage[T1]{fontenc}    
\usepackage{hyperref}       
\usepackage{url}            
\usepackage{booktabs}       
\usepackage{amsfonts}       
\usepackage{nicefrac}       
\usepackage{microtype}      
\usepackage{xcolor}         

\usepackage{amsmath}
\usepackage{amsthm}
\usepackage{amssymb}            
\usepackage{mathtools}          
\usepackage{mathrsfs}           
\theoremstyle{plain}
\newtheorem{theorem}{Theorem}[section]

\theoremstyle{definition}
\newtheorem{definition}[theorem]{Definition}

\theoremstyle{remark}

\usepackage[inline]{enumitem}
\setlist[enumerate]{label={\arabic*)}}
\usepackage{wrapfig}
\definecolor{gray}{cmyk}{0,0,0,0.8}
\usepackage{algorithm}
\usepackage{algorithmic}

\usepackage{graphicx}           
\usepackage{subcaption}         
\usepackage[space]{grffile}     
\usepackage{multirow} 
\usepackage[capitalize,noabbrev]{cleveref}
\usepackage{comment}
\usepackage{dsfont}

\title{Scaling Curriculum Learning For Autonomous Driving}

\author{%
  Cevahir Koprulu\thanks{Corresponding author: \texttt{cevahir.koprulu@utexas.edu}}~~\textsuperscript{1},~~~David Paz\textsuperscript{2},~~~Feng Tao\textsuperscript{2},~~~Yuliang Guo\textsuperscript{2},~~~Xinyu Huang\textsuperscript{2}\\
  \textbf{Ufuk Topcu\textsuperscript{1},~~~Liu Ren\textsuperscript{2}} \\
  \textsuperscript{1}The University of Texas at Austin,~\textsuperscript{2}Bosch Center for AI, North America\\
}

\begin{document}

\newcommand{\Reals}{\mathbb{R}}
\newcommand{\NonNegativeReals}{\mathbb{R}_{\geq0}}
\newcommand{\PositiveIntegers}{\mathbb{Z}^+}
\newcommand{\Probability}{\mathbb{P}}
\newcommand{\Expectation}{\mathbb{E}}
\newcommand{\Indicator}{\mathbf{1}}
\newcommand{\ProbSimplex}{\Delta}
\newcommand{\Mod}[1]{\ (\mathrm{mod}\ #1)}

\newcommand{\POSG}{\mathcal{G}}
\newcommand{\POSGAgentSet}{\mathcal{N}}
\newcommand{\POSGNumberOfAgents}{\text{N}}
\newcommand{\POSGStateSpace}{\mathcal{S}}
\newcommand{\POSGActionSpace}{\mathcal{A}}
\newcommand{\POSGObservationSpace}{\mathcal{O}}
\newcommand{\POSGState}{\mathbf{s}}
\newcommand{\POSGAction}{\mathbf{a}}
\newcommand{\POSGObservation}{\mathbf{o}}
\newcommand{\POSGReward}{r}
\newcommand{\POSGTransitionFunction}{T}
\newcommand{\POSGObservationFunction}{Z}
\newcommand{\POSGRewardFunction}{R}
\newcommand{\POSGInitialStateDistribution}{I}
\newcommand{\POSGDiscount}{\gamma}
\newcommand{\POSGHorizon}{\text{H}}

\newcommand{\UPOSGScenarioParameterSet}{\Theta}
\newcommand{\UPOSGScenarioParameter}{\theta}
\newcommand{\UPOSG}{\POSG^{\UPOSGScenarioParameterSet}}
\newcommand{\UPOSGAgentSet}{\POSGAgentSet^{\UPOSGScenarioParameterSet}}
\newcommand{\UPOSGNumberOfAgents}{\text{N}}
\newcommand{\UPOSGStateSpace}{\POSGStateSpace}
\newcommand{\UPOSGActionSpace}{\POSGActionSpace^{\UPOSGScenarioParameterSet}}
\newcommand{\UPOSGObservationSpace}{\POSGObservationSpace^{\UPOSGScenarioParameterSet}}
\newcommand{\UPOSGState}{\POSGState}
\newcommand{\UPOSGAction}{\POSGAction}
\newcommand{\UPOSGObservation}{\POSGObservation}
\newcommand{\UPOSGPosition}{\mathbf{x}}
\newcommand{\UPOSGReward}{\POSGReward}
\newcommand{\UPOSGTransitionFunction}{\POSGTransitionFunction^{\UPOSGScenarioParameterSet}}
\newcommand{\UPOSGObservationFunction}{\POSGObservationFunction^{\UPOSGScenarioParameterSet}}
\newcommand{\UPOSGRewardFunction}{\POSGRewardFunction^{\UPOSGScenarioParameterSet}}
\newcommand{\UPOSGInitialStateDistribution}{\POSGInitialStateDistribution^{\UPOSGScenarioParameterSet}}
\newcommand{\UPOSGDiscount}{\POSGDiscount}
\newcommand{\UPOSGHorizon}{\POSGHorizon}
\newcommand{\UPOSGNumberOfScenarios}{\text{M}}
\newcommand{\UPOSGGoalStates}{\UPOSGStateSpace^{\UPOSGScenarioParameter}_{\text{Goal}}}
\newcommand{\UPOSGGoalStatesAgent}{\UPOSGStateSpace^{\UPOSGScenarioParameter,n}_{\text{Goal}}}

\newcommand{\UEDLevelGenerator}{\Lambda}
\newcommand{\UEDPolicySpace}{\Pi}
\newcommand{\UEDDistributionOverLevelS}{\ProbSimplex(\UPOSGScenarioParameterSet)}
\newcommand{\UEDUtilityFunction}{U}
\newcommand{\UEDConstantUtility}{\text{C}}
\newcommand{\UEDRegretUtilityFunction}{\UEDUtilityFunction^{\text{Regret}}}
\newcommand{\UEDLearnabilityUtilityFunction}{\UEDUtilityFunction^{\text{Learn}}}
\newcommand{\UEDRealismUtilityFunction}{\UEDUtilityFunction^{\text{Real}}}

\newcommand{\AMGAE}{\UEDUtilityFunction^{\text{AMGAE}}}
\newcommand{\PVL}{\UEDUtilityFunction^{\text{PVL}}}
\newcommand{\MaxMC}{\UEDUtilityFunction^{\text{MaxMC}}}
\newcommand{\LearnabilityHard}{\UEDUtilityFunction^{\text{Learn-hard}}}
\newcommand{\Learnability}{\UEDUtilityFunction^{\text{Learn}}}
\newcommand{\GCADE}{\UEDUtilityFunction^{\text{GC-ADE}}}
\newcommand{\ACTMAE}{\UEDUtilityFunction^{\text{Act-MAE}}}

\newcommand{\PLRBuffer}{\mathcal{B}}
\newcommand{\PLRReplayDistribution}{\Probability_{\text{replay}}}
\newcommand{\PLRScoreDistribution}{\Probability_{\text{utility}}}
\newcommand{\PLRStalenessDistribution}{\Probability_{\text{staleness}}}
\newcommand{\PLRIterationNumber}{l}
\newcommand{\PLRStalenessCoefficient}{\rho}
\newcommand{\PLRScoreTemperature}{\beta}
\newcommand{\PLRTrainingLevels}{\UPOSGScenarioParameterSet^{\text{train}}}
\newcommand{\PLRReplayRate}{d}
\newcommand{\PLRMaxBufferSize}{\text{B}^{\text{max}}}
\newcommand{\Policy}{\pi}
\newcommand{\Value}[1][]{V^{#1}}
\newcommand{\GAEDiscount}{\lambda}
\newcommand{\TDError}{\delta}
\newcommand{\MaximumReturn}{\text{R}_{\text{max}}^{\UPOSGScenarioParameter}}
\newcommand{\MaximumReturnAgent}[1][n]{\text{R}_{\text{max}}^{\UPOSGScenarioParameter,#1}}
\newcommand{\SuccessRate}{p}
\newcommand{\PolicyParameter}{\phi}
\newcommand{\TotalIterations}{\text{T}^{\text{train}}}
\newcommand{\ScenarioSamplingInterval}{\text{T}^{\text{sce}}}
\newcommand{\PolicyUpdateInterval}{\text{T}^{\text{pol}}}
\newcommand{\SampleFromCurriculum}[1][]{{\textsc{SampleFromCurriculum}}\textbf{(}{#1}\textbf{)}}
\newcommand{\UpdateCurriculum}[1][]{{\textsc{UpdateCurriculum}}\textbf{(}{#1}\textbf{)}}
\newcommand{\NumberOfWorlds}{\text{W}}
\newcommand{\InteractionSet}{\mathcal{D}}
\newcommand{\UpdatePolicy}[1][]{{\Phi}\textbf{(}{#1}\textbf{)}}
\newcommand{\EndOfEpisodeFlag}{e}
\newcommand{\Rollout}{\tau}

\newcommand{\NumberOfEpisodes}{\text{K}}

\newcommand{\GIGAFLOW}{\textsc{GigaFlow}}
\newcommand{\GPUDRIVE}{\textsc{GPUDrive}}

\newcommand{\CLForAD}{\textsc{CL4AD}}

\maketitle

\begin{abstract}
  Batched simulators for autonomous driving have recently enabled training reinforcement learning (RL) agents at scale, encompassing thousands of traffic scenarios and billions of interactions within a matter of days. 
Although such high-throughput feeds RL algorithms faster than ever, their sample-efficiency has not kept pace: 
As the standard training scheme, domain randomization uniformly samples scenarios, thereby consuming a vast number of interactions on cases that contribute little to learning.
Curriculum learning offers a remedy by adaptively prioritizing scenarios that matter most to policy improvement. 
We present $\CLForAD$, the first integration of curriculum learning into batched autonomous driving simulators by framing scenario selection as an unsupervised environment design problem.
We introduce utility functions that shape curricula based on success rates and the realism of the agent's behavior, in addition to existing regret-estimation functions.
Large-scale experiments in $\GPUDRIVE$ demonstrate that curriculum learning achieves a $99\%$ success rate a billion steps earlier than domain randomization, reducing wall-clock time by $77\%$, and outperforms heuristic curricula with static and dynamic attributes, with only one exception at the largest scale.
An ablation under limited compute shows that curriculum learning improves sample efficiency by $67\%$.
We also investigate how utility functions behave at scale, and how prioritized scenarios evolve during training. 
We release \href{https://anonymous.4open.science/r/gpudrive-37D3/README.md}{an implementation of $\CLForAD$ in $\GPUDRIVE$}.
\end{abstract}

\section{Introduction}\
\label{sec:introduction}
Batched simulators for autonomous driving (AD) have recently empowered sample-inefficient but effective reinforcement learning (RL) algorithms by enabling training for billions of interactions within days \citep{cusumano-towner2025robust,kazemkhani2025gpudrive}.  
These simulators achieve such scale by training RL agents on hundreds to thousands of scenarios in parallel through self-play \citep{silver2017mastering}, where a single policy controls all vehicles, taking millions of actions per second.
Agents trained on $\GPUDRIVE$ \citep{kazemkhani2025gpudrive} using the Waymo Open Motion Dataset (WOMD) \citep{ettinger2021large} generalize to unseen test cases in less than a day.
$\GIGAFLOW$ \citep{cusumano-towner2025robust}, further scales self-play to 1.6 billion kilometers of simulated driving within 10 days, producing generalist driving policies that outperform benchmark-specific agents on CARLA \citep{dosovitskiy2016learning}, nuPlan \citep{caesar2021nuplan}, and Waymax \citep{gulino2023waymax} without any training on these benchmarks.

Despite advances in high simulation throughput, training RL agents in batched driving simulators remains sample-inefficient due to a standard training strategy: uniform scenario sampling, i.e., domain randomization (DR). 
This approach wastes interactions on scenarios that are either too easy to provide a sufficient learning signal or too difficult for the current policy to make progress on. 
Curriculum learning (CL) offers a remedy by adaptively prioritizing scenarios that contribute the most to policy improvement \citep{narvekar2020curriculum}. 
In particular, curriculum learning has successfully fulfilled that promise in multiple large-scale RL domains. 
For example, \citep{bauer2023human} demonstrate that scaling meta-RL with automated curricula yields agents capable of human-timescale adaptation across thousands of procedurally generated environments. 
\citep{zhang2024omni} introduce curricula for open-ended environments with infinitely many tasks, showing that curriculum learning enables faster and broader skill acquisition. 

\begin{figure*}[tbp]
\centering
    \centering
    \includegraphics[width=\textwidth]{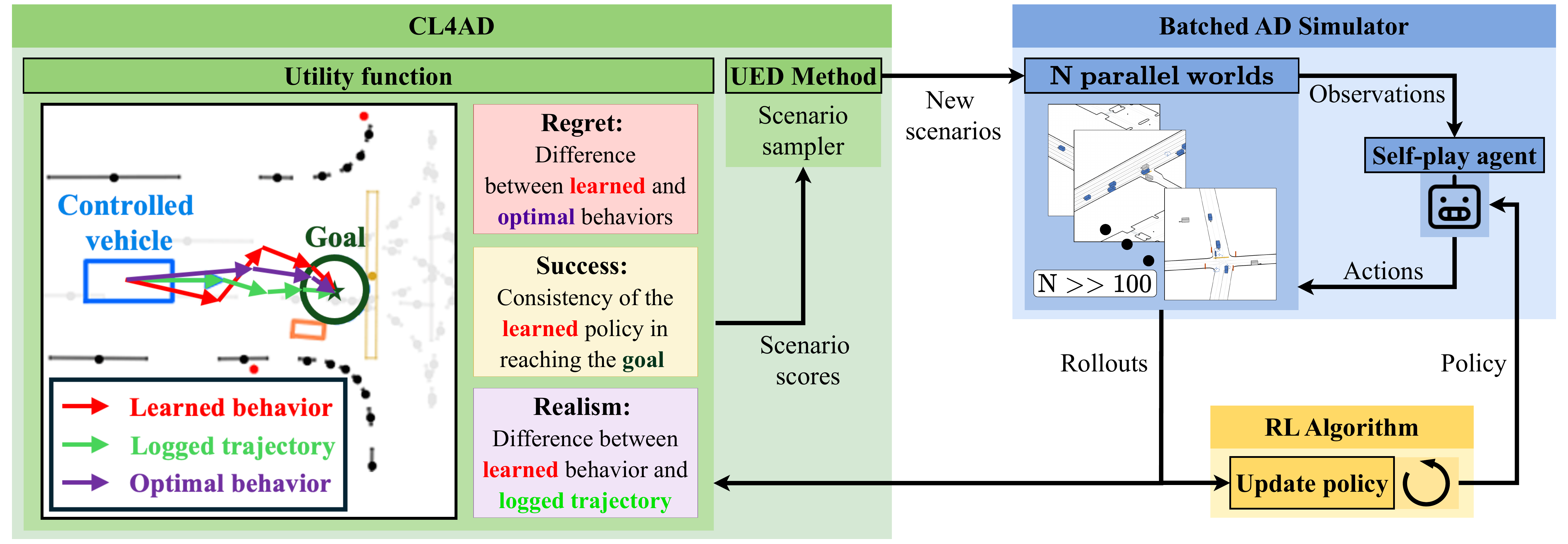}
    \vspace*{-5mm}
    \caption{$\CLForAD$ integrates a UED method into a batched AD simulator to adaptively prioritize traffic scenarios based on three types of utility functions: regret, success, and realism. \textbf{(blue)} A self-play agent interacts with sampled scenarios across concurrent worlds. \textbf{(yellow)} An RL algorithm of choice updates the policy using collected rollouts in said scenarios. \textbf{(green)} $\CLForAD$ scores scenarios based on a chosen utility function and updates its sampler via the integrated UED method.}
    \label{fig:cl4ad}
\end{figure*}

Inspired by the success of CL in large-scale RL, we introduce \emph{Curriculum Learning for Autonomous Driving} ($\CLForAD$) (see \cref{fig:cl4ad}), the first integration of automated curricula into a batched AD simulator. We frame scenario selection as an unsupervised environment design problem (UED), and equip a UED method with utility functions that adaptively shape training. Therefore, the curricula prioritize scenarios at the edge of the agent's capabilities. Across experiments, we show that CL accelerates RL training by hundreds of millions of steps compared to DR and heuristic curricula. 

Our key contributions are three-fold:
\vspace*{-1.5mm}
\begin{itemize}[nolistsep,noitemsep,leftmargin=*]
    \item We present \textbf{$\CLForAD$}, the first integration of curriculum learning methods from unsupervised environment design to the scale of GPU-accelerated, self-play simulators for AD, and provide an implementation in an open-source batched AD simulator, $\GPUDRIVE$.
    \item We propose \textbf{novel utility functions} based on realism, and a safety-critical success criterion that balances robustness and safety, exposing the gap between optimal and realistic driving.
    \item We conduct a \textbf{large-scale empirical study} \begin{enumerate*}
        \item showing that $\CLForAD$ accelerates RL training by up to a billion interactions compared to DR and outperforms heuristic curricula;
        \item investigating how prioritized scenarios evolve during training;
        \item analyzing the correlation among utility functions and performance metrics;         
        \item describing how multi-agent self-play creates a persistent learning frontier; 
        \item demonstrating that reward optimality and behavioral realism are distinct axes; and
        \item providing practitioner guidelines for curriculum learning in batched AD simulators.
    \end{enumerate*}
\end{itemize}
\vspace*{-1.5mm}

\section{Related Work}
\label{sec:related_worlds}
\textbf{Autonomous driving simulators} have enabled RL to train self-driving agents in multiple ways: Waymax \citep{gulino2023waymax} and Nocturne \citep{vinitsky2022nocturne} use traffic scenarios from open-source driving datasets such as WOMD \citep{ettinger2021large}, whereas CARLA \citep{dosovitskiy2017carla} is not data-driven, and Metadrive enables procedural scenario generation as well as integration of real driving data. Attempts to scale RL for AD have led to batched simulators such as Waymax, $\GPUDRIVE$ \citep{kazemkhani2025gpudrive}, and $\GIGAFLOW$ \citep{cusumano-towner2025robust}, which significantly increase data throughput for RL algorithms but still rely on random scenario generation/sampling.

\textbf{Curriculum learning for RL} accelerates learning optimal policies by sequencing different configurations of the environment \citep{narvekar2020curriculum}. Automated curriculum generation studies goal-conditioned domains \citep{baranes2010intrinsically,florensa2018automatic,tzannetos2023proximal}, contextual settings \citep{klink2022curriculum,koprulu2023risk,sayar2024diffusion},  and more popularly UED \citep{dennis2020emergent}. Across many settings, a curriculum requires a signal that measures the contribution of a task to policy improvement, e.g., learning progress, i.e., the change in an agent's competence, which \cite{portelas2020teacher} estimates over a continuous parameter space and \cite{kanitscheider2021multi} measures as the change in success probability. UED methods can generate levels through trained teachers or by sampling free parameters. PAIRED \citep{dennis2020emergent} trains a teacher that emits level parameters maximizing regret against an antagonist, and RE-PAIRED \citep{jiang2021replay} stabilizes this game by replaying high-regret levels. ACCEL \citep{parker2022evolving} instead mutates the parameters of high-regret levels in a replay buffer and curates the mutants that score highly. PLR \cite{jiang2021prioritized}, as one of the earlier UED approaches, neither trains a teacher nor mutates parameters; instead, it scores and replays levels it has already encountered. PLR has also shown evidence of scalability in meta RL \citep{bauer2023human,jackson2023discovering}, and open-ended environments \citep{zhang2024omni}. Beyond individual methods, libraries such as DCD \citep{jiang2021replay}, minimax \cite{jiang2023minimax}, and Syllabus \cite{sullivan2025syllabus} implement curricula for batched or asynchronous training, though they target procedurally generated environments outside AD.

\textbf{Curriculum learning for AD} aims to speed up training self-driving policies via RL, e.g., ScenarioNet \citep{li2023scenarionet}, which unifies heterogenous data for simulation, showcases benefits of heuristic-based curricula. Similarly, \citep{anzalone2021reinforced,anzalone2022end} propose a multi-stage curriculum learning method for CARLA, making the number of agents, their initial positions, or weather conditions incrementally more difficult. In contrast to manual curricula, \citep{qiao2018automatically} develops an automated method for urban intersections.  Recently, \citep{brunnbauer2024scenario} and \citep{abouelazm2025automatic} have demonstrated that UED methods, RE-PAIRED and ACCEL, respectively, accelerate training AD agents in CARLA. However, there has been no investigation into whether CL can scale with the high throughput and scenario diversity enabled by batched simulators such as $\GPUDRIVE$, which trains RL agents on real-world scenarios from the Waymo Open Motion Dataset (WOMD) \citep{ettinger2021large}. Other curriculum strategies for AD include adversarial scenario generation \cite{zhang2023cat}, continual policy adaptation with individualized curricula \cite{niu2024continual}, and VLM-guided safety-critical curriculum design \cite{sheng2026curricuvlm}. These works train single-ego agents on a smaller scale, in contrast to CL4AD, which enables training with real-world datasets across thousands of concurrent self-play agents in hundreds of parallel worlds.

\section{Background}
\label{sec:background}
We model a traffic scenario as a \emph{partially observable stochastic game} (POSG) \citep{brunnbauer2024scenario},  to accommodate the multi-agent nature of driving. Then, we frame curriculum learning for autonomous driving as \emph{unsupervised environment design}, and describe how to measure the utility of traffic scenarios.

\subsection{Traffic scenarios as partially observable stochastic games}
\label{sec:posg}
\begin{definition}
A \emph{POSG} is a tuple $\POSG=\langle \POSGAgentSet,\POSGStateSpace,\POSGActionSpace,\POSGObservationSpace,\POSGTransitionFunction,\POSGObservationFunction,\POSGRewardFunction, \POSGInitialStateDistribution,\POSGDiscount \rangle$, where $\POSGAgentSet=[\POSGNumberOfAgents]$ is the set of agents with $\POSGNumberOfAgents\in\PositiveIntegers$, $\POSGStateSpace$ is the state space, $\POSGActionSpace=\prod_{i=1}^{\POSGNumberOfAgents}\POSGActionSpace_i$ and $\POSGObservationSpace=\prod_{i=1}^{\POSGNumberOfAgents}\POSGObservationSpace_i$ are the joint action and observation spaces. $\POSGTransitionFunction: \POSGStateSpace \times \POSGActionSpace \to \ProbSimplex(\POSGStateSpace)$ is the stochastic dynamics, i.e, the probability of transitioning from state $\POSGState\in\POSGStateSpace$ to state $\POSGState'\in\POSGStateSpace$ given joint action $\POSGAction\in\POSGActionSpace$. 
$\POSGObservationFunction: \POSGStateSpace \times \POSGActionSpace \to \ProbSimplex(\POSGObservationFunction)$ determines the probability of observing $\POSGObservation=(\POSGObservation_1,\POSGObservation_2,\cdots,\POSGObservation_N)\in\POSGObservationSpace$ in state $\POSGState$ taking joint action $\POSGAction$.
$\POSGRewardFunction: \POSGStateSpace \times \POSGActionSpace \to \Reals^N$ is the reward function, i.e., $\POSGRewardFunction(\POSGState,\POSGAction)=(\POSGRewardFunction_i(\POSGState,\POSGAction))_{i \in [\POSGNumberOfAgents]}$ where $\POSGRewardFunction_i(\POSGState,\POSGAction)\in\Reals$ is the reward for agent $i\in\POSGAgentSet$. $\POSGInitialStateDistribution\in\ProbSimplex(\POSGStateSpace)$ is the initial state distribution, and $\POSGDiscount\in[0,1]$ is the discount factor.
\end{definition}
\vspace{-.15cm}
A policy $\Policy_i:\POSGObservationSpace_i\to\ProbSimplex(\POSGActionSpace_i)$ describes the behavior of agent $i$ in POSG $\POSG$. The value function for $\Policy_i$ is the expected cumulative discounted rewards over a horizon of $\POSGHorizon$ steps, i.e., $
    \Value(\Policy_i) = \Expectation_{\POSGTransitionFunction,\POSGObservationFunction}\left[ \sum_{t=0}^{\POSGHorizon-1} \POSGDiscount^t\POSGRewardFunction_i(\POSGState_t,\POSGAction_t) | \POSGState_o\sim\POSGInitialStateDistribution, \POSGAction_t=(\POSGAction_{j,t})_{j\in\POSGAgentSet}\right]
    $ 
    where $\POSGAction_{j,t}\sim\Policy_j(\POSGObservation_{j,t})
$. Agent $i$ aims to find an optimal policy $\Policy_i^*$, which maximizes its value $\Value(\Policy_i)$ in POSG $\POSG$. 

In a traffic scenario modeled as $\POSG$, consider $\Policy_i$ as a policy that controls vehicle $i$. The road layout, traffic rules, and collision dynamics in a scenario specify the dynamics $\POSGTransitionFunction$. Initial state $\POSGState_0\sim\POSGInitialStateDistribution$ consists of the initial positions of all vehicles, pedestrians, cyclists, etc. Observation $\POSGObservation_{i,t}$ of vehicle $i$ at time $t\in[\POSGHorizon]$ is what the controller perceives about the surroundings based on its sensors as well as specific attributes, e.g., the type of vehicle, its velocity, acceleration, etc. The reward $\POSGReward_i=\POSGRewardFunction_i(\POSGState_t,\POSGAction_t)$ can incentivize the policy to reach a goal location, stay within lanes, and avoid collisions. To model multiple traffic scenarios, we formalize them as an \emph{underspecified} POSG (UPOSG).

\begin{definition}
    An \emph{underspecified POSG} $\UPOSG=\langle \UPOSGScenarioParameterSet, \UPOSGAgentSet, \UPOSGStateSpace, \UPOSGActionSpace, \UPOSGObservationSpace, \UPOSGTransitionFunction, \UPOSGObservationFunction, \UPOSGRewardFunction, \UPOSGInitialStateDistribution, \UPOSGDiscount \rangle$ models a set of POSGs through parameters $\UPOSGScenarioParameter\in\UPOSGScenarioParameterSet$ that determine all attributes of a POSG $\UPOSGScenarioParameter\in\UPOSGScenarioParameterSet$ depending on its agents, such as the dynamics $\UPOSGTransitionFunction:\UPOSGStateSpace\times\UPOSGActionSpace\times\UPOSGScenarioParameterSet\to\ProbSimplex(\UPOSGStateSpace)$.
\end{definition}
\vspace{-.15cm}
Consider scenarios $\UPOSGScenarioParameterSet=\{\UPOSGScenarioParameter_m\}_{m\in[\UPOSGNumberOfScenarios]}$ in WOMD, where $\UPOSGNumberOfScenarios\approx100,000$. A scenario $\UPOSGScenarioParameter_m$ may correspond to an urban intersection or a highway, with varying speed limits, number of vehicles, etc. In practice, $\UPOSGScenarioParameter_m$ is merely an identification number, i.e., $\UPOSGScenarioParameter_m\in[\UPOSGNumberOfScenarios]$, hence it is underspecified.

\subsection{Unsupervised Environment Design}
\label{sec:ued}
UED \citep{dennis2020emergent} aims to generate a sequence of \emph{levels}\footnote{As \emph{level} is the common term in the UED literature to describe $\UPOSGScenarioParameter$, we use it interchangeably with \emph{scenario}.}, i.e., scenarios $\UPOSGScenarioParameter\in\UPOSGScenarioParameterSet$ in the case of AD, to accelerate learning a policy that generalizes across all levels. One solution to UED is a level generator $\UEDLevelGenerator: \UEDPolicySpace \to \UEDDistributionOverLevelS$ that produces a distribution over the set of levels $\UPOSGScenarioParameterSet$ given a policy $\Policy\in\UEDPolicySpace$. A level generator $\UEDLevelGenerator$ maximizes a utility function $\UEDUtilityFunction(\Policy,\UPOSGScenarioParameter)$ measuring the contribution of $\UPOSGScenarioParameter$ for improving $\Policy$.

Domain randomization, i.e., uniformly sampling levels throughout the training, is the default way of training an RL agent where the utility is constant for each level, namely, $\UEDUtilityFunction(\Policy,\UPOSGScenarioParameter)=\UEDConstantUtility\in\Reals, \forall\UPOSGScenarioParameter\in\UPOSGScenarioParameterSet$. UED methods primarily differ in their utility functions of choice. There are two common categories of utility functions: regret and success-based. 

\begin{wrapfigure}[23]{r}{0.6\textwidth}
\centering
    \vspace{-1.5\baselineskip}
    \includegraphics[width=\linewidth,trim={0pt 0pt 0pt 0pt},clip]{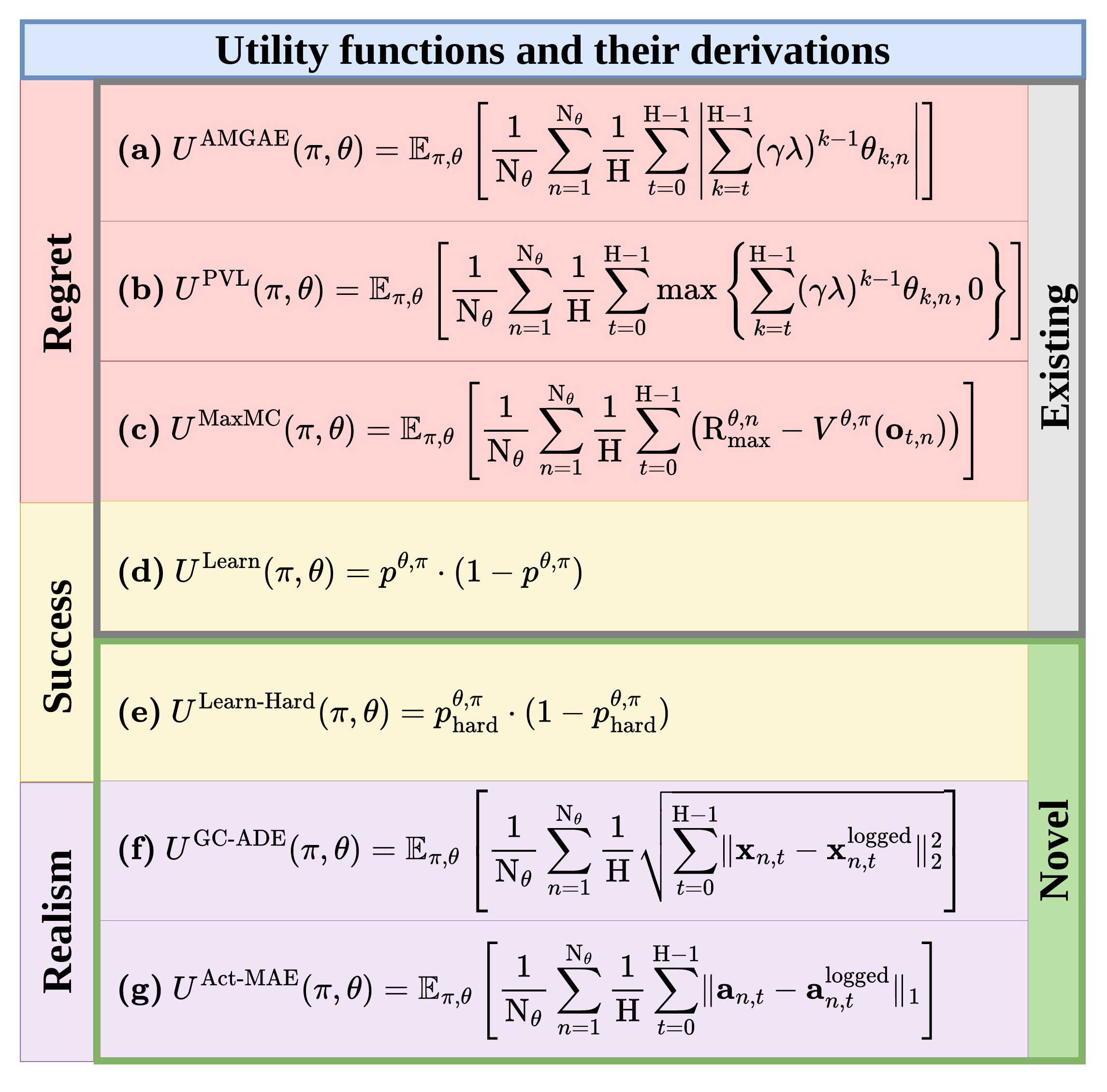}
    \vspace*{-.7cm}
    \caption{Utility functions: $\CLForAD$ adapts existing \textbf{(a-d)} utility functions as well as proposes \textbf{(e-g)} novel ones.}
\label{fig:cl4ad_utility}
\end{wrapfigure}

Regret, i.e., the difference between the expected discounted return of the current policy and the optimal one, enables prioritizing the easiest levels that the agent cannot currently solve \citep{dennis2020emergent}. More formally, a regret-based utility is
$
\UEDRegretUtilityFunction(\Policy,\UPOSGScenarioParameter)=\Value[\UPOSGScenarioParameter](\Policy^{*}_{\UPOSGScenarioParameter})-\Value[\UPOSGScenarioParameter](\Policy),
$
where $\Policy^{*}_{\UPOSGScenarioParameter}$ is an optimal policy in level $\UPOSGScenarioParameter$, i.e., a policy collecting the maximum expected discounted return $\Value[\UPOSGScenarioParameter](\Policy^{*}_{\UPOSGScenarioParameter})$. However, as the optimal expected discounted return or the optimal policy for each level is rarely available, UED methods estimate regret in various ways. \citep{jiang2021prioritized} propose learning potential, i.e., \emph{average magnitude of the generalized advantage estimate} (AMGAE) \citep{schulman2015high}, (a) in \cref{fig:cl4ad_utility}, as a utility function that estimates regret over a single episode.
Here, $\TDError_{k.n}=\POSGReward_{k,n}+\UPOSGDiscount \Value[\UPOSGScenarioParameter,\Policy](\UPOSGObservation_{k+1,n})-\Value[\UPOSGScenarioParameter,\Policy](\UPOSGObservation_{k,n})$ is the temporal difference error at timestep $k$ of agent $n\in[\UPOSGNumberOfAgents_{\UPOSGScenarioParameter}]$. $
\Value[\UPOSGScenarioParameter,\Policy](\POSGObservation_{k,n})=\Expectation_{\UPOSGTransitionFunction,\UPOSGObservationFunction}\left[ \sum_{t=k}^{\POSGHorizon-k-1} \POSGDiscount^{t-k}\UPOSGRewardFunction(\POSGState_{t,n},\POSGAction_{t,n},\UPOSGScenarioParameter) | \POSGAction_{t,n}\sim\Policy(\POSGObservation_{t,n})\right]
$ is the expected discounted return of agent $n$ in $\UPOSGState_{k}$ on level $\UPOSGScenarioParameter$, and $\GAEDiscount$ is the discount factor for GAE. Alternatively, \citep{jiang2021replay} and \citep{parker2022evolving} employ \emph{positive value loss} (PVL), (b) in \cref{fig:cl4ad_utility}.
As PVL uses the bootstrapped value target to compute the temporal difference error, \citep{jiang2021replay} also propose \emph{maximum Monte Carlo} (MaxMC), (c) in \cref{fig:cl4ad_utility}, which instead utilizes the highest return obtained by agent $n$ on level $\UPOSGScenarioParameter$ to mitigate potential bias issues,
where $\MaximumReturnAgent $ is the maximum discounted return achieved in level $\UPOSGScenarioParameter$ so far during training.

Success-based utility functions address settings where a level $\UPOSGScenarioParameter$ is considered solved for agent $n$ when a policy $\Policy$ reaches a goal state $\UPOSGState\in\UPOSGGoalStatesAgent\subset\UPOSGStateSpace$. Such utility functions use the success rate $\SuccessRate^{\UPOSGScenarioParameter,\Policy}$, i.e., the fraction of controlled agents that reach a goal state in an episode of level $\UPOSGScenarioParameter$ under policy $\Policy$,
$
\SuccessRate^{\UPOSGScenarioParameter,\Policy}=\frac{1}{\UPOSGNumberOfAgents}\sum_{n\in\UPOSGNumberOfAgents}\mathds{1}[\exists t\in[\UPOSGHorizon]:\UPOSGState_t\in\UPOSGGoalStatesAgent].
$
Inspired by \citep{tzannetos2023proximal}, \citep{rutherford2024no} propose \emph{Sampling for Learnability} (SFL), along with \emph{learnability}, (d) in \cref{fig:cl4ad_utility},
a utility function corresponding to the variance of a Bernoulli distribution with parameter $\SuccessRate^{\UPOSGScenarioParameter,\Policy}$, namely, how inconsistently policy $\Policy$ solves $\UPOSGScenarioParameter$ across all controlled agents. \citep{rutherford2024no} argues that, in sparse-reward settings, regret-based utility functions exhibit low correlation with success rates, as regret-based utility functions become noisy in such settings, causing inaccurate identification of the learning frontier. In AD, rewards commonly occur after sparse events, such as goal completion or collisions; thus, learnability becomes a viable alternative.

\section{Curriculum Learning for Autonomous Driving at Scale}
\label{sec:cl4ad}
\begin{algorithm*}[t]
\caption{\textbf{C}urriculum \textbf{L}earning for \textbf{A}utonomous \textbf{D}riving ($\CLForAD$)}
\label{alg:cl4ad}
\textbf{Input}: Set of training scenarios $\PLRTrainingLevels$ \\
\textbf{Parameters}: Replay rate $\PLRReplayRate$, Staleness coefficient $\PLRStalenessCoefficient$, temperature $\PLRScoreTemperature$, utility function $\UEDUtilityFunction$, max buffer size $\PLRMaxBufferSize$, total number of iterations $\TotalIterations$, scenario sampling interval $\ScenarioSamplingInterval$, policy update interval $\PolicyUpdateInterval$, number of worlds $\NumberOfWorlds$ \\
\textbf{Output}: Final policy $\Policy_{\PolicyParameter}$
\begin{algorithmic}[1] 
    \STATE $\PLRBuffer\leftarrow()$, $\InteractionSet\leftarrow()$ $t\leftarrow0$,$l\leftarrow0$, $\Policy_{\PolicyParameter}\leftarrow\Policy_{\PolicyParameter_0}$
    \COMMENT{Reset scenario/experience buffers, iterators, and policy}
    \WHILE{$t<\TotalIterations$}
    \IF{$0\equiv t\mod{\ScenarioSamplingInterval}$}
    \STATE $\PLRIterationNumber\leftarrow\PLRIterationNumber+1$
    \COMMENT{Increment sampling iteration}
    \STATE $(\UPOSGScenarioParameter_{w})_{w=1}^{\NumberOfWorlds},\PLRBuffer\leftarrow\SampleFromCurriculum[\PLRBuffer,\PLRTrainingLevels,\PLRIterationNumber]$
    \COMMENT{Sample scenarios for worlds}
    \ENDIF
    \STATE $\InteractionSet_t=\{\{\UPOSGObservation_{n,w},\UPOSGAction_{n,w},\UPOSGObservation'_{n,w},\UPOSGReward_{n,w},\EndOfEpisodeFlag_{n,w}\}_{n\in{[\UPOSGNumberOfAgents_{\UPOSGScenarioParameter_w}]}}\}_{w\in[\NumberOfWorlds]}$
    \COMMENT{Record experiences over a single step}
    \STATE $\PLRBuffer\leftarrow\UpdateCurriculum[\InteractionSet_t,\UEDUtilityFunction,\PLRBuffer]$
    \COMMENT{Update curriculum with the scores of terminated scenarios}
    \STATE $\InteractionSet\leftarrow\InteractionSet\cup\InteractionSet_t$
    \COMMENT{Update experience buffer with new interactions}
    \IF{$0\equiv t\mod{\PolicyUpdateInterval}$}
    \STATE $\Policy,\InteractionSet\leftarrow\UpdatePolicy[\InteractionSet]$
    \COMMENT{Update self-play policy via RL algorithm $\Phi$, and reset the experience buffer $\InteractionSet$}
    \ENDIF
    \STATE $t\leftarrow t + |\InteractionSet_t|$
    \COMMENT{Update training iteration}
    \ENDWHILE
\end{algorithmic}
\end{algorithm*}
\textbf{$\CLForAD$ scales UED to batched AD simulators.} 
$\CLForAD$ integrates variants of a UED method, \emph{prioritized level replay} (PLR) \citep{jiang2021prioritized}, which lays the foundation for approaches such as Robust PLR \citep{jiang2021replay}, REPAIRED, ACCEL, and SFL. $\CLForAD$ scales them up for a batched AD simulator on four axes:
\begin{enumerate*}[label=\textbf{(\arabic*)}] 
    \item concurrent simulation of hundreds of scenarios,
    \item tracking tens of agents per scenario,
    \item training in tens of thousands of scenarios, and
    \item for billions of steps.
\end{enumerate*}

\textbf{$\CLForAD$ efficiently scales via asynchronous episode management.} $\CLForAD$ tracks the behavior of all controlled agents in all concurrent scenarios to compute their utility, which captures the expected collective behavior of the self-play policy. In $\GPUDRIVE$, where we implement $\CLForAD$, simulated scenarios come from real-world datasets, and each scenario has a specific horizon $\UPOSGHorizon$ due to the nature of the logged data. Since scenarios have different horizons and agents terminate at different times, $\CLForAD$ processes episode terminations asynchronously: it computes the utility of a scenario as soon as all its agents terminate, updates the score by averaging over the episodes of that scenario since the last sampling call, and frees the rollout buffer for the next episode. This keeps memory usage proportional to the number of active episodes rather than the total number of episodes between curriculum sampling steps. Wall-clock measurements show that curriculum updates account for approximately $1\%$ of total training time (see \cref{app:cl4ad}).

\textbf{CL4AD operates over large training sets by adapting PLR's sampling mechanism,} which has two parts: uniformly sampling scenarios from the training set $\PLRTrainingLevels$, and replaying levels from a rolling buffer $\PLRBuffer$. At the beginning of the training, PLR uniformly randomly samples scenarios, and scores them based on a utility function $\UEDUtilityFunction$. Then, it adds scenarios with the highest scores to its buffer. Subsequently, PLR makes a random decision with probability $\PLRReplayRate$ to sample unseen levels in $\PLRTrainingLevels$ or seen levels from the buffer via a distribution based on their scores and staleness, namely,
\begin{equation}
\PLRReplayDistribution(\UPOSGScenarioParameter_i|\PLRBuffer,\UEDUtilityFunction,\PLRIterationNumber)=(1-\PLRStalenessCoefficient)\cdot\PLRScoreDistribution(\UPOSGScenarioParameter_i|\PLRBuffer,\UEDUtilityFunction)+\PLRStalenessCoefficient\cdot\PLRStalenessDistribution(\UPOSGScenarioParameter_i|\PLRBuffer,\PLRIterationNumber),
    \label{eq:plr_replay}
\end{equation}
where $\PLRScoreDistribution(\UPOSGScenarioParameter_i|\PLRBuffer,\UEDUtilityFunction)$ emphasizes levels with higher ranks with respect to their scores, and $\PLRStalenessDistribution(\UPOSGScenarioParameter_i|\PLRBuffer,\PLRIterationNumber)$ assigns a higher likelihood for levels that have not been sampled for longer (see \cref{app:cl4ad}). This distribution aims to prevent the scores of seen levels from becoming off-policy, as they may remain in the buffer for a while without being sampled.

\textbf{$\CLForAD$ introduces three novel utility functions,} (e-g) in \cref{fig:cl4ad_utility}: \emph{learnability-hard} $\LearnabilityHard$, \emph{goal-conditioned average distance error} (GC-ADE) $\GCADE$, and \emph{action mean absolute error} (Act-MAE) $\ACTMAE$.
$\LearnabilityHard$ is a success-based utility function that, in contrast to $\Learnability$, utilizes the rate at which agents reach their goals without colliding or going off-road in scenario $\UPOSGScenarioParameter$ via self-play policy $\Policy$. Since episodes do not terminate on collision or off-road events, an agent can complete its goal unsafely, which $\Learnability$ counts as a success and $\LearnabilityHard$ does not. $\LearnabilityHard$ therefore applies learnability to a success criterion that AD works use, as it captures both robustness and safety \citep{cusumano-towner2025robust}. $\GCADE$ and $\ACTMAE$ are realism-based utility functions that compute the distance between the positions and actions of RL agents and the logged trajectories, respectively, evaluating the plausibility of behavior \citep{caesar2021nuplan,gulino2023waymax,cornelisse2024humancompatible}.

\textbf{\cref{alg:cl4ad} illustrates $\CLForAD$.} At the beginning of the training, we initialize self-play policy $\Policy_{\PolicyParameter}$, and reset scenario and experience buffers $\PLRBuffer$ and $\InteractionSet$, as well as the training and scenario sampling iterations, $t$ and $\PLRIterationNumber$, respectively (Line 1). Until training iteration reaches $\TotalIterations$, $\CLForAD$ first checks if it is time to sample new scenarios based on its replay buffer $\PLRBuffer$ (Line 3-5). If so, PLR samples new scenarios, and $\CLForAD$ sets them to concurrently simulated worlds. Note that PLR only keeps $\PLRMaxBufferSize$ highest ranking scenarios in the buffer for sampling. Then, the self-play policy $\Policy_{\PolicyParameter}$ takes a step in all scenarios, and $\InteractionSet_t$ records them (Line 7). $\CLForAD$ updates the curriculum buffer using the utility of terminated scenarios (Line 8). Finally, an RL algorithm updates the policy using the experience buffer $\InteractionSet$ (Line 9-12) every $\PolicyUpdateInterval$ steps. We refer the reader to \cref{app:cl4ad} for more details.

\section{Experimental Results}
\label{sec:results}
\begin{figure*}[t]
    \centering
    \begin{subfigure}[b]{\textwidth}
        \centering
        \includegraphics[width=\textwidth,trim={0 240pt 0pt 0pt},clip]{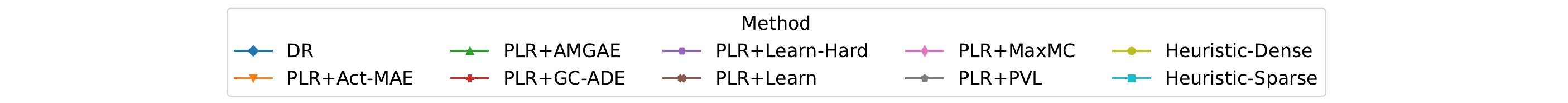}
    \end{subfigure}
    \begin{subfigure}[b]{.49\textwidth}
        \centering
        \includegraphics[width=\textwidth,trim={0 0pt 675pt 90pt},clip]{figures/performance_shadow_plots_train_experiments_step1.pdf}
        \vspace*{-6mm}
        \caption{Evaluation in training scenarios}
        \label{fig:experiment_step1_performance_train}
    \end{subfigure}
    \vspace{-.5mm}
    \begin{subfigure}[b]{.49\textwidth}
        \centering
        \includegraphics[width=\textwidth,trim={0 0pt 675pt 95pt},clip]{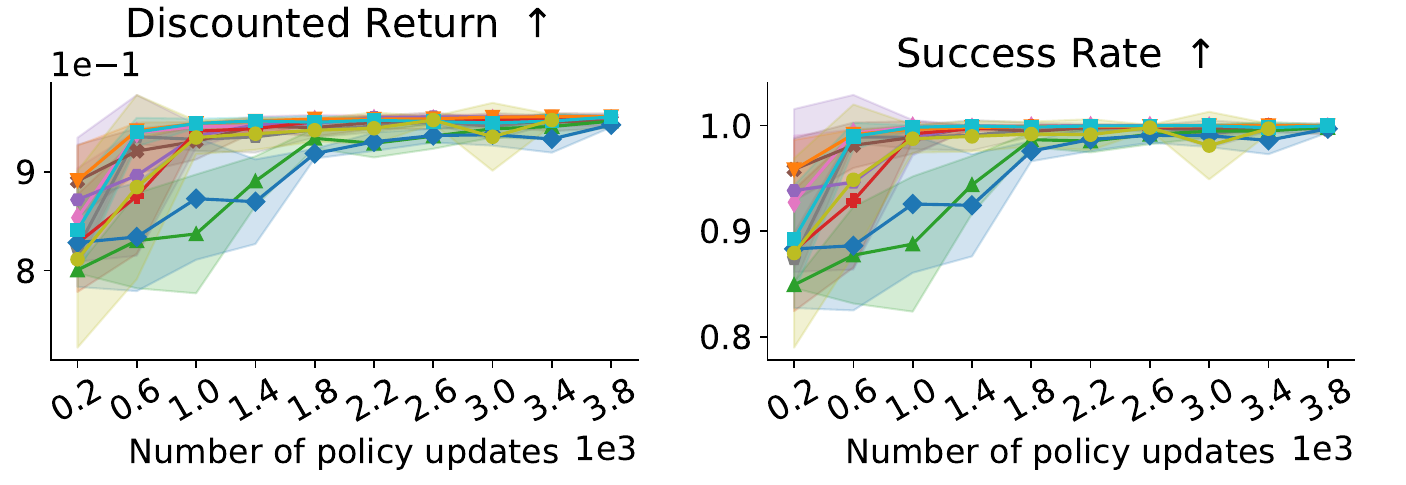}
        \vspace*{-6mm}
        \caption{Evaluation in unseen scenarios}
        \label{fig:experiment_step1_performance_test}
    \end{subfigure}
    \begin{subfigure}[b]{.45\textwidth}
        \centering
        \includegraphics[width=.6\linewidth,trim={50 0pt 45 160pt},clip]
{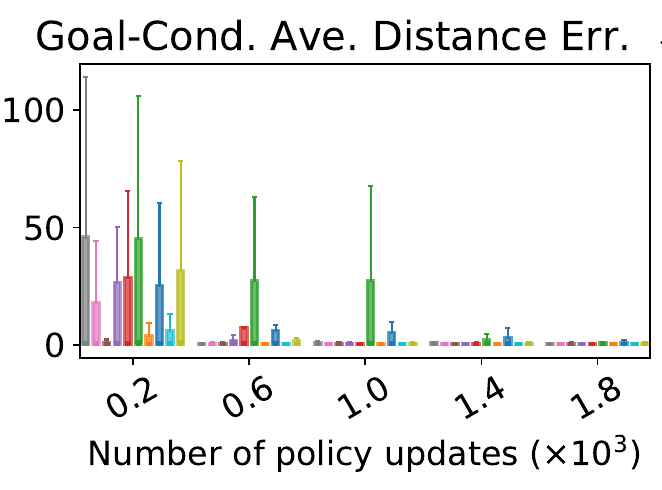}
        \vspace*{-3mm}
        \caption{Realism progression in unseen scenarios}
        \label{fig:experiment_step1_realism_test}
    \end{subfigure}
    \begin{subfigure}[b]{.45\textwidth}
        \centering
        \includegraphics[width=.6\linewidth,trim={405 0pt 50pt 120pt},clip]{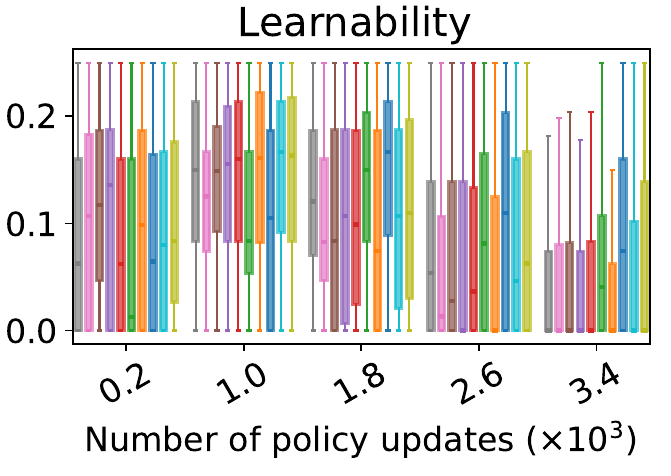}        \vspace*{-3mm}
        \caption{Learnability progression in training scenarios}
        \label{fig:experiment_step1_learnability_train}
    \end{subfigure}
    \vspace*{-2mm}
    \caption{Case 1: Training with 1000 scenarios. \textbf{(top)} We demonstrate performance progression in \textbf{(a)} training, and \textbf{(b)} test partitions, where bold markers indicate the mean and the shaded area covers one standard deviation across three independent runs. \textbf{(bottom)} Box plots illustrate quartiles of two utility functions: \textbf{(c)} $\GCADE$ and \textbf{(d)} $\Learnability$ progression in test/training splits.}
    \label{fig:experiment_step1_performance}
\end{figure*}

We implement $\CLForAD$ in $\GPUDRIVE$ \citep{kazemkhani2025gpudrive} and conduct experiments using traffic scenarios from WOMD \citep{ettinger2021large}. We assess performance via return and success rates, safety via collision and off-road rates, and realism via GC-ADE and metrics from the Waymo Open Sim Agents Challenge (WOSAC)\citep{montali2023waymo}. We train RL agents using self-play PPO following \cite{kazemkhani2025gpudrive,cornelisse2025building}, with sparse rewards for goal completion, collisions, and going off-road. Note that an episode does not terminate upon collision/off-road. Qualitatively, we analyze curriculum evolution, the correlation among utility functions and performance metrics, and how multi-agent self-play shapes the learning frontier. We compare DR against four heuristic curricula and seven PLR variants combined with the utility functions in \cref{fig:cl4ad_utility}. Heuristic-Dense/Sparse prioritize scenarios with high/low vehicle counts; Heuristic-Fast/Far prioritize by descending mean velocity and pairwise inter-agent distance, respectively. Note that we adopt $\PVL$ and $\MaxMC$ from Robust PLR and $\Learnability$ from SFL, rather than the algorithms themselves. See \cref{app:experimental_details} for more details on experiments.

\subsection{Can $\CLForAD$ accelerate learning?}
\textbf{PLR variants significantly accelerate learning, outperforming DR and heuristic curricula in both sample-efficiency and realism.} \cref{fig:experiment_step1_performance_train,fig:experiment_step1_performance_test} shows that PLR, with all utility functions except $\AMGAE$, achieves the highest returns and success rates in training scenarios. \cref{fig:experiments_step1_tts} evidences that, in test scenarios, PLR achieves $99\%$ success rate a billion steps earlier than DR, reducing wall-clock time by $77\%$. Compared with Heuristic-Sparse/Dense, PLR accelerates training to achieve the same success rate by 40\% and 66\%, respectively. Note that PLR with $\AMGAE$ outperforms DR with a small margin in terms of return. PLR also yields realistic policies faster than DR (see \cref{fig:experiment_step1_realism_test}). Lastly, \cref{fig:experiment_step1_learnability_train} illustrates that approaches that converge early obtain high learnability early on and achieve the lowest learnability fastest at the end.

\begin{figure*}[t]
    \centering
    \begin{subfigure}[b]{.36\linewidth}
    \centering
    \includegraphics[width=\linewidth,trim={0 0pt 0pt 0pt},clip]{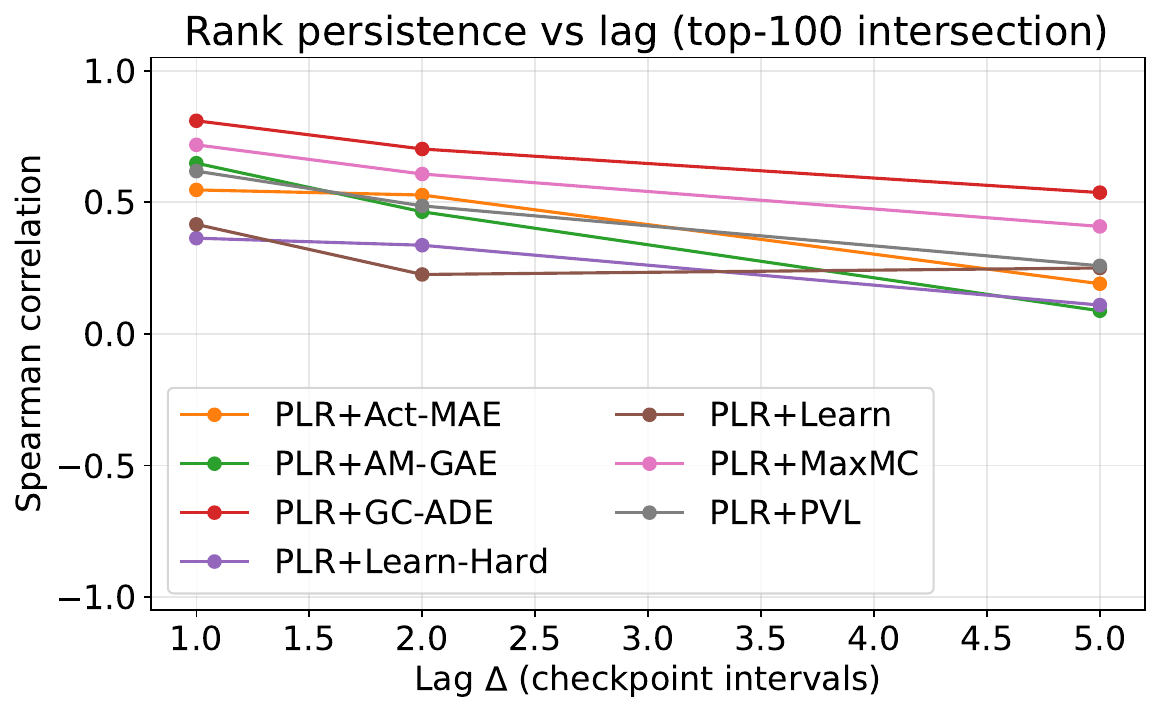}\\
    \includegraphics[width=\linewidth,trim={0 0pt 0pt 0pt},clip]{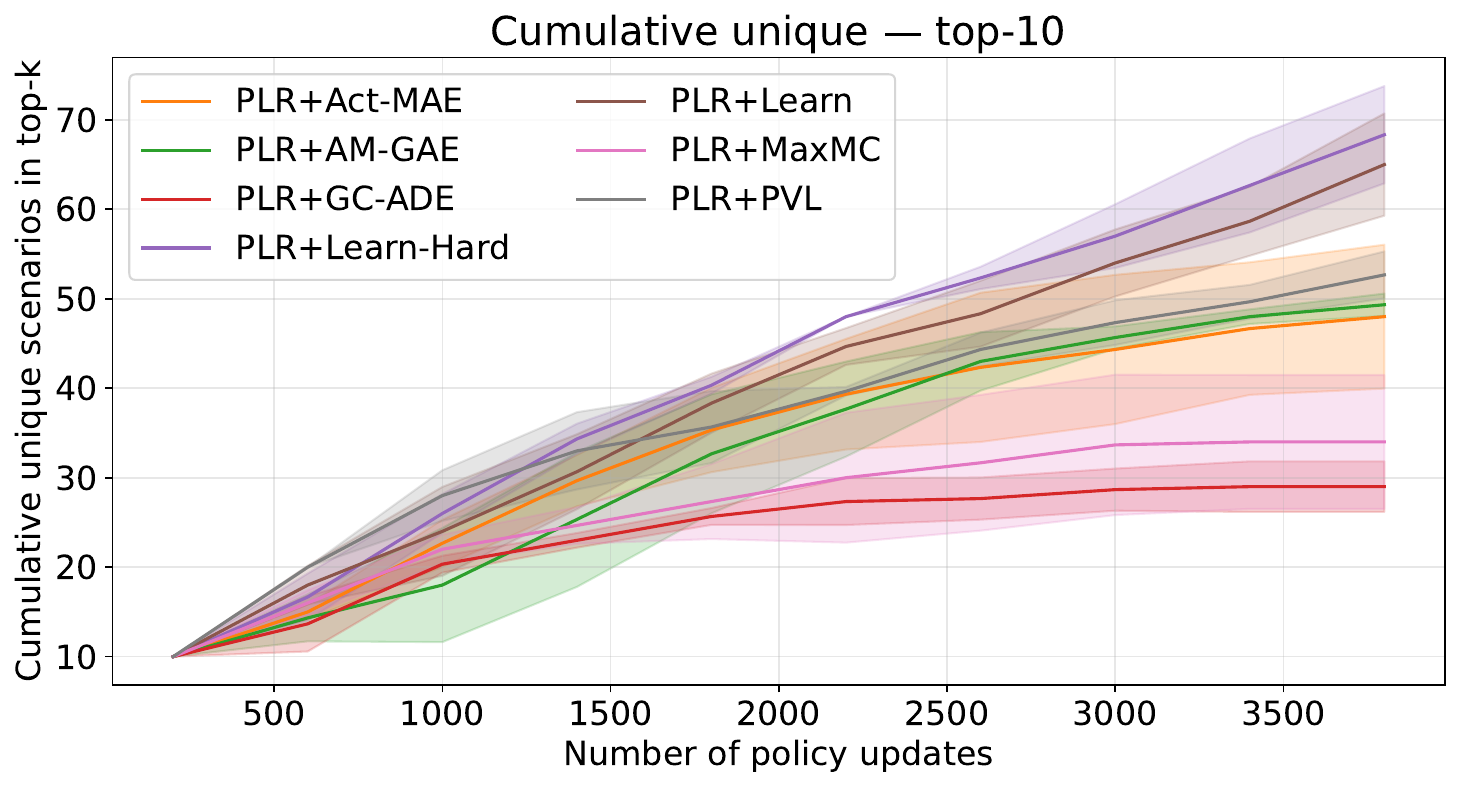}\\
    \includegraphics[width=\linewidth,trim={0 0pt 0pt 0pt},clip]{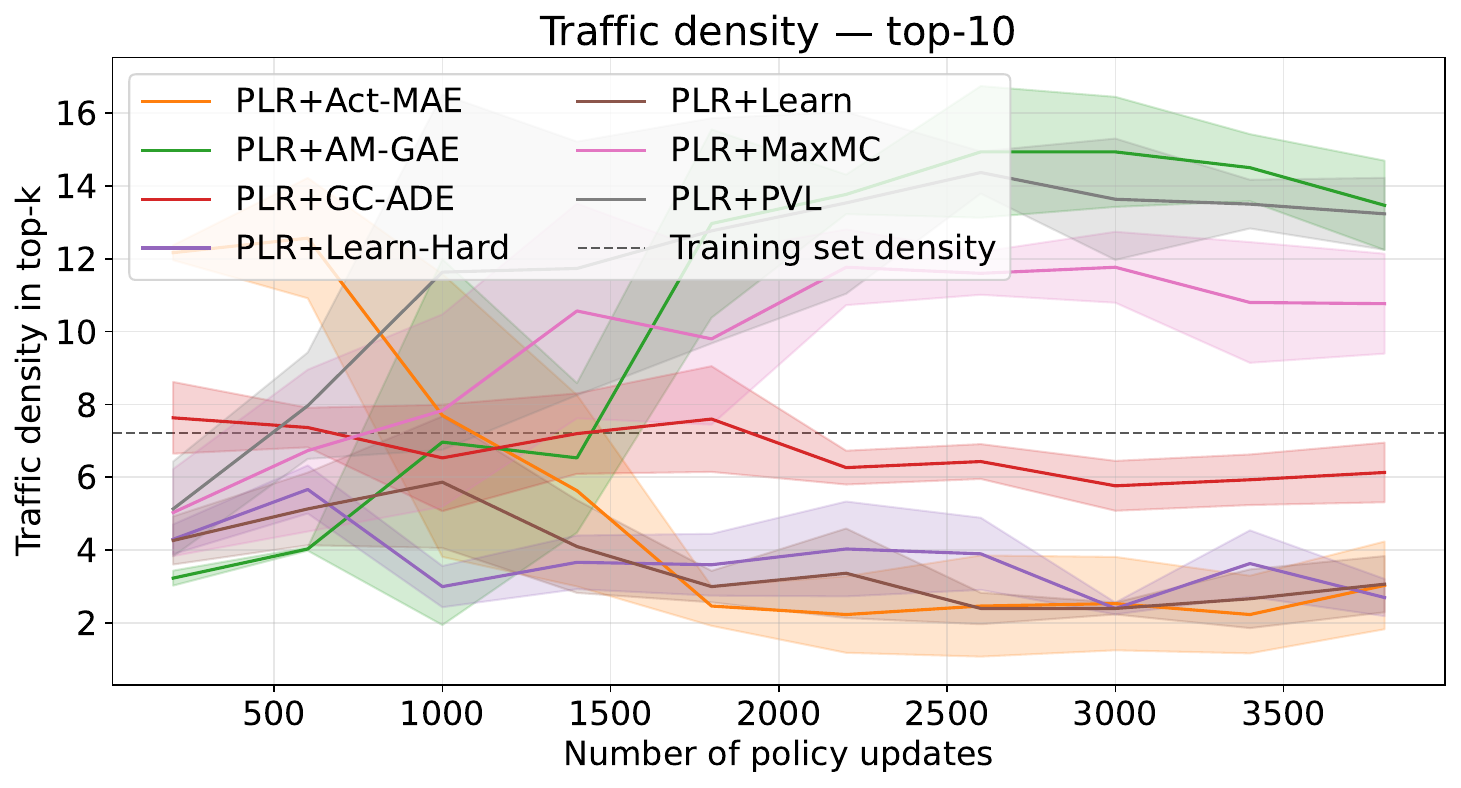}
    \vspace{-6mm}
    \caption{Score ranking evolution.}
    \label{fig:experiment_step1_score_ranking_evolution}
    \end{subfigure}
    \begin{subfigure}[b]{.61\linewidth}
    \centering
    \includegraphics[width=\linewidth,trim={0 0pt 0pt 00pt},clip]
    {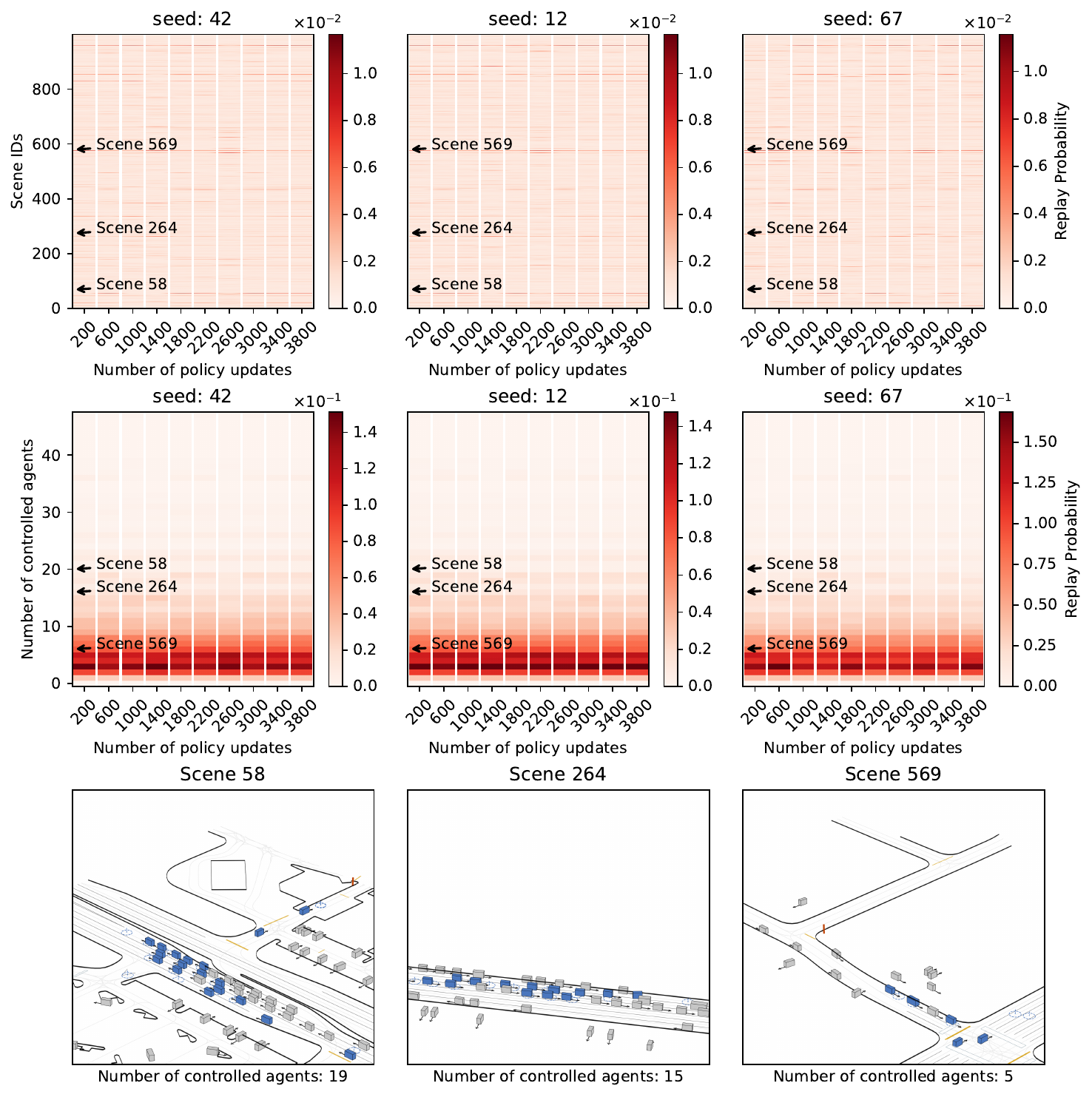}
    \vspace{-6mm}
    \caption{Progression of $\PLRReplayDistribution$ for $\MaxMC$}
    \label{fig:experiment_step1_replay_maxmc}
    \end{subfigure}
    \vspace{-2mm}
    \caption{Case 1: How CL4AD guides scenario selection. \textbf{(a)} We analyze score rankings: (top) Spearman rank correlation at increasing lags ($\Delta=1,2,5$) for the top-$100$ scenarios, where lower correlation indicates faster reshuffling; (middle) cumulative number of unique scenarios appearing in the top-$10$ over training; (bottom) mean number of controlled agents in the top-$10$ scenarios, with the dashed line indicating the training set average. \textbf{(b)} $\PLRReplayDistribution$ progression of PLR with $\MaxMC$: (top)  the evolution of $\PLRReplayDistribution$, darker segments are scenarios with higher likelihood, (middle) replay distribution with respect to the number of controlled agents, (bottom) three frequently replayed scenarios.}
\end{figure*}
\subsection{How does $\CLForAD$ guide scenario selection?}
\label{sec:experiments_step1_qualitative}
\textbf{$\CLForAD$ actively reshuffles score rankings, though the rate varies across utility functions.} The rank persistence plot (top row in \cref{fig:experiment_step1_score_ranking_evolution}) shows the Spearman rank correlation between score rankings. $\GCADE$ and $\MaxMC$ are the most rank-persistent, maintaining correlations of $0.4-0.55$ at $\Delta=5$. Other utility functions, particularly success-based and $\ACTMAE$, decorrelate more rapidly, reaching $0.1-0.25$ at $\Delta=5$. Cumulative number of unique scenarios (middle row) corroborates this: PLR with $\GCADE$ and $\MaxMC$ see $29$ and $34$, respectively, unique scenarios enter their top-10 over the full training run, while success-based and regret-based functions cycle through $48-70$.

\textbf{Different utility functions favor qualitatively distinct instances.} The progression of traffic density across the top-10 ranked scenarios (bottom row of \cref{fig:experiment_step1_score_ranking_evolution}) reveals that utility functions guide the curriculum to qualitatively distinct scenarios. Regret-based functions increasingly prioritize scenarios with traffic density above the dataset average. For example, \cref{fig:experiment_step1_replay_maxmc} shows the replay distribution progression of PLR with $\MaxMC$: dense scenarios such as Scenes $58$ and $264$  maintain high probability across runs, consistent with this mechanism. Success-based functions shift toward scenarios with low density as training progresses: scenarios with few controlled agents have the highest Bernoulli variance, since a single failure has a larger impact on the success rate.

\subsection{Is $\CLForAD$ effective under limited compute?}
\label{sec:ablation_compute_constraint_quantitative}
\begin{figure*}[t]
    \centering
    \begin{subfigure}[b]{.75\linewidth}
        \centering
    \includegraphics[width=\textwidth,trim={0pt 0pt 0pt 0pt},clip]{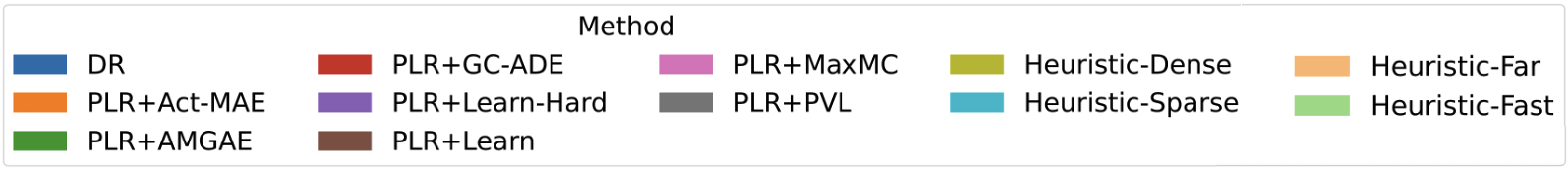}
    \end{subfigure}
    \begin{subfigure}[b]{0.35\linewidth}
        \centering
        \includegraphics[width=\textwidth,trim={75pt 0pt 75pt 80pt},clip]{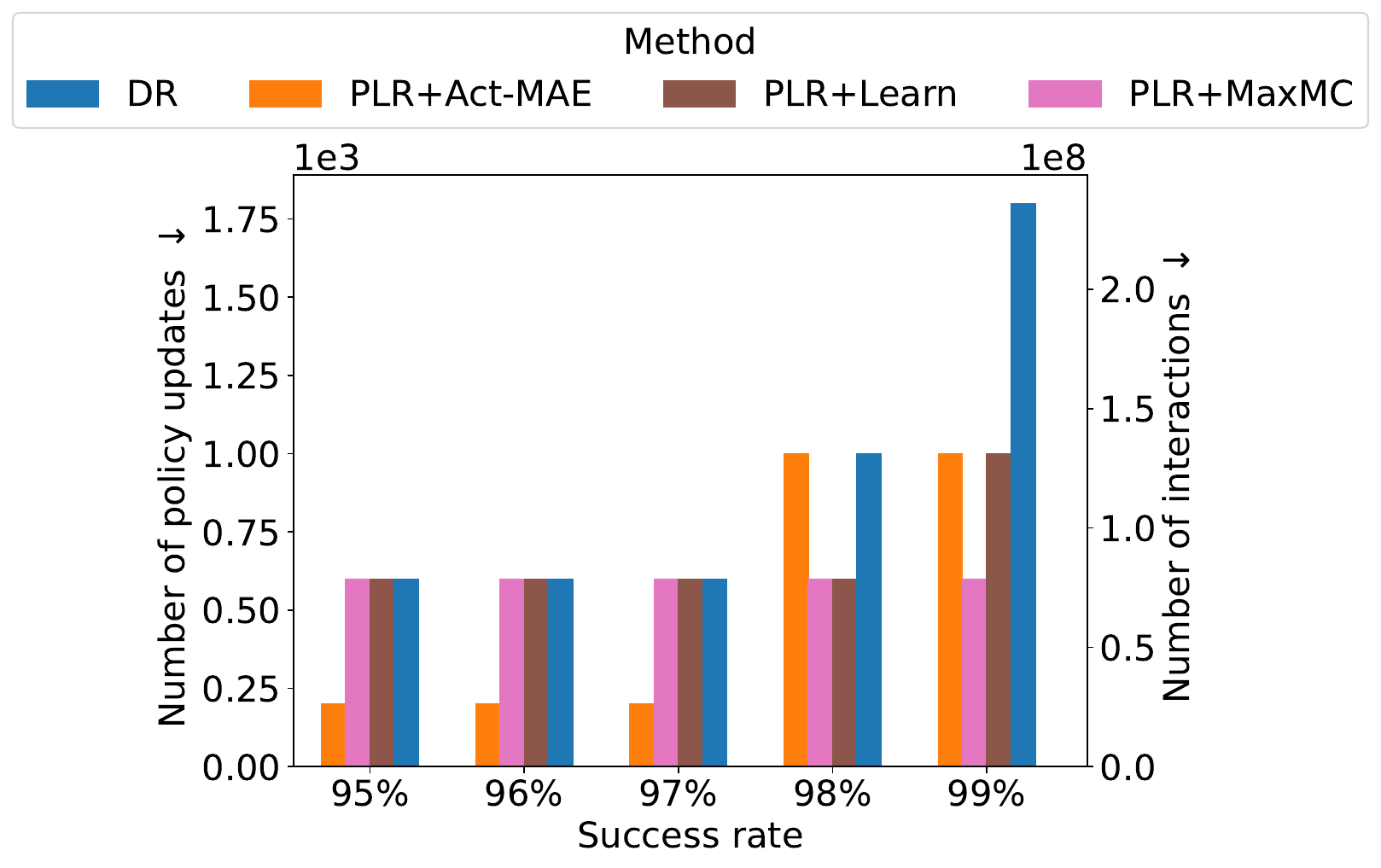}
        \vspace*{-6mm}
        \caption{Ablation: 1,000 scenarios}
        \label{fig:ablation_tts}
    \end{subfigure}
    \begin{subfigure}[b]{0.32\linewidth}
        \centering
        \includegraphics[width=\textwidth,trim={200pt 0pt 200pt 140pt},clip]{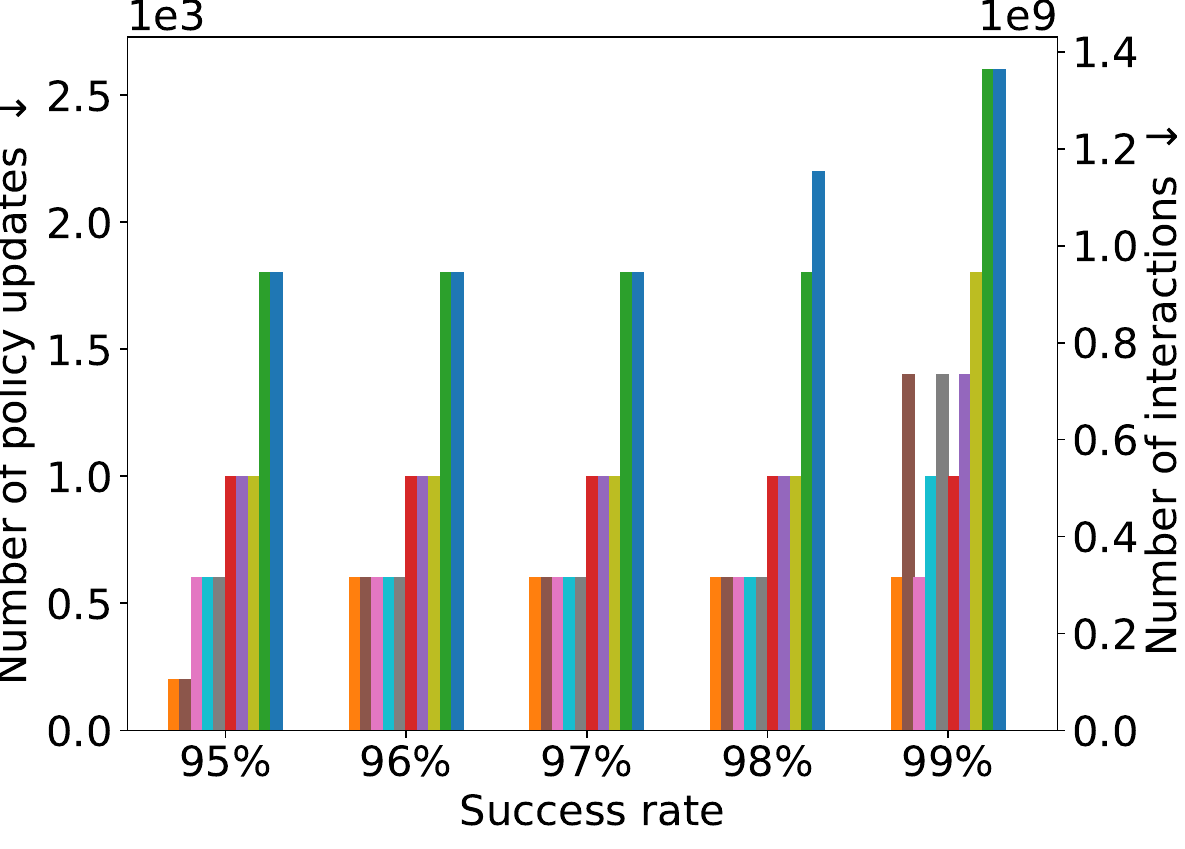}
        \vspace*{-6mm}
        \caption{Case 1: 1,000 scenarios}
        \label{fig:experiments_step1_tts}
    \end{subfigure}
    \\
    \begin{subfigure}[b]{0.32\linewidth}
        \centering
        \includegraphics[width=\textwidth,trim={175pt 0pt 175pt 110pt},clip]{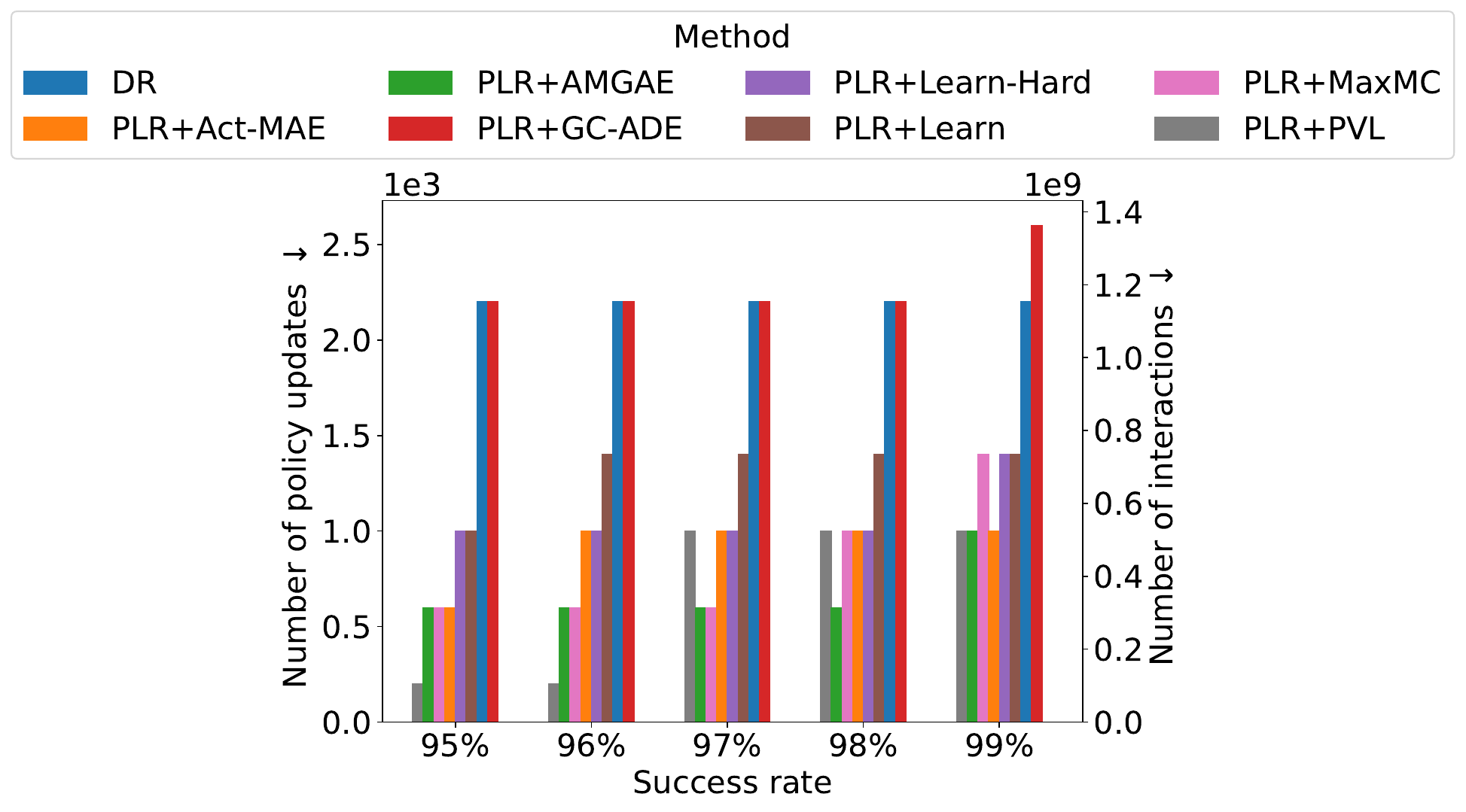}
        \vspace*{-6mm}
        \caption{Case 2: 10,000 scenarios}
        \label{fig:experiments_step2_tts}
    \end{subfigure}
    \hspace{3mm}
    \begin{subfigure}[b]{0.32\linewidth}
        \centering
        \includegraphics[width=\textwidth,trim={175pt 0pt 175pt 110pt},clip]{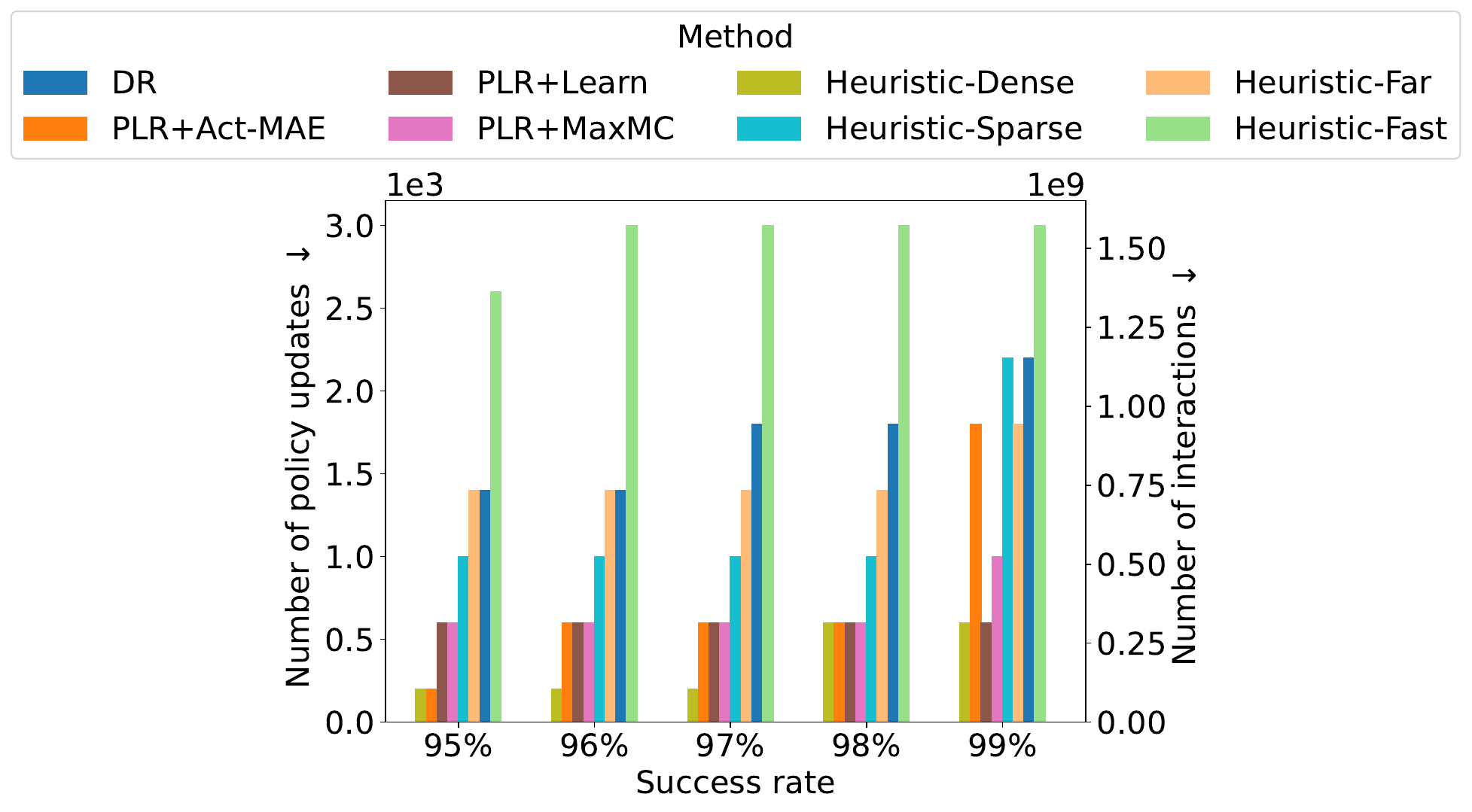}
        \vspace*{-6mm}
        \caption{Case 3: 80,000 scenarios}
        \label{fig:experiments_step3_tts}
    \end{subfigure}
    \vspace*{-2mm}
    \caption{Success progression in unseen test scenarios: \textbf{(a)} ablation study under compute constraints, \textbf{(b-d)} training in 1,000, 10,000, and 80,000 scenarios, respectively.}
    \label{fig:ablation_and_step1_and_step2_and_step3}
\end{figure*}

\textbf{$\CLForAD$ remains effective under strict computational constraints, outperforming DR in time to success.} \cref{fig:ablation_tts} illustrates results from an ablation study using a GPU with significantly smaller memory, reducing the number of worlds $\NumberOfWorlds$ and the experience buffer $\InteractionSet$ (see \cref{app:computational_resources} ). Although a smaller buffer results in more frequent policy updates, this setup takes about four times as long in wall-clock time and limits the diversity of scenarios used for updates. DR needs fewer interactions than the regular set-up, yet PLR reaches a $99\%$ success rate $67\%$ faster than DR.

\subsection{How does $\CLForAD$ scale up the dataset?}
\label{sec:experiments_step2_and_step3_quantitative}
\textbf{$\CLForAD$ continues to deliver sample-efficiency gains as the number of training scenarios increases to tens of thousands.} We train self-play agents in (case 2) 10,000 and (case 3) 80,000 scenarios from WOMD. \cref{fig:experiments_step2_tts} demonstrates that PLR reduces the number of interactions needed to reach a $99\%$ success rate by over $55\%$, when combined with $\MaxMC$ and $\ACTMAE$ in case 2. Similarly, \cref{fig:experiments_step3_tts} shows that PLR improves sample-efficiency by $72\%$ when combined with $\Learnability$ in case 3. Here, only Heuristic-Dense matches PLR, whereas the other heuristics offer no advantage.

\subsection{Do utility functions correlate with each other and performance metrics?}
\label{sec:correlation}
\begin{figure*}[t]{}
    \centering
    \centering\vspace{-0.5\baselineskip}
    \begin{subfigure}[b]{0.3\linewidth}
        \centering
        \includegraphics[width=\textwidth,trim={0pt 0pt 620pt 25pt},clip]{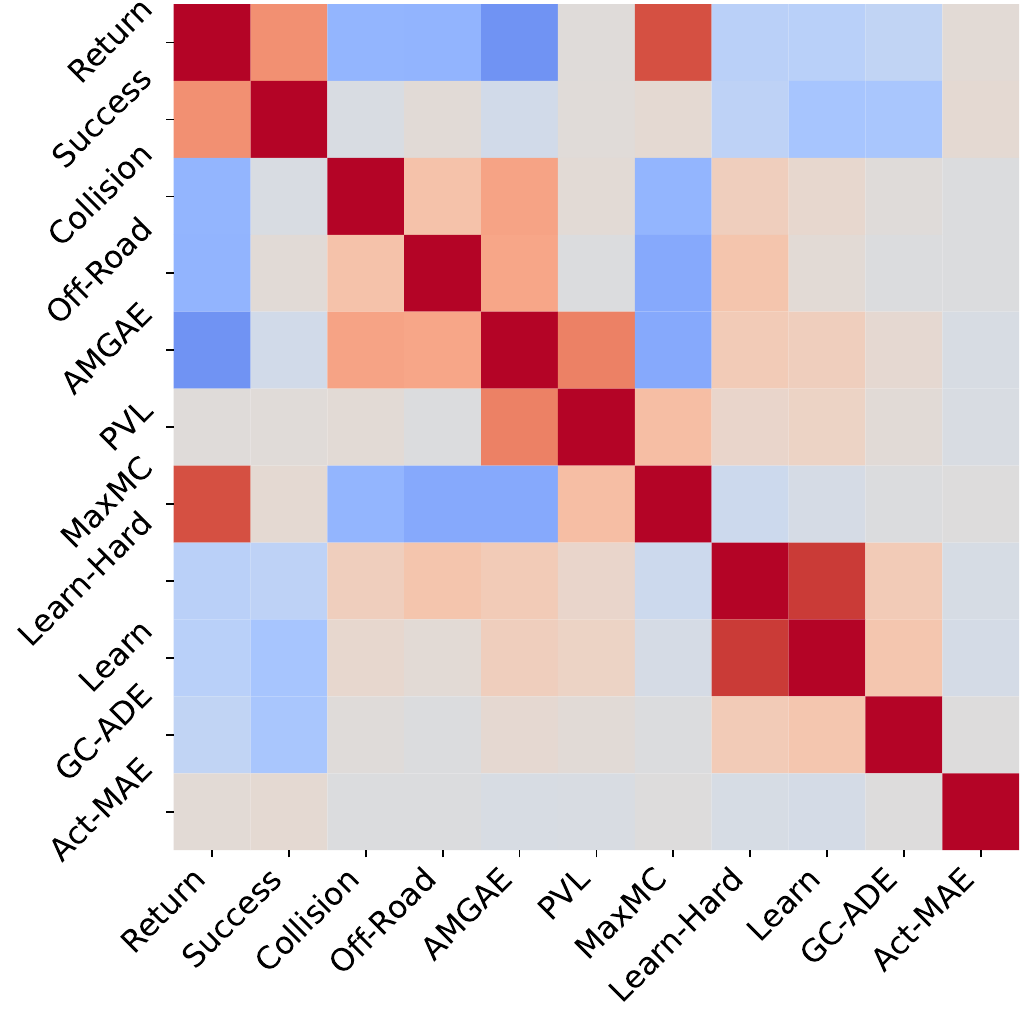}
        \vspace*{-6.5mm}
        \caption{Case 1: 1,000 scenarios}
        \label{fig:correlation_step1}
    \end{subfigure}
    \begin{subfigure}[b]{0.3\linewidth}
        \centering
        \includegraphics[width=\textwidth,trim={0pt 0pt 620pt 25pt},clip]{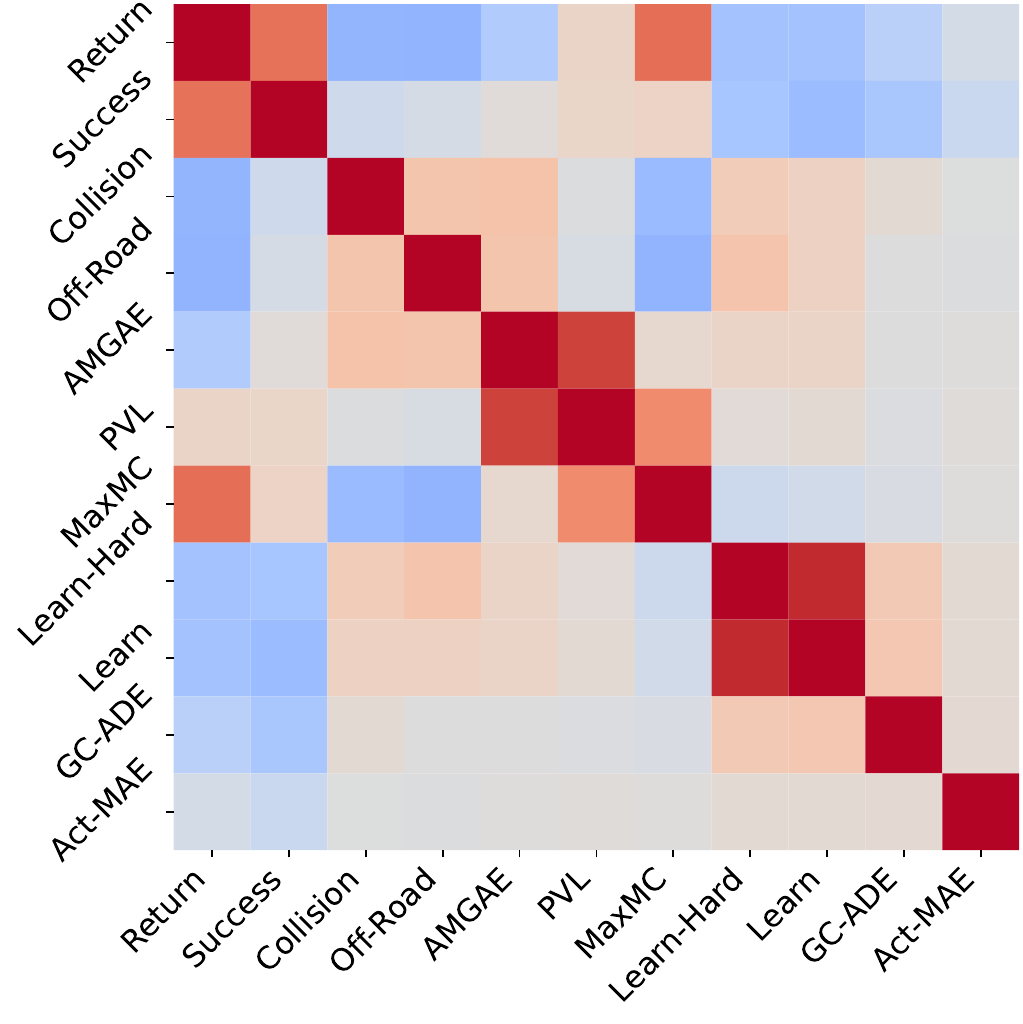}
        \vspace*{-6.5mm}
        \caption{Case 2: 10,000 scenarios}
        \label{fig:correlation_step2}
    \end{subfigure}
    \begin{subfigure}[b]{0.3\linewidth}
        \centering
        \includegraphics[width=\textwidth,trim={0pt 0pt 620pt 25pt},clip]{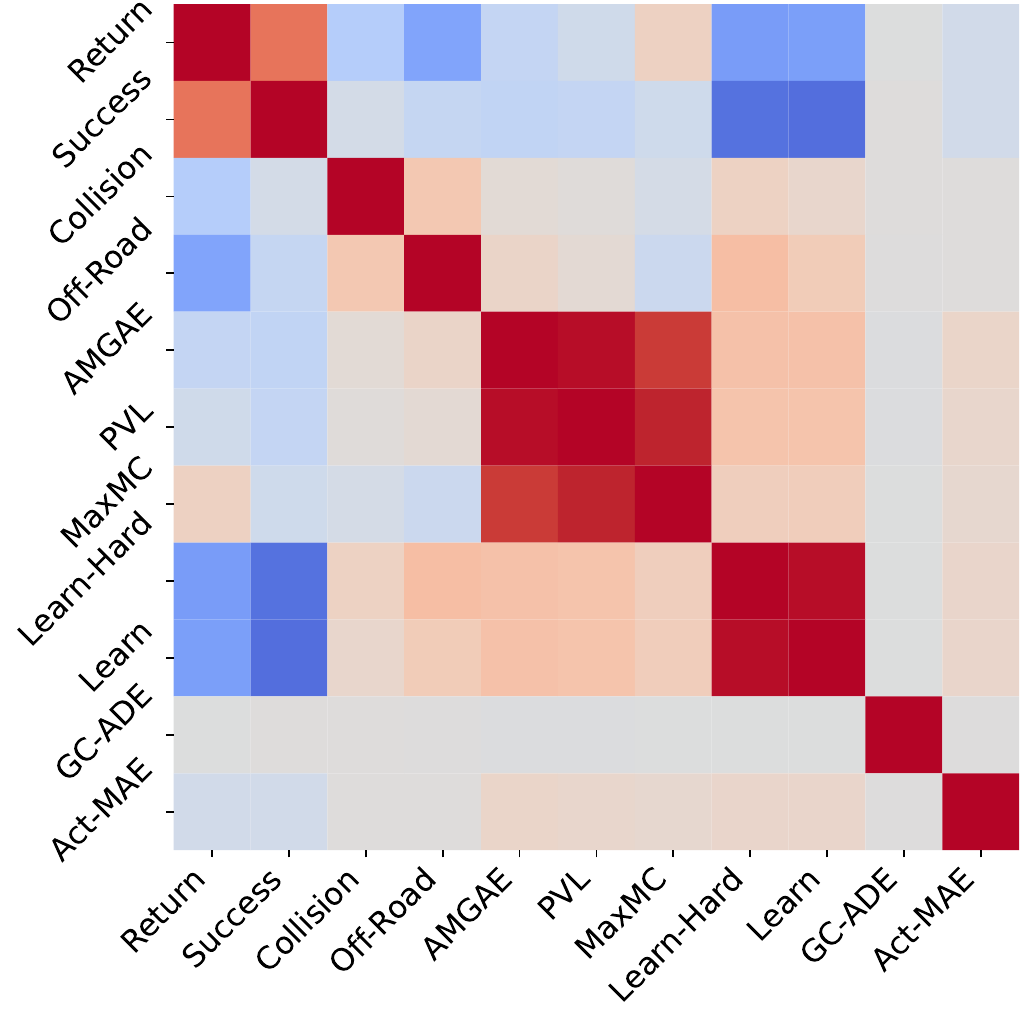}
        \vspace*{-6.5mm}
        \caption{Case 3: 80,000 scenarios}
        \label{fig:correlation_step3}
    \end{subfigure}
    \begin{subfigure}[b]{0.075\linewidth}
        \centering
        \includegraphics[width=\textwidth,trim={1000pt 0pt 0pt 20pt},clip]{figures/correlation_heatmaps_experiments_step3.pdf}
        \vspace*{-3mm}
        \label{fig:correlation_step3}
    \end{subfigure}
    \vspace{-2mm}
    \caption{Pearson correlation between utility functions and performance metrics, i.e, success, collision, and off-road rates: Results from training in \textbf{(a)} 1,000 \textbf{(b)} 10,000, and \textbf{(c)} 80,000 scenarios.}
    \label{fig:correlation}
\end{figure*}

\cref{fig:correlation} presents correlations between utility functions and performance metrics across all training cases. We include policies from multiple checkpoints to capture agents at varying stages of learning.

\textbf{Regret-based functions differ in how they handle sparse rewards, and their approximations improve with dataset scale.} $\AMGAE$ takes the absolute value of GAE, amplifying collision and off-road penalties into high-utility signals since episodes do not terminate on collision/off-road events. This explains $\AMGAE$'s positive correlation with collision/off-road rates in \cref{fig:correlation}. $\PVL$ discards negative TD errors via a ramp function, avoiding crash amplification, but its bootstrapped value targets are noisy early in training, explaining its lack of correlation with performance metrics. $\MaxMC$ uses empirical returns, avoiding both issues. The correlation between regret-based functions increases as the training set grows, because more diverse experience improves value estimation, consistent with $\AMGAE$ and $\PVL$ performing better in case 2 than in case 1.

\textbf{Success-based functions target the learning frontier by construction.} $\Learnability$ measures Bernoulli variance $\SuccessRate(1-\SuccessRate)$, peaking at $\SuccessRate=0.5$ and prioritizing scenarios where the policy succeeds half of the time. The negative correlation with return and success in \cref{fig:correlation} makes sense as the dominant signal comes from later training phases, where high success ($\SuccessRate\to1$) corresponds to low learnability. The positive correlation with collision/off-road rates follows the same logic: scenarios where collisions still occur have intermediate success rates and hence higher Bernoulli variance. $\LearnabilityHard$ counts only collision-free, on-road goal completions, making it a stricter measure of frontier scenarios.

\textbf{Realism-based utilities capture information orthogonal to performance.} $\GCADE$ and $\ACTMAE$ are standard AD evaluation metrics repurposed as utility functions. They affect only scenario prioritization, whereas the reward function captures only goal completion and collision/off-road avoidance. Therefore, realism-based functions cannot steer behavior toward realism. This explains their lack of correlation with performance metrics in \cref{fig:correlation}. $\GCADE$ and $\ACTMAE$ do not correlate with each other, either, as trajectory divergence compounds over time, while action divergence does not. However, they still accelerate training relative to DR, because early in training, high behavioral divergence from logged trajectories serves as an implicit proxy for scenario difficulty. Once the policy reaches goals reliably, divergence reflects only the gap between reward-optimal and logged driving, which explains the lack of correlation across checkpoints.

\textbf{Cross-category combinations are viable.} The correlation analysis suggests that pairs drawn from different categories carry complementary information. We combine two utility functions by ranking scenarios independently under each and sampling from the averaged rank. \cref{app:combination} reports three such pairs in case 1, which converge with their individual components and exceed DR throughout, and $\MaxMC$+$\LearnabilityHard$ sits at or above both of its components early in training. Combining a success-based function with a regret-based one therefore does not average their behavior.

\subsection{How does multi-agent self-play affect curricula at scale?}
\label{sec:multi_agent}
\textbf{Multi-agent self-play creates a persistent learning frontier that is exaggerated at scale.} As discussed in \cref{sec:cl4ad}, the utility of a scenario reflects the collective performance of the self-play policy. When a policy update improves some agents, it changes their trajectories, creating new conflicts for others in the same scenario. This coupling keeps scenario utility elevated, as improving some agents continuously introduces new difficulty for others. Dense scenarios amplify this as more agents means more coupled interactions. At the scale of batched simulators, each policy update consumes over half a million interactions across hundreds of scenarios, causing large policy shifts that further exaggerate the disruption and rapidly make buffer scores off-policy. We address this via higher score temperatures (see \cref{tab:parameters_plr}), which spread sampling probability across more scenarios.

\subsection{Does reward-optimal driving imply realistic driving?}
\label{sec:realism_vs_optimality}
\textbf{Reward optimality and behavioral realism can be orthogonal objectives.} $\GPUDRIVE$'s reward function incentivizes goal completion and penalizes collisions/off-road events, but does not account for realism, e.g., comfort, speed limit compliance, or smooth lane changes. An RL agent can therefore behave optimally yet not so realistically. We run an evaluation using WOSAC metrics to assess realism (see \cref{app:wosac}). DR achieves higher map-based scores than PLR variants because WOSAC measures distributional similarity to logged behavior, and uniform sampling stays closer to the data distribution. PLR variants achieve lower displacement error, consistent with more efficient goal-reaching, but diverge from human-like driving. Curricula cannot steer towards realism, as utility functions affect only scenario prioritization, not the reward. Improving both performance and realism requires reward functions that explicitly incentivize realistic or penalize unrealistic behavior.

\subsection{What guidelines can we derive for practitioners?}

\textbf{A full-size replay buffer and a score temperature well above prior work perform best at this scale.} A replay buffer equal to the training set works consistently as the dataset grows, in contrast to prior UED work, which uses buffers much smaller than the level space. Score temperatures above the typical range of $\PLRScoreTemperature\in[0.1, 1.0]$ perform better, and higher values become preferable as the dataset scales, since low temperatures concentrate sampling on top-ranked scenarios. The staleness coefficient remains in the range used in prior work \cite{jiang2021prioritized}, as higher temperature already spreads sampling across more scenarios. In addition, \cref{app:sensitivity} varies the sampling interval and the buffer size in case 1, where all configurations reach the same performance and exceed DR throughout, though a small buffer carries higher collision and off-road rates early in training.

\textbf{On utility functions, no single choice dominates across scales, though patterns emerge.} $\MaxMC$ is the most reliable regret-based function, as it does not depend on value function accuracy, whereas $\AMGAE$ and $\PVL$ underperform in case 1 and improve as the dataset grows. Success-based functions are effective at every scale and strongest in case 3, where success variance identifies the learning frontier. Realism-based functions accelerate training over DR but cannot improve realism, since they affect prioritization rather than the reward. Functions within a category prioritize similar scenarios, especially as the dataset scales, with realism as the exception. See \cref{app:practitioner_guidelines} for details.

\section{Conclusion}
\label{sec:conclusion}
In this work, we introduce $\CLForAD$, the first integration of CL into batched AD simulators. $\CLForAD$ frames scenario selection as a UED problem, enabling adaptive prioritization of traffic scenarios via PLR \citep{jiang2021prioritized} combined with utility functions that measure regret, success, and realism. Large-scale experiments on $\GPUDRIVE$ show that CL achieves $99\%$ success, up to $77\%$ faster than domain randomization, and outperforms heuristic curricula, with one exception at the largest scale, where the same heuristic offers no advantage at smaller ones. Our analysis reveals that different utility functions prioritize qualitatively different scenarios at varying rates; utility functions within the same category may correlate, but across categories capture distinct information; multi-agent self-play creates a persistent learning frontier that scale amplifies; and reward optimality and behavioral realism can be distinct axes. We derive practitioner guidelines for applying CL in batched AD simulators.

\textbf{Limitations and future work.} $\CLForAD$ evidences that CL scales up to the high-throughput of batched AD simulators. However, $\CLForAD$ is currently limited to variants of PLR and thus requires access to a real self-driving dataset as a source of traffic scenarios for sampling, e.g., WOMD, since $\GPUDRIVE$ operates on predefined scenarios. To address these limitations, future work will explore UED methods such as ACCEL \citep{parker2022evolving}, which randomly mutates prioritized scenarios, hence increasing scenario diversity for training robust policies. Applying ACCEL to GPUDrive requires a mutation operator over scenarios, e.g., one that perturbs the initial positions and velocities of agents, inserts or removes agents, or edits their goals, while keeping the scene drivable and consistent with the road graph. PAIRED and RE-PAIRED instead require a teacher policy that emits such scenario parameters directly, which entails a generative scenario representation rather than the identification numbers GPUDrive currently operates with. In addition, synthetic scenario generation tools, e.g., Scenario Dreamer \citep{rowe2025scenario}, can enable $\CLForAD$ to further accelerate training and improve the robustness and generalization capabilities of trained agents by creating safety-critical or out-of-distribution scenarios that the agent struggles with. $\CLForAD$ also trains on all sampled scenarios, whereas Robust PLR suppresses gradient updates on newly sampled levels and reports improved zero-shot transfer under sharper prioritization than our score temperatures induce. Whether that restriction transfers to batched self-play remains untested.

\clearpage
\bibliographystyle{plainnat}
\bibliography{neurips}

@inproceedings{
cusumano-towner2025robust,
title={Robust Autonomy Emerges from Self-Play},
author={Marco Francis Cusumano-Towner and David Hafner and Alexander Hertzberg and Brody Huval and Aleksei Petrenko and Eugene Vinitsky and Erik Wijmans and Taylor W. Killian and Stuart Bowers and Ozan Sener and Philipp Kraehenbuehl and Vladlen Koltun},
booktitle={Forty-second International Conference on Machine Learning},
year={2025},
}

@inproceedings{
kazemkhani2025gpudrive,
title={{GPUD}rive: Data-driven, multi-agent driving simulation at 1 million {FPS}},
author={Saman Kazemkhani and Aarav Pandya and Daphne Cornelisse and Brennan Shacklett and Eugene Vinitsky},
booktitle={The Thirteenth International Conference on Learning Representations},
year={2025},
}

@article{cornelisse2025building,
  title={Building reliable sim driving agents by scaling self-play},
  author={Cornelisse, Daphne and Pandya, Aarav and Joseph, Kevin and Su{\'a}rez, Joseph and Vinitsky, Eugene},
  journal={arXiv preprint arXiv:2502.14706},
  year={2025}
}

@inproceedings{ettinger2021large,
  title={Large scale interactive motion forecasting for autonomous driving: The waymo open motion dataset},
  author={Ettinger, Scott and Cheng, Shuyang and Caine, Benjamin and Liu, Chenxi and Zhao, Hang and Pradhan, Sabeek and Chai, Yuning and Sapp, Ben and Qi, Charles R and Zhou, Yin and others},
  booktitle={Proceedings of the IEEE/CVF international conference on computer vision},
  pages={9710--9719},
  year={2021}
}

@article{dosovitskiy2016learning,
  title={Learning to act by predicting the future},
  author={Dosovitskiy, Alexey and Koltun, Vladlen},
  journal={arXiv preprint arXiv:1611.01779},
  year={2016}
}

@article{caesar2021nuplan,
  title={nuplan: A closed-loop ml-based planning benchmark for autonomous vehicles},
  author={Caesar, Holger and Kabzan, Juraj and Tan, Kok Seang and Fong, Whye Kit and Wolff, Eric and Lang, Alex and Fletcher, Luke and Beijbom, Oscar and Omari, Sammy},
  journal={arXiv preprint arXiv:2106.11810},
  year={2021}
}

@article{silver2017mastering,
  title={Mastering the game of go without human knowledge},
  author={Silver, David and Schrittwieser, Julian and Simonyan, Karen and Antonoglou, Ioannis and Huang, Aja and Guez, Arthur and Hubert, Thomas and Baker, Lucas and Lai, Matthew and Bolton, Adrian and others},
  journal={nature},
  volume={550},
  number={7676},
  pages={354--359},
  year={2017},
  publisher={Nature Publishing Group UK London}
}

@inproceedings{bauer2023human,
  title={Human-timescale adaptation in an open-ended task space},
  author={Bauer, Jakob and Baumli, Kate and Behbahani, Feryal and Bhoopchand, Avishkar and Bradley-Schmieg, Nathalie and Chang, Michael and Clay, Natalie and Collister, Adrian and Dasagi, Vibhavari and Gonzalez, Lucy and others},
  booktitle={International Conference on Machine Learning},
  pages={1887--1935},
  year={2023},
  organization={PMLR}
}

@inproceedings{
zhang2024omni,
title={{OMNI}: Open-endedness via Models of human Notions of Interestingness},
author={Jenny Zhang and Joel Lehman and Kenneth Stanley and Jeff Clune},
booktitle={The Twelfth International Conference on Learning Representations},
year={2024},
}

@article{brunnbauer2024scenario,
  title={Scenario-based curriculum generation for multi-agent autonomous driving},
  author={Brunnbauer, Axel and Berducci, Luigi and Priller, Peter and Nickovic, Dejan and Grosu, Radu},
  journal={arXiv preprint arXiv:2403.17805},
  year={2024}
}

@article{dennis2020emergent,
  title={Emergent complexity and zero-shot transfer via unsupervised environment design},
  author={Dennis, Michael and Jaques, Natasha and Vinitsky, Eugene and Bayen, Alexandre and Russell, Stuart and Critch, Andrew and Levine, Sergey},
  journal={Advances in neural information processing systems},
  volume={33},
  pages={13049--13061},
  year={2020}
}

@inproceedings{jiang2021prioritized,
  title={Prioritized level replay},
  author={Jiang, Minqi and Grefenstette, Edward and Rockt{\"a}schel, Tim},
  booktitle={International Conference on Machine Learning},
  pages={4940--4950},
  year={2021},
  organization={PMLR}
}

@article{jiang2021replay,
  title={Replay-guided adversarial environment design},
  author={Jiang, Minqi and Dennis, Michael and Parker-Holder, Jack and Foerster, Jakob and Grefenstette, Edward and Rockt{\"a}schel, Tim},
  journal={Advances in Neural Information Processing Systems},
  volume={34},
  pages={1884--1897},
  year={2021}
}

@inproceedings{parker2022evolving,
  title={Evolving curricula with regret-based environment design},
  author={Parker-Holder, Jack and Jiang, Minqi and Dennis, Michael and Samvelyan, Mikayel and Foerster, Jakob and Grefenstette, Edward and Rockt{\"a}schel, Tim},
  booktitle={International Conference on Machine Learning},
  pages={17473--17498},
  year={2022},
  organization={PMLR}
}

@article{rutherford2024no,
  title={No regrets: Investigating and improving regret approximations for curriculum discovery},
  author={Rutherford, Alexander and Beukman, Michael and Willi, Timon and Lacerda, Bruno and Hawes, Nick and Foerster, Jakob},
  journal={Advances in Neural Information Processing Systems},
  volume={37},
  pages={16071--16101},
  year={2024}
}

@article{schulman2015high,
  title={High-dimensional continuous control using generalized advantage estimation},
  author={Schulman, John and Moritz, Philipp and Levine, Sergey and Jordan, Michael and Abbeel, Pieter},
  journal={arXiv preprint arXiv:1506.02438},
  year={2015}
}

@article{
tzannetos2023proximal,
title={Proximal Curriculum for Reinforcement Learning Agents},
author={Georgios Tzannetos and B{\'a}rbara Gomes Ribeiro and Parameswaran Kamalaruban and Adish Singla},
journal={Transactions on Machine Learning Research},
issn={2835-8856},
year={2023},
note={}
}

@article{schulman2017proximal,
  title={Proximal policy optimization algorithms},
  author={Schulman, John and Wolski, Filip and Dhariwal, Prafulla and Radford, Alec and Klimov, Oleg},
  journal={arXiv preprint arXiv:1707.06347},
  year={2017}
}

@inproceedings{
cornelisse2024humancompatible,
title={Human-compatible driving agents through data-regularized self-play reinforcement learning},
author={Daphne Cornelisse and Eugene Vinitsky},
booktitle={Reinforcement Learning Conference},
year={2024},
}

@article{gulino2023waymax,
  title={Waymax: An accelerated, data-driven simulator for large-scale autonomous driving research},
  author={Gulino, Cole and Fu, Justin and Luo, Wenjie and Tucker, George and Bronstein, Eli and Lu, Yiren and Harb, Jean and Pan, Xinlei and Wang, Yan and Chen, Xiangyu and others},
  journal={Advances in Neural Information Processing Systems},
  volume={36},
  pages={7730--7742},
  year={2023}
}

@article{vinitsky2022nocturne,
  title={Nocturne: a scalable driving benchmark for bringing multi-agent learning one step closer to the real world},
  author={Vinitsky, Eugene and Lichtl{\'e}, Nathan and Yang, Xiaomeng and Amos, Brandon and Foerster, Jakob},
  journal={Advances in Neural Information Processing Systems},
  volume={35},
  pages={3962--3974},
  year={2022}
}

@inproceedings{dosovitskiy2017carla,
  title={CARLA: An open urban driving simulator},
  author={Dosovitskiy, Alexey and Ros, German and Codevilla, Felipe and Lopez, Antonio and Koltun, Vladlen},
  booktitle={Conference on robot learning},
  pages={1--16},
  year={2017},
  organization={PMLR}
}

@article{narvekar2020curriculum,
  title={Curriculum learning for reinforcement learning domains: A framework and survey},
  author={Narvekar, Sanmit and Peng, Bei and Leonetti, Matteo and Sinapov, Jivko and Taylor, Matthew E and Stone, Peter},
  journal={Journal of Machine Learning Research},
  volume={21},
  number={181},
  pages={1--50},
  year={2020}
}

@inproceedings{baranes2010intrinsically,
  title={Intrinsically motivated goal exploration for active motor learning in robots: A case study},
  author={Baranes, Adrien and Oudeyer, Pierre-Yves},
  booktitle={2010 IEEE/RSJ International Conference on Intelligent Robots and Systems},
  pages={1766--1773},
  year={2010},
  organization={IEEE}
}

@inproceedings{florensa2018automatic,
  title={Automatic goal generation for reinforcement learning agents},
  author={Florensa, Carlos and Held, David and Geng, Xinyang and Abbeel, Pieter},
  booktitle={International conference on machine learning},
  pages={1515--1528},
  year={2018},
  organization={PMLR}
}

@inproceedings{klink2022curriculum,
  title={Curriculum reinforcement learning via constrained optimal transport},
  author={Klink, Pascal and Yang, Haoyi and D’Eramo, Carlo and Peters, Jan and Pajarinen, Joni},
  booktitle={International Conference on Machine Learning},
  pages={11341--11358},
  year={2022},
  organization={PMLR}
}

@inproceedings{koprulu2023risk,
  title={Risk-aware curriculum generation for heavy-tailed task distributions},
  author={Koprulu, Cevahir and Sim{\~a}o, Thiago D and Jansen, Nils and Topcu, Ufuk},
  booktitle={Uncertainty in Artificial Intelligence},
  pages={1132--1142},
  year={2023},
  organization={PMLR}
}

@article{sayar2024diffusion,
  title={Diffusion-based curriculum reinforcement learning},
  author={Sayar, Erdi and Iacca, Giovanni and Oguz, Ozgur S and Knoll, Alois},
  journal={Advances in Neural Information Processing Systems},
  volume={37},
  pages={97587--97617},
  year={2024}
}

@inproceedings{anzalone2021reinforced,
  title={Reinforced curriculum learning for autonomous driving in carla},
  author={Anzalone, Luca and Barra, Silvio and Nappi, Michele},
  booktitle={2021 IEEE International Conference on Image Processing (ICIP)},
  pages={3318--3322},
  year={2021},
  organization={IEEE}
}

@article{anzalone2022end,
  title={An end-to-end curriculum learning approach for autonomous driving scenarios},
  author={Anzalone, Luca and Barra, Paola and Barra, Silvio and Castiglione, Aniello and Nappi, Michele},
  journal={IEEE Transactions on Intelligent Transportation Systems},
  volume={23},
  number={10},
  pages={19817--19826},
  year={2022},
  publisher={IEEE}
}

@inproceedings{qiao2018automatically,
  title={Automatically generated curriculum based reinforcement learning for autonomous vehicles in urban environment},
  author={Qiao, Zhiqian and Muelling, Katharina and Dolan, John M and Palanisamy, Praveen and Mudalige, Priyantha},
  booktitle={2018 IEEE Intelligent Vehicles Symposium (IV)},
  pages={1233--1238},
  year={2018},
  organization={IEEE}
}

@article{abouelazm2025automatic,
  title={Automatic Curriculum Learning for Driving Scenarios: Towards Robust and Efficient Reinforcement Learning},
  author={Abouelazm, Ahmed and Weinstein, Tim and Joseph, Tim and Sch{\"o}rner, Philip and Z{\"o}llner, J Marius},
  journal={arXiv preprint arXiv:2505.08264},
  year={2025}
}

@inproceedings{rowe2025scenario,
  title={Scenario dreamer: Vectorized latent diffusion for generating driving simulation environments},
  author={Rowe, Luke and Girgis, Roger and Gosselin, Anthony and Paull, Liam and Pal, Christopher and Heide, Felix},
  booktitle={Proceedings of the Computer Vision and Pattern Recognition Conference},
  pages={17207--17218},
  year={2025}
}

@article{jackson2023discovering,
  title={Discovering general reinforcement learning algorithms with adversarial environment design},
  author={Jackson, Matthew T and Jiang, Minqi and Parker-Holder, Jack and Vuorio, Risto and Lu, Chris and Farquhar, Greg and Whiteson, Shimon and Foerster, Jakob},
  journal={Advances in Neural Information Processing Systems},
  volume={36},
  pages={79980--79998},
  year={2023}
}

@article{li2023scenarionet,
  title={Scenarionet: Open-source platform for large-scale traffic scenario simulation and modeling},
  author={Li, Quanyi and Peng, Zhenghao Mark and Feng, Lan and Liu, Zhizheng and Duan, Chenda and Mo, Wenjie and Zhou, Bolei},
  journal={Advances in neural information processing systems},
  volume={36},
  pages={3894--3920},
  year={2023}
}

@inproceedings{zhang2023cat,
  title={Cat: Closed-loop adversarial training for safe end-to-end driving},
  author={Zhang, Linrui and Peng, Zhenghao and Li, Quanyi and Zhou, Bolei},
  booktitle={Conference on Robot Learning},
  pages={2357--2372},
  year={2023},
  organization={PMLR}
}

@inproceedings{niu2024continual,
  title={Continual driving policy optimization with closed-loop individualized curricula},
  author={Niu, Haoyi and Xu, Yizhou and Jiang, Xingjian and Hu, Jianming},
  booktitle={2024 IEEE International Conference on Robotics and Automation (ICRA)},
  pages={6850--6857},
  year={2024},
  organization={IEEE}
}

@article{sheng2026curricuvlm,
  title={Curricuvlm: Towards safe autonomous driving via personalized safety-critical curriculum learning with vision-language models},
  author={Sheng, Zihao and Huang, Zilin and Qu, Yansong and Leng, Yue and Bhavanam, Sruthi and Chen, Sikai},
  journal={Transportation Research Part C: Emerging Technologies},
  volume={185},
  pages={105549},
  year={2026},
  publisher={Elsevier}
}

@article{montali2023waymo,
  title={The waymo open sim agents challenge},
  author={Montali, Nico and Lambert, John and Mougin, Paul and Kuefler, Alex and Rhinehart, Nicholas and Li, Michelle and Gulino, Cole and Emrich, Tristan and Yang, Zoey and Whiteson, Shimon and others},
  journal={Advances in Neural Information Processing Systems},
  volume={36},
  pages={59151--59171},
  year={2023}
}

@inproceedings{portelas2020teacher,
  title={Teacher algorithms for curriculum learning of deep rl in continuously parameterized environments},
  author={Portelas, R{\'e}my and Colas, C{\'e}dric and Hofmann, Katja and Oudeyer, Pierre-Yves},
  booktitle={Conference on Robot Learning},
  pages={835--853},
  year={2020},
  organization={PMLR}
}

@article{kanitscheider2021multi,
  title={Multi-task curriculum learning in a complex, visual, hard-exploration domain: Minecraft},
  author={Kanitscheider, Ingmar and Huizinga, Joost and Farhi, David and Guss, William Hebgen and Houghton, Brandon and Sampedro, Raul and Zhokhov, Peter and Baker, Bowen and Ecoffet, Adrien and Tang, Jie and others},
  journal={arXiv preprint arXiv:2106.14876},
  year={2021}
}

@inproceedings{
jiang2023minimax,
title={minimax: Efficient Baselines for Autocurricula in {JAX}},
author={Minqi Jiang and Michael D Dennis and Edward Grefenstette and Tim Rockt{\"a}schel},
booktitle={Second Agent Learning in Open-Endedness Workshop},
year={2023},
}

@article{sullivan2025syllabus,
    title={Syllabus: {P}ortable Curricula for Reinforcement Learning Agents},
    author={Sullivan, Ryan and P{\'{e}}goud, Ryan and Rehman, Ameen Ur and Yang, Xinchen and Huang, Junyun and Verma, Aayush and Mitra, Nistha and Dickerson, John P},
    journal={Reinforcement Learning Journal},
    volume={6},
    pages={1816--1855},
    year={2025}
}

\newpage
\appendix

\section{Nomenclature}
\label{app:nomenclature}
\bgroup
\def\arraystretch{1.5}
\begin{tabular}{p{1in}p{4.5in}}
$\POSG$ & POSG \\
$\POSGAgentSet$, $\POSGNumberOfAgents$ & Set of agents, number of agents ($|\POSGAgentSet|=\POSGNumberOfAgents$) in POSG \\
$\POSGStateSpace$, $\POSGActionSpace$, $\POSGObservationSpace$ & State, action and observation spaces in POSG \\ 
$\POSGState$, $\POSGAction$, $\POSGObservation$, $\POSGReward$ & State, action, observation, and reward in POSG\\
$\POSGTransitionFunction$, $\POSGObservationFunction$, $\POSGRewardFunction$ & Transition, observation, and reward functions in POSG\\
$\POSGInitialStateDistribution$ & Initial state distribution in POSG\\
$\POSGDiscount$, $\POSGHorizon$ & Discount factor and horizon in POSG\\
$\UPOSG$, $\UPOSGScenarioParameterSet$, $\UPOSGScenarioParameter$ & UPOSG, its set of scenarios and a scenario, i.e., $\UPOSGScenarioParameter\in\UPOSGScenarioParameterSet$ \\
$\UPOSGAgentSet$, $\UPOSGNumberOfAgents$ & Set of agents, number of agents ($|\UPOSGAgentSet|=\UPOSGNumberOfAgents$) in UPOSG \\
$\UPOSGStateSpace$, $\UPOSGActionSpace$, $\UPOSGObservationSpace$ & State, action and observation spaces in UPOSG \\ 
$\UPOSGState$, $\UPOSGAction$, $\UPOSGObservation$, $\UPOSGReward$ & State, action, observation, and reward in UPOSG\\
$\UPOSGTransitionFunction$, $\UPOSGObservationFunction$, $\UPOSGRewardFunction$, $\UPOSGInitialStateDistribution$ & Transition, observation, reward functions and initial state distribution in UPOSG\\
$\UPOSGDiscount$, $\UPOSGHorizon$ & Discount factor and horizon in UPOSG\\
$\UPOSGPosition$ & Position of an agent in a scenario in UPOSG \\
$\UPOSGNumberOfScenarios$ & Number of scenarios in a UPOSG, i.e., $|\UPOSGScenarioParameterSet|=\UPOSGNumberOfScenarios$ \\
$\UPOSGGoalStates$ & Goal states in a UPOSG \\
$\UEDLevelGenerator$ & Level generator for UED \\
$\UEDPolicySpace$ & Policy space in UED \\
$\UEDDistributionOverLevelS$ & Distribution over levels in UED \\
$\UEDUtilityFunction$, $\UEDConstantUtility$ & Utility function in UED, constant utility in UED \\
$\PLRBuffer$, $\PLRReplayDistribution$ & Replay buffer and distribution in PLR \\
$\PLRScoreDistribution$, $\PLRStalenessDistribution$  & Score and staleness distribution in PLR \\
$\PLRIterationNumber$ & Scenario sampling iteration in PLR \\
$\PLRStalenessCoefficient$, $\PLRScoreTemperature$, $\PLRReplayRate$, $\PLRMaxBufferSize$ & Staleness coefficient, score temperature, replay rate, max replay buffer size \\
$\Policy$, $\Value$ & Policy, value function \\
$\GAEDiscount$, $\TDError$ & GAE discount factor, TD error \\ 
$\MaximumReturnAgent$ & Maximum return of an agent \\
$\SuccessRate$ & Success rate \\
$\PolicyParameter$ & Trainable policy parameter \\
$\TotalIterations$, $\ScenarioSamplingInterval$, $\PolicyUpdateInterval$ & Number of interactions for training, sampling scenarios, and updating policy \\
$\NumberOfWorlds$ & Number of concurrent worlds \\
$\InteractionSet$ & Experience buffer \\
$\UpdatePolicy$ & RL algorithm of choice to update policy \\
$\EndOfEpisodeFlag$ & End of episode flag \\ 
$\Rollout$ & Rollout \\
\end{tabular}
\egroup

\newpage
\section{Details of $\CLForAD$}
\label{app:cl4ad}
\begin{algorithm}[t]
\caption{$\SampleFromCurriculum$}
\label{alg:cl4ad_sample}
\textbf{Input}: Replay buffer $\PLRBuffer$, set of training scenarios $\PLRTrainingLevels$, sampling iteration $\PLRIterationNumber$ \\
\textbf{Parameters}: Replay rate  $\PLRReplayRate$, staleness $\PLRStalenessCoefficient$, temperature $\PLRScoreTemperature$, max buffer size $\PLRMaxBufferSize$, number of worlds $\NumberOfWorlds$ \\
\textbf{Output}: Sampled scenarios $(\UPOSGScenarioParameter_{w})_{w=1}^{\NumberOfWorlds}$, and buffer $\PLRBuffer$ with updated staleness
\begin{algorithmic}[1] 
    \STATE $\PLRBuffer\leftarrow\textsc{DiscardLowestRankingScenarios}(\PLRBuffer,\PLRMaxBufferSize)$
    \IF{$|\PLRBuffer|\equiv 0~\OR~(\text{Bernoulli}(\PLRReplayRate)\equiv 0~\AND~|\PLRTrainingLevels-\PLRBuffer^{\text{scenario}}|>0$)}
    \STATE $\Probability_{\text{sample}}\leftarrow\text{Uniform}(\PLRTrainingLevels-\PLRBuffer^{\text{scenario}})$
    \COMMENT{Uniformly randomly sample scenarios}
    \ELSE
    \STATE $\Probability_{\text{sample}}\leftarrow\PLRReplayDistribution$
    \COMMENT{Replay scenarios based on $\PLRReplayDistribution$}
    \ENDIF
    \STATE $(\UPOSGScenarioParameter_{w})_{w=1}^{\NumberOfWorlds}\leftarrow\text{Sample}(\Probability_{\text{sample}},\NumberOfWorlds)$
    \COMMENT{Sample $\NumberOfWorlds$-many scenarios based on $\Probability_{\text{sample}}$}
    \STATE $\PLRBuffer^{\text{scenario}}\leftarrow\PLRBuffer^{\text{scenario}}\cup(\UPOSGScenarioParameter_{w})_{w=1}^{\NumberOfWorlds}$
    \COMMENT{Update scenarios in the replay buffer}
    \STATE $\PLRIterationNumber_{\UPOSGScenarioParameter_w}\leftarrow\PLRIterationNumber, \forall w\in[W]$
    \COMMENT{Update sampling iteration for staleness distribution}
    \STATE $\Rollout_{\UPOSGScenarioParameter_w}\leftarrow(), \forall w\in[W]$
    \COMMENT{Reset the rollout}
    
\end{algorithmic}

\end{algorithm}

\begin{algorithm}[t]
\caption{$\UpdateCurriculum$}
\label{alg:cl4ad_update}
\textbf{Input}: Interaction set $\InteractionSet_t$, utility function $\UEDUtilityFunction$, replay buffer $\PLRBuffer$ \\
\textbf{Output}: Updated replay buffer $\PLRBuffer$
\begin{algorithmic}[1] 
    \FOR{$w\in[\NumberOfWorlds]$}
    \IF{$\EndOfEpisodeFlag_{n,w}\text{ is True }\forall n\in[\UPOSGNumberOfAgents_{\UPOSGScenarioParameter_w}]$}
    \STATE $\text{score}_{\UPOSGScenarioParameter_w,t} \leftarrow\UEDUtilityFunction(\Rollout_{\UPOSGScenarioParameter_w})$
    \COMMENT{Compute utility score for terminated episode}
    \STATE $\text{score}_{\UPOSGScenarioParameter_w}\leftarrow\text{MovingAverage}(\text{score}_{\UPOSGScenarioParameter_w},\text{score}_{\UPOSGScenarioParameter_w,t})$
    \COMMENT{Update the score in the buffer}
    \STATE $\Rollout_{\UPOSGScenarioParameter_w}\leftarrow()$
    \COMMENT{Reset the rollout}
    \ELSE     
    \STATE $\Rollout_{\UPOSGScenarioParameter_w}\leftarrow\Rollout_{\UPOSGScenarioParameter_w}\cup\{\UPOSGObservation_{n,w},\UPOSGAction_{n,w},\UPOSGObservation'_{n,w},\UPOSGReward_{n,w},\EndOfEpisodeFlag_{n,w}\}_{n\in{[\UPOSGNumberOfAgents_{\UPOSGScenarioParameter_w}]}}$
    \COMMENT{Update the rollout with new interactions}
    \ENDIF
    \ENDFOR
\end{algorithmic}
\end{algorithm}

In this section, we provide a more detailed look into how $\CLForAD$ works to support the material in Section $\ref{sec:cl4ad}$ and share the wall-clock overhead of curriculum updates and sampling. 

\subsection{Curriculum Sampling in $\CLForAD$}
The score distribution $\PLRScoreDistribution(\UPOSGScenarioParameter_i|\PLRBuffer,\UEDUtilityFunction)$ is based on the ranking of seen levels,
\begin{equation}
    \PLRScoreDistribution(\UPOSGScenarioParameter_i|\PLRBuffer,\UEDUtilityFunction)=\frac{\text{rank}(\UPOSGScenarioParameter_i|\PLRBuffer)^{-1/\PLRScoreTemperature}}{\sum_{j\in\PLRBuffer^{\text{scenario}}}\text{rank}(\UPOSGScenarioParameter_j|\PLRBuffer)^{-1/\PLRScoreTemperature}},
    \label{eq:p_utility}
\end{equation}
with a temperature parameter $\PLRScoreTemperature$ tuning the impact of ranking. The staleness distribution assigns a higher likelihood for levels that has not been sampled for longer, namely,
\begin{equation}
    \PLRStalenessDistribution(\UPOSGScenarioParameter_i|\PLRBuffer,\PLRIterationNumber)=\frac{\PLRIterationNumber-\PLRIterationNumber_{\UPOSGScenarioParameter_i}}{\sum_{j\in\PLRBuffer^{\text{scenario}}} \PLRIterationNumber - \PLRIterationNumber_{\UPOSGScenarioParameter_j}},
    \label{eq:p_staleness}
\end{equation}
where $\PLRIterationNumber$ is the total number of sampling iterations so far, and $\PLRIterationNumber_{\UPOSGScenarioParameter_j}$ is the iteration at which scenario $\UPOSGScenarioParameter_j$ was last sampled. Algorithm \ref{alg:cl4ad_sample} is a pseudocode for how $\CLForAD$ samples new scenarios during training via PLR. First, $\CLForAD$ removes scenarios with ranking lower than $\PLRMaxBufferSize$ in the buffer (Line 1), where $\PLRMaxBufferSize$ is the maximum size of $\PLRBuffer$ for sampling. If the buffer size is smaller than or equal to $\PLRMaxBufferSize$, then no scenario is removed. Then, it determines whether to sample traffic scenarios from the replay buffer. If the replay buffer is empty, or the random replay decision is False, conditioned on the fact that there are still unseen scenarios, then $\CLForAD$ uniformly randomly samples unseen scenarios from the training dataset. Otherwise, it uses the replay distribution $\PLRReplayDistribution$ to sample from the replay buffer $\PLRBuffer$ (lines 2-6). Then, $\CLForAD$ updates the scenarios in the buffer with the newly sampled ones and sets their corresponding last sampling iteration to the current one for staleness computation later on (lines 7-9).  

\subsection{Curriculum Updates in $\CLForAD$}
Algorithm \ref{alg:cl4ad_update} is a pseudocode for how $\CLForAD$ updates the buffer. $\CLForAD$ goes through every world and checks whether an episode has terminated. If so, it computes the utility of that episode based on the rollout that $\CLForAD$ has kept track of. Then it stores this score, and at the next sampling call, it averages the scores of all episodes in that scenario to update the buffer. Finally, the rollout is reset for a new episode to save memory. If the episode continues, $\CLForAD$ updates the rollouts with the latest interactions.

\subsection{Practitioner Guidelines}
\label{app:practitioner_guidelines}

Based on our experimental findings, we offer the following recommendations to practitioners.
 
\paragraph{PLR configuration at scale.}
\begin{enumerate}[nolistsep,noitemsep]
    \item \textbf{Set the replay buffer size equal to the training dataset size.} Prior UED work uses buffers much smaller than the level space; our experiments show that a full-size buffer works consistently as the dataset scales.
    \item \textbf{Use a score temperature significantly higher than prior work.} We use $\beta\in\{2,4\}$, well above the typical range of $\beta\in[0.1,1.0]$. Lower temperatures concentrate sampling on a few top-ranked scenarios, which becomes problematic as the buffer grows. Increasing $\beta$ generally leads to better performance as the dataset scales up. Higher temperature also indirectly mitigates staleness by spreading sampling across more scenarios.
    \item \textbf{The staleness coefficient can remain in the same range as prior work,} $\rho\in\{0.1,0.3\}$ \cite{jiang2021prioritized}, as higher score temperature already alleviates staleness.
\end{enumerate}

\paragraph{Utility function selection.}
\begin{enumerate}[nolistsep,noitemsep]
    \item \textbf{No single utility function dominates across all scales.} However, several patterns emerge from our results (see \cref{sec:experiments_step1_qualitative,sec:correlation}).
    \item \textbf{Among regret-based functions, MaxMC is the most reliable across scales,} as it does not depend on value function accuracy. AMGAE and PVL rely on value estimation and can underperform at smaller scale (case 1). As the dataset scales, their performance improves (case 2), consistent with the increasing correlation between regret-based functions in \cref{fig:correlation}.
    \item \textbf{Success-based functions are effective across all scales} and particularly strong at large scale (case 3), where identifying the learning frontier through success variance is effective.
    \item \textbf{Realism-based functions accelerate training over DR but cannot improve realism,} as they only affect scenario prioritization, not the reward function.
    \item \textbf{Utility functions within the same category tend to prioritize similar scenarios,} especially as the dataset scales, except the realism category (see \cref{fig:correlation}).
\end{enumerate}

\subsection{Computational overhead of $\CLForAD$}
To quantify the overhead of curriculum learning, we measured wall-clock times for all $\CLForAD$ components during training. Curriculum updates (utility computation and buffer score updates) and scenario sampling (PLR sampling and scenario assignment to worlds) occur every $\ScenarioSamplingInterval=2,000,000$ interactions.

Per inter-sampling segment, curriculum update time totals $\sim4.3s$, which is $\sim1\%$ of the segment's evaluation/rollout time ($\sim428.5s$). The PLR sampling operation itself takes $\sim0.66ms$ per call, the remaining $\sim6.15s$ of each sampling block is spent assigning scenarios to worlds and resetting the simulator, which is simulator overhead, not $\CLForAD$ overhead.

Over a full training run of one billion steps ($\sim110$ hours on an A5000, \cref{app:computational_resources}), there are approximately $500$ curriculum sampling steps. The total curriculum update time is approximately $36$ minutes, and the total PLR sampling time is approximately $0.3$ seconds. The overhead of assignment of scenarios ($\sim51$ minutes total) is attributed to the simulator, not to $\CLForAD$. In total, $\CLForAD$ adds less than $1\%$ to the training wall-clock time.

\begin{table}[t!]
    \centering
    \caption{Wall-clock time per inter-sampling segment}
    \begin{tabular}{ccc}
        \hline
        Component  & Time & Percentage\\
        \hline
        Curriculum update (total) & $4.30\pm0.21$s & $1.04\%$ \\
        Evaluation/rollout time & $428.50\pm88.44$s & $100\%$ \\
        \hline
    \end{tabular}
    \label{tab:wallclock_intersampling}
\end{table}

\begin{table}[t!]
    \centering
    \caption{$\CLForAD$ sampling vs. scenario assignment per curriculum sampling call}
    \begin{tabular}{ccc}
        \hline
        Component  & Time & Percentage\\
        \hline
        Curriculum sampling & $0.66\pm0.02$ms & $0.01\%$ \\
        Scenario assignment to worlds & $6.15\pm0.40$s & $99.99\%$ \\
        \hline
    \end{tabular}
    \label{tab:wallclock_intersampling}
\end{table}

\section{Experimental Details}
\label{app:experimental_details}
In this section, we describe the process of hyperparameter selection for our experiments.

\subsection{Simulation Set-up}
Our integration of $\CLForAD$ into $\GPUDRIVE$ follows the simulation set-up in \citet{kazemkhani2025gpudrive}, where the simulator ignores collisions and going off-road, i.e., they do not lead to episode termination; the observation of a vehicle is its bird-eye-view of a radius of 50m; non-vehicle objects are omitted; a goal is considered to be achieved if an agent is in its proximity by 2m; the action consists of two discrete random variables for steering and acceleration inputs, divided into evenly spaced grids, 13 and 7, respectively; maximum number of controlled agents in a scenario is 64; the agents only observe the current time step; and the episode takes 91 timesteps, amounting to 9 seconds, with 1 second of history followed by 8 seconds of future. The public test partition withholds the future trajectories, as they are the prediction targets of the motion forecasting benchmark, leaving 11 logged steps per scenario. Since GPUDrive sets each agent's goal to its final logged position, goals in test scenarios lie roughly one second of driving ahead, and agents are exposed to fewer timesteps in which collisions and off-road events can occur. For more details, we refer the reader to the default PufferLib configuration (see \texttt{environment} section) in the repository published by \citet{kazemkhani2025gpudrive}.

\subsection{Self-play PPO Training}
Self-play RL is an RL scheme for multi-agent settings where each agent samples their actions from a shared, decentralized policy. More formally, this scheme samples the action $\UPOSGAction_{i,t}\sim\Policy_{\PolicyParameter}(\UPOSGObservation_{i,t})$ of agent $i\in\UPOSGAgentSet$ via a policy $\Policy_{\PolicyParameter}$ parameterized by $\PolicyParameter$, e.g., a neural network with learnable parameters $\PolicyParameter$, given the observation $\UPOSGObservation_{i,t}$ of said agent at time $t$. Batched AD simulators $\GIGAFLOW$ and $\GPUDRIVE$ use self-play RL as the strategy to train a single policy that controls all vehicles in a scenario in parallel. Their batched structure empowers parallelization further by concurrently simulating hundreds to thousands of traffic scenarios to accelerate experience collection. Both works employ an on-policy RL algorithm, proximal policy optimization (PPO) \citep{schulman2017proximal}, where policy updates occur once the simultaneous data collection fills an experience buffer. As a result, batched simulation accelerates experience collection via parallelized scenarios, while self-play RL saves compute time and memory by training a single policy. We implement $\CLForAD$ on $\GPUDRIVE$, which samples hundreds of traffic scenarios every couple of million interactions, with initial positions and goals from logged data in WOMD. The default scenario sampling is uniform, where each traffic scenario has equal likelihood.


Table \ref{tab:parameters_ppo} lists the hyperparameters for self-play PPO training in cases 1, 2, and 3, as well as the ablation study. As the ablation study investigates limited compute resources, i.e., the use of fewer worlds and lower batch sizes, we essentially set them according to the hyperparameters in \citet{kazemkhani2025gpudrive}, where the number of worlds $\NumberOfWorlds=50$. In comparison, cases 1, 2, and 3 studies a larger scale in terms of throughput, hence utilize significantly more concurrent worlds and a larger experience buffer. As a result, their hyperparameters come from \citet{cornelisse2025building}, which focuses on a similar scale. The weights for collision/off-road penalties and goal completion rewards also come from \citet{cornelisse2025building}. The experiments are over three independent runs, utilizing seeds $42$, $12$, and $67$. The network architecture also follows the settings in \citet{cornelisse2025building}.

\begin{table}[t!]
    \centering
    \caption{Self-play PPO Hyperparameters}
    \begin{tabular}{lcc}
        \hline
        Parameter  & Case 1,2,3 & Ablation\\
        \hline
        \texttt{total\_timesteps} $\TotalIterations$    & $2,000,000,000$ & $1,000,000,000$ \\
        \texttt{num\_worlds} $\NumberOfWorlds$    & $800$ & $100$ \\
        \texttt{batch\_size} $\PolicyUpdateInterval$    & $524,288$ & $131,072$ \\
        \texttt{minibatch\_size}  & $16,384$ & $8,192$ \\
        \texttt{learning\_rate}  & $0.0003$ & $0.0003$ \\
        \texttt{anneal\_lr}  & \texttt{false} & \texttt{false}  \\
        \texttt{gamma} $\UPOSGDiscount$  & $0.99$ & $0.99$  \\
        \texttt{gae\_gamma} $\GAEDiscount$  & $0.95$ & $0.95$ \\
        \texttt{update\_epochs}  & $2$ & $4$ \\
        \texttt{norm\_adv}  & \texttt{true} & \texttt{true}  \\
        \texttt{clip\_coef}  & $0.2$ & $0.2$ \\
        \texttt{clip\_vloss}  & \texttt{false} & \texttt{false} \\
        \texttt{vf\_clip\_coef}  & $0.2$ & $0.2$ \\
        \texttt{ent\_coef}  & $0.0001$ & $0.0001$ \\
        \texttt{vf\_coef}  & $0.5$ & $0.3$ \\
        \texttt{max\_grad\_norm}  & $0.5$ & $0.5$ \\
        \texttt{target\_kl}  & $\texttt{null}$ & $\texttt{null}$ \\
        \texttt{collision\_weight}  & $-0.75$ & $-0.75$ \\
        \texttt{off\_road\_weight}  & $-0.75$ & $-0.75$ \\
        \texttt{goal\_achieved\_weight}  & $1.0$ & $1.0$ \\
        \hline
    \end{tabular}
    \label{tab:parameters_ppo}
\end{table}

\subsection{Scenario Sampling Details}

Table \ref{tab:parameters_plr} demonstrates the hyperparameters used for the experiments we report in cases 1, 2, 3, and the ablation study. The search space for PLR hyperparameters is as follows: staleness coefficient $\PLRStalenessCoefficient\in\{0.1,0.3\}$ and score temperature $\PLRScoreTemperature\in\{2,4\}$, based mainly on \citet{jiang2021prioritized}. We first conduct a grid search in Case 1, where we train agents using all score functions on three independent runs for one billion interactions. Then we select the pair that yields the highest success rate, the fastest at test-time. \citet{jiang2021replay} suggests a lower temperature; however, our experiments indicate that a higher temperature, especially considering the size of the training dataset, is more performant in large-scale training. Case 3 and the ablation study also utilize these hyperparameters. In case 2, we find that a higher temperature yields better results. We set the replay buffer size to the size of the training dataset, and sample scenarios every $2,000,000$ interactions. For heuristic curricula, once scenarios are ranked using the chosen heuristic, we sample from a score-rank distribution $\PLRScoreDistribution$ with fixed utility and $\PLRScoreTemperature\in\{2,4\}$.

\begin{table}[t!]
    \centering
    \caption{PLR Hyperparameters}
    \begin{tabular}{clccc}
        \hline
        &Utility Function  & $\PLRReplayRate$ & $\PLRScoreTemperature$ & $\PLRStalenessCoefficient$\\
        \hline
        \\
        \multirow{7}{*}{\rotatebox[origin=c]{90}{Case 1}} & $\ACTMAE$ & 0.5 & 2 & 0.3 \\
        &$\AMGAE$ & 0.5 & 4 & 0.3 \\
        &$\GCADE$ & 0.5 & 4 & 0.3 \\
        &$\Learnability$ & 0.5 & 4 & 0.1 \\
        &$\LearnabilityHard$ & 0.5 & 2 & 0.1 \\
        &$\MaxMC$ & 0.5 & 2 & 0.3 \\
        &$\PVL$ & 0.5 & 4 & 0.3 \\
        \\
        \hline
        \\
        \multirow{3}{*}{\rotatebox[origin=c]{90}{Case 2}} & $\ACTMAE$ & 0.5 & 4 & 0.3 \\
        &$\Learnability$ & 0.5 & 4 & 0.1 \\
        &$\MaxMC$ & 0.5 & 4 & 0.3 \\
        \\
        \hline
        \\
        \multirow{3}{*}{\rotatebox[origin=c]{90}{Case 3}} & $\ACTMAE$ & 0.5 & 4 & 0.3 \\
        &$\Learnability$ & 0.5 & 4 & 0.1 \\
        &$\MaxMC$ & 0.5 & 2 & 0.3 \\
        \\
        \hline
        \\
        \multirow{3}{*}{\rotatebox[origin=c]{90}{Ablation}} & $\ACTMAE$ & 0.5 & 2 & 0.3 \\
        &$\Learnability$ & 0.5 & 4 & 0.1 \\
        &$\MaxMC$ & 0.5 & 2 & 0.3 \\
        \\
        \hline
    \end{tabular}
    \label{tab:parameters_plr}
\end{table}


\section{Computational Resources}
\label{app:computational_resources}
We run our experiments in cases 1, 2, and 3 on an NVIDIA H200, which has 141 GB of GPU memory. One training run, which amounts to 2 billion steps and approximately 3,800 policy updates, takes around 60 hours. For the ablation study, we train agents on NVIDIA RTX A5000, which has a GPU memory of 24GB, for a billion interactions, which takes over 110 hours.

\section{Detailed Results}
\label{app:detailed_results}
\subsection{WOSAC Evaluation}
\label{app:wosac}

We evaluate the final trained policies using the Waymo Open Sim Agents Challenge (WOSAC) metrics \citep{montali2023waymo} in two settings: (1) all controlled agents via self-play, and (2) ego-only with other agents replaying logged trajectories. \cref{tab:wosac_selfplay} and \cref{tab:wosac_ego} report the results for case 1 (150 test scenarios, 3 seeds).

\begin{table}[h]
\centering
\caption{WOSAC metrics, self-play evaluation with final trained policies (case 1: 150 test scenarios).}
\label{tab:wosac_selfplay}
\resizebox{\textwidth}{!}{
\begin{tabular}{lcccccccccccc}
\toprule
Method & Realism $\uparrow$ & minADE $\downarrow$ & Lin Spd $\uparrow$ & Lin Acc $\uparrow$ & Ang Spd $\uparrow$ & Ang Acc $\uparrow$ & Dist Obj $\uparrow$ & Collis $\uparrow$ & TTC $\uparrow$ & Dist Edge $\uparrow$ & Offroad $\uparrow$ \\
\midrule
Oracle & 0.832 & 0.00 & 0.493 & 0.448 & 0.578 & 0.694 & 0.430 & 0.999 & 0.900 & 0.772 & 0.999 \\
DR & 0.689 & 10.28 & 0.161 & 0.246 & 0.510 & 0.658 & 0.163 & 0.855 & 0.856 & 0.703 & 0.901 \\
PLR+MaxMC & 0.657 & 9.20 & 0.165 & 0.252 & 0.514 & 0.662 & 0.158 & 0.830 & 0.857 & 0.701 & 0.804 \\
PLR+GC-ADE & 0.652 & 9.58 & 0.158 & 0.229 & 0.509 & 0.660 & 0.152 & 0.840 & 0.856 & 0.691 & 0.783 \\
PLR+Learn-Hard & 0.649 & 9.76 & 0.158 & 0.230 & 0.508 & 0.660 & 0.150 & 0.837 & 0.856 & 0.694 & 0.778 \\
PLR+AMGAE & 0.644 & 9.94 & 0.159 & 0.244 & 0.508 & 0.659 & 0.156 & 0.849 & 0.854 & 0.674 & 0.747 \\
PLR+PVL & 0.640 & 9.71 & 0.157 & 0.231 & 0.510 & 0.660 & 0.152 & 0.858 & 0.857 & 0.670 & 0.724 \\
PLR+Act-MAE & 0.631 & 9.40 & 0.158 & 0.212 & 0.512 & 0.664 & 0.147 & 0.824 & 0.856 & 0.689 & 0.724 \\
PLR+Learn & 0.628 & 10.10 & 0.153 & 0.217 & 0.507 & 0.663 & 0.146 & 0.863 & 0.856 & 0.648 & 0.685 \\
\bottomrule
\end{tabular}
}
\end{table}

\begin{table}[h]
\centering
\caption{WOSAC metrics, ego-only evaluation with final trained policies (case 1: 150 test scenarios).}
\label{tab:wosac_ego}
\resizebox{\textwidth}{!}{
\begin{tabular}{lcccccccccccc}
\toprule
Method & Realism $\uparrow$ & minADE $\downarrow$ & Lin Spd $\uparrow$ & Lin Acc $\uparrow$ & Ang Spd $\uparrow$ & Ang Acc $\uparrow$ & Dist Obj $\uparrow$ & Collis $\uparrow$ & TTC $\uparrow$ & Dist Edge $\uparrow$ & Offroad $\uparrow$ \\
\midrule
Oracle & 0.871 & 0.00 & 0.582 & 0.694 & 0.721 & 0.917 & 0.482 & 0.999 & 0.918 & 0.854 & 0.999 \\
DR & 0.669 & 9.14 & 0.279 & 0.402 & 0.617 & 0.864 & 0.144 & 0.769 & 0.843 & 0.724 & 0.804 \\
PLR+MaxMC & 0.624 & 8.25 & 0.277 & 0.405 & 0.620 & 0.862 & 0.136 & 0.765 & 0.842 & 0.674 & 0.649 \\
PLR+PVL & 0.608 & 8.47 & 0.274 & 0.386 & 0.615 & 0.866 & 0.125 & 0.766 & 0.844 & 0.659 & 0.597 \\
PLR+GC-ADE & 0.603 & 8.58 & 0.273 & 0.385 & 0.612 & 0.863 & 0.132 & 0.753 & 0.842 & 0.661 & 0.591 \\
PLR+AMGAE & 0.595 & 8.72 & 0.273 & 0.388 & 0.612 & 0.864 & 0.128 & 0.749 & 0.840 & 0.637 & 0.572 \\
PLR+Learn-Hard & 0.586 & 8.45 & 0.274 & 0.385 & 0.613 & 0.868 & 0.126 & 0.757 & 0.842 & 0.636 & 0.529 \\
PLR+Act-MAE & 0.572 & 8.35 & 0.271 & 0.369 & 0.615 & 0.864 & 0.124 & 0.761 & 0.843 & 0.625 & 0.478 \\
PLR+Learn & 0.556 & 9.00 & 0.267 & 0.373 & 0.611 & 0.861 & 0.120 & 0.768 & 0.841 & 0.565 & 0.424 \\
\bottomrule
\end{tabular}
}
\end{table}

Kinematics and interaction scores are comparable across all methods, indicating that the curriculum choice does not meaningfully affect these categories. The main difference is in map-based metrics (Dist Edge, Offroad), where DR scores higher because WOSAC measures distributional similarity to logged behavior, and uniform sampling stays closer to the data distribution. PLR variants achieve lower displacement error, consistent with more efficient goal reaching, but diverge from human driving patterns in map-based metrics. This gap widens in the ego-only setting. All methods remain far from the oracle, with kinematics representing the largest gap. These results support the analysis in \cref{sec:realism_vs_optimality}: the curriculum improves task performance but cannot improve realism without changes to the reward function.

\subsection{Quantitative Results}

\begin{figure}[t]
    \centering
    \includegraphics[width=\textwidth,trim={0 240pt 0pt 0pt},clip]{figures/performance_shadow_plots_train_experiments_step1.pdf}
    \begin{subfigure}[b]{\textwidth}
        \centering
        \includegraphics[width=\textwidth,trim={0pt 0pt 0pt 100pt},clip]{figures/performance_shadow_plots_train_experiments_step1.pdf}
        \includegraphics[width=.7\textwidth,trim={0 0pt 0pt 90pt},clip]{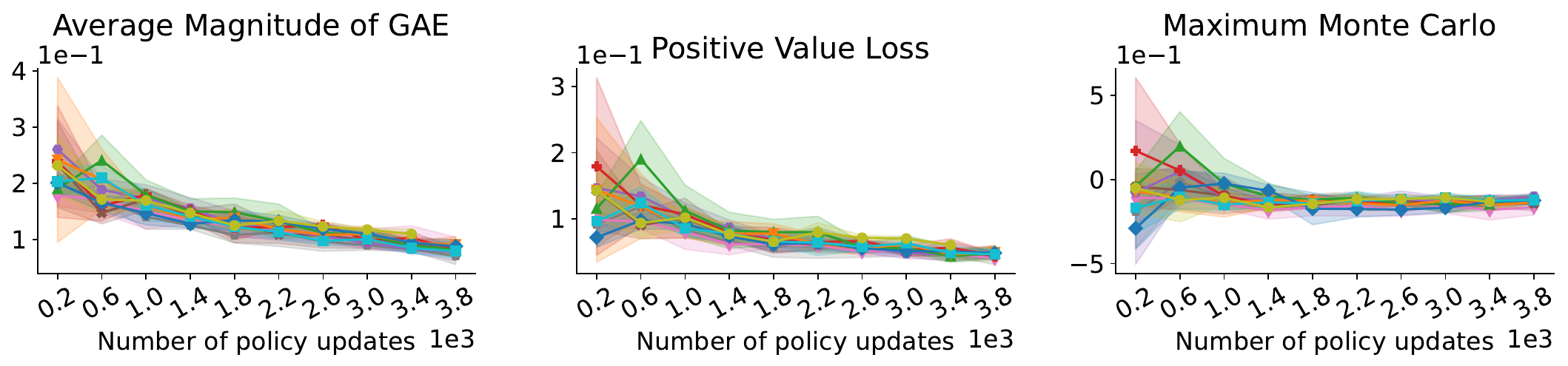}
            \includegraphics[width=0.29\textwidth,trim={0 0pt 0pt 165pt},clip]{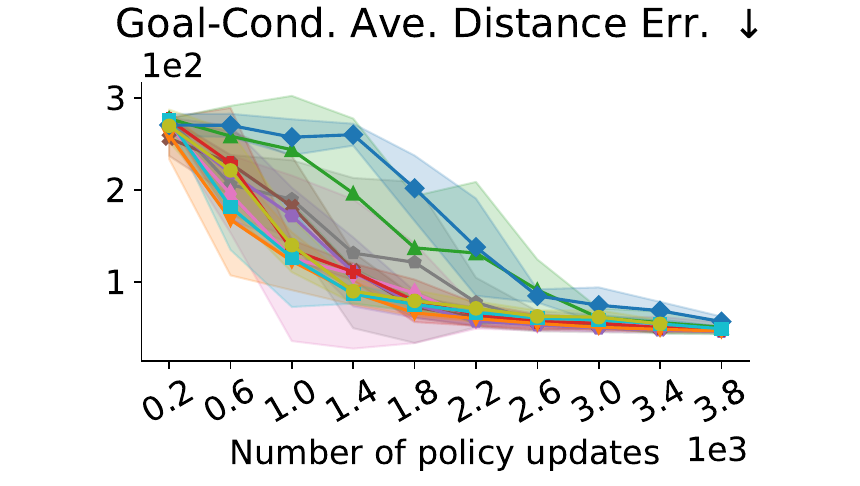}
            \caption{Evaluation on training partition}
    \end{subfigure}
    \begin{subfigure}[b]{\textwidth}
        \centering
        \includegraphics[width=\textwidth,trim={0pt 0pt 0pt 90pt},clip]{figures/performance_shadow_plots_test_experiments_step1.pdf}
        \includegraphics[width=0.7\textwidth,trim={0 0pt 0pt 90pt},clip]{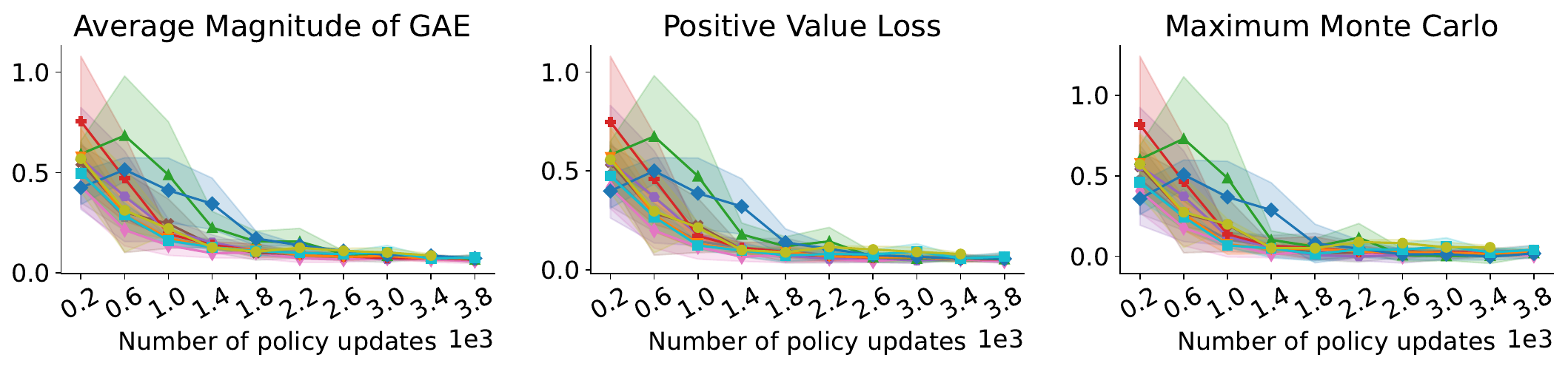}
        \includegraphics[width=0.6\textwidth,trim={0 0pt 0pt 110pt},clip]{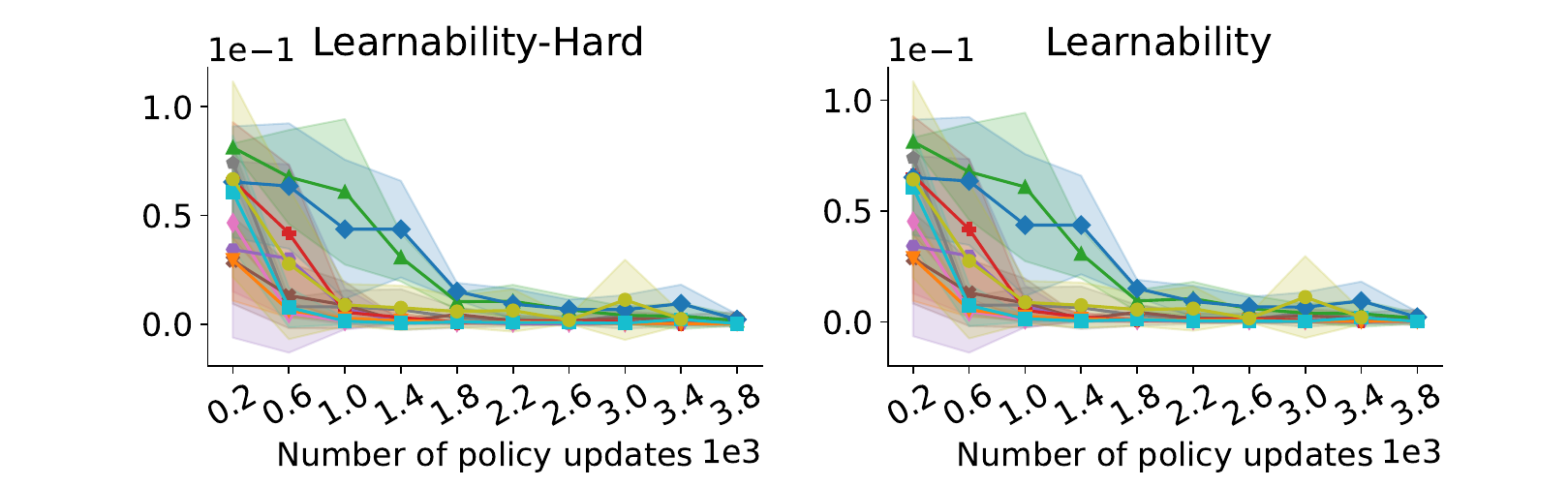}
            \caption{Evaluation on test partition}
    \end{subfigure}
    \vspace*{-5mm}
    \caption{Case 1: Regret ($\AMGAE$, $\PVL$, $\MaxMC$), realism ($\GCADE$), and learnability ($\Learnability$, $\LearnabilityHard$), progression during training with 1000 scenarios from WOMD: We evaluate in \textbf{(a)} training partition, and \textbf{(b)} 150 test scenarios. Bold markers indicate the mean, whereas the shaded area covers one standard deviation around it across three independent training runs.}
    \label{fig:experiment_step1_detailed_results}
\end{figure}

\begin{figure}[t]
    \centering
    \includegraphics[width=\textwidth,trim={0 240pt 0pt 0pt},clip]{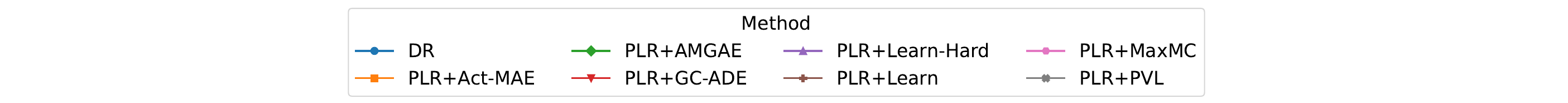}
    \begin{subfigure}[b]{\textwidth}
        \centering
        \includegraphics[width=\textwidth,trim={0 0pt 0pt 90pt},clip]{figures/performance_shadow_plots_train_experiments_step2.pdf}
        \includegraphics[width=0.75\textwidth,trim={0 0pt 0pt 90pt},clip]{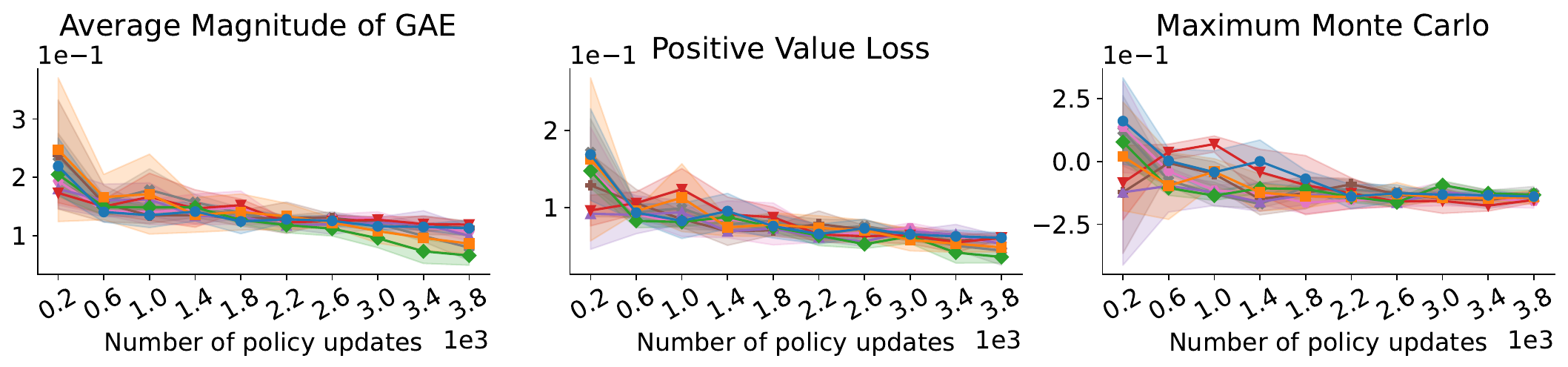}
        \includegraphics[width=0.24\textwidth,trim={30 0pt 30pt 140pt},clip]{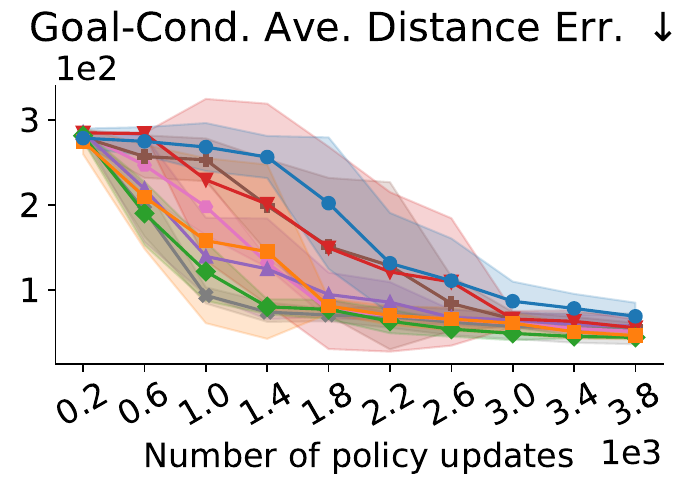}
        \includegraphics[width=0.55\textwidth,trim={0 0pt 0pt 90pt},clip]{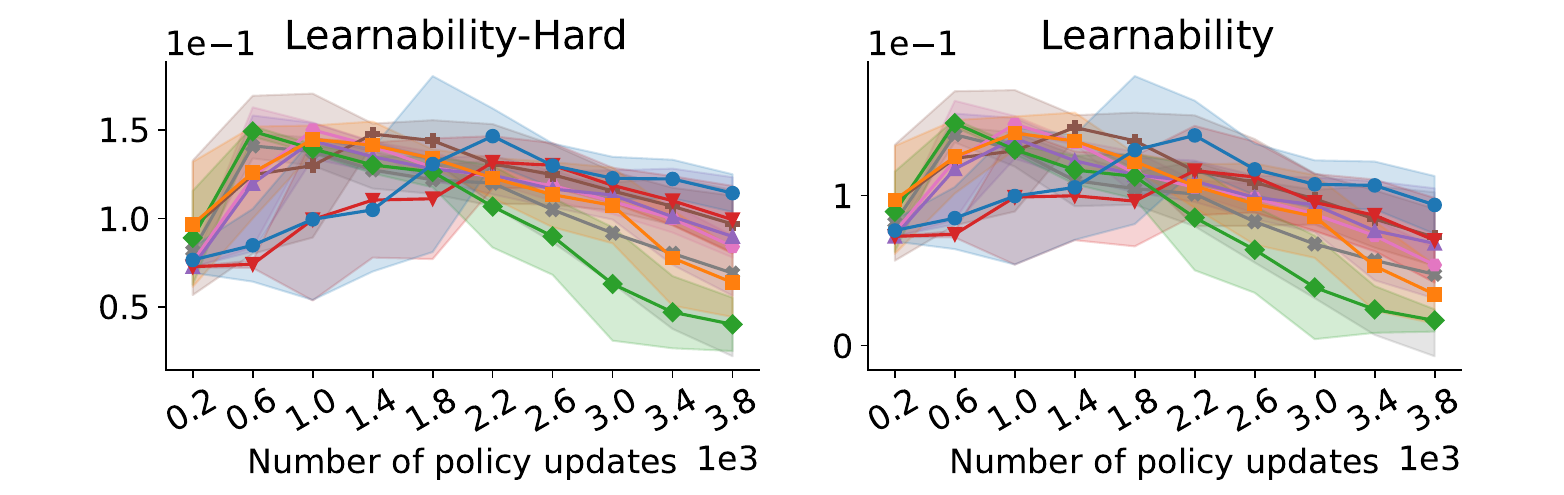}
        \caption{Evaluation on training partition}
    \end{subfigure}
    \begin{subfigure}[b]{\textwidth}
        \centering
        \includegraphics[width=\textwidth,trim={0 0pt 0pt 90pt},clip]{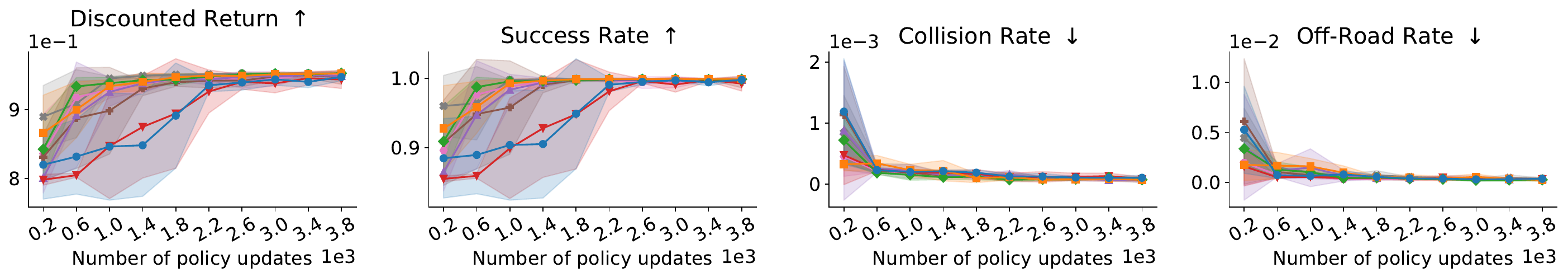}
        \includegraphics[width=0.75\textwidth,trim={0 0pt 0pt 90pt},clip]{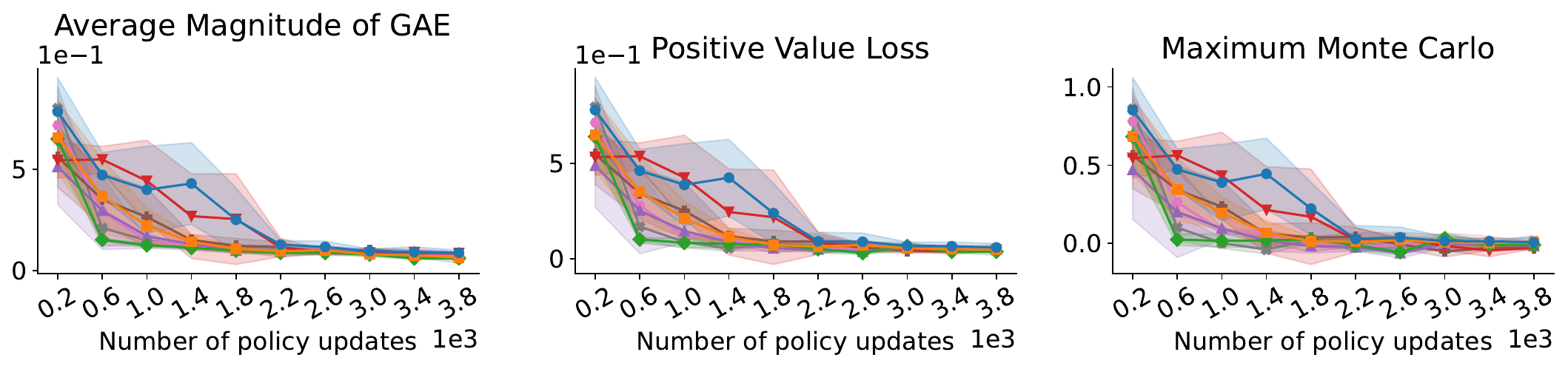}
        \includegraphics[width=0.24\textwidth,trim={30 0pt 30pt 140pt},clip]{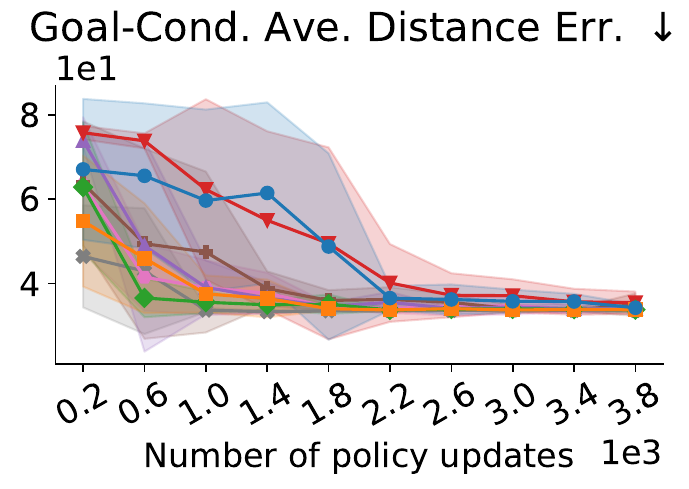}
        \includegraphics[width=0.55\textwidth,trim={0 0pt 0pt 90pt},clip]{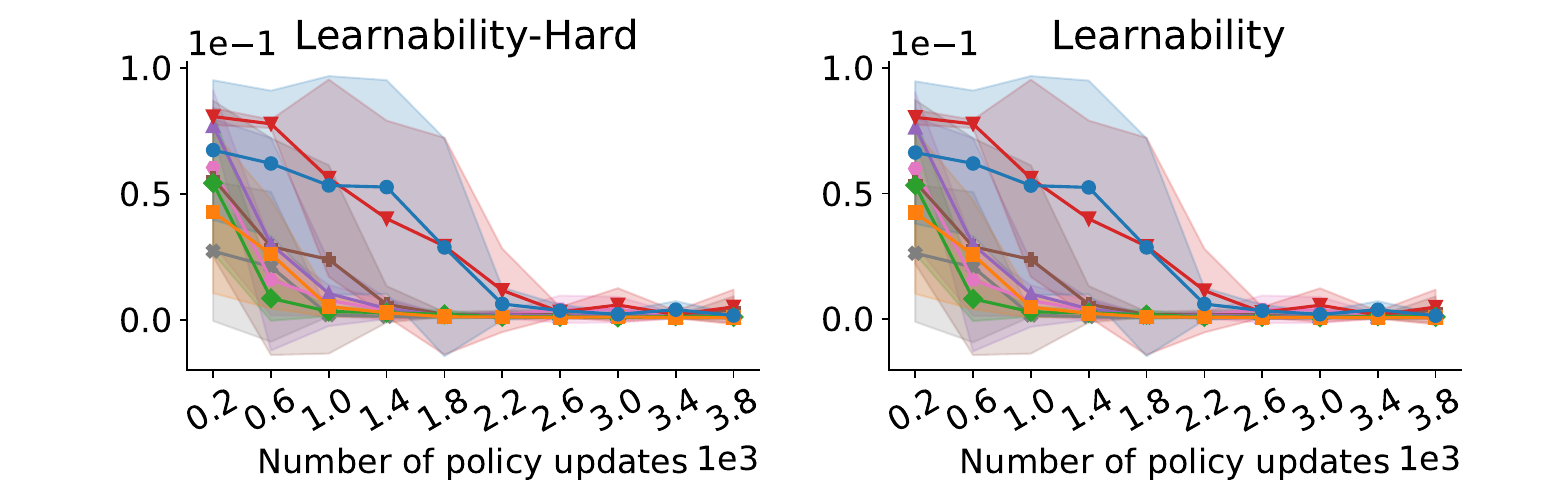}
        \caption{Evaluation on test partition}
    \end{subfigure}
    \vspace*{-5mm}
    \caption{Case 2: Performance, Regret ($\AMGAE$, $\PVL$, $\MaxMC$), realism ($\GCADE$), and learnability ($\Learnability$, $\LearnabilityHard$), progression during training with 10,000 scenarios from WOMD: We evaluate in \textbf{(a)} training partition, and \textbf{(b)} 10,000 test scenarios. Bold markers indicate the mean, whereas the shaded area covers one standard deviation around it across three training runs.}
    \label{fig:experiment_step2_detailed_results}
\end{figure}

\begin{figure}[t]
    \centering
    \includegraphics[width=\textwidth,trim={0 240pt 0pt 0pt},clip]{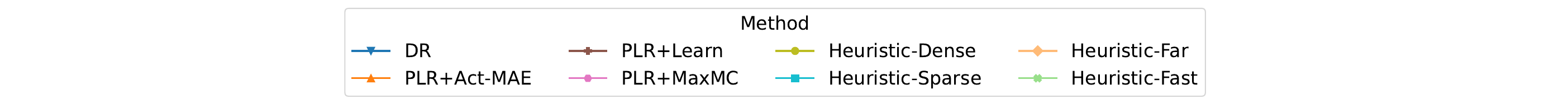}
    \includegraphics[width=\textwidth,trim={0 0pt 0pt 90pt},clip]{figures/performance_shadow_plots_test_experiments_step3.pdf}
    \includegraphics[width=0.72\textwidth,trim={0 0pt 0pt 90pt},clip]{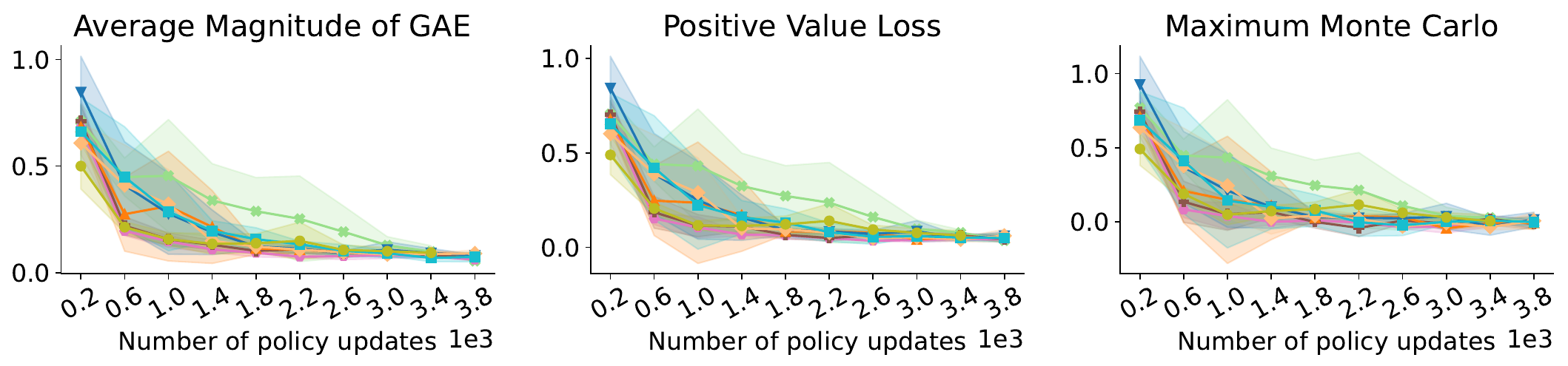}
    \includegraphics[width=0.27\textwidth,trim={0 0pt 0pt 140pt},clip]{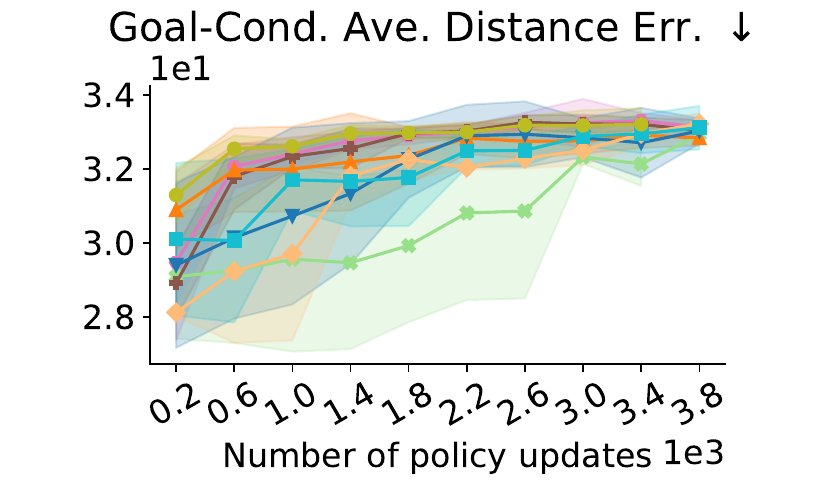}
    \includegraphics[width=0.55\textwidth,trim={0 0pt 0pt 90pt},clip]{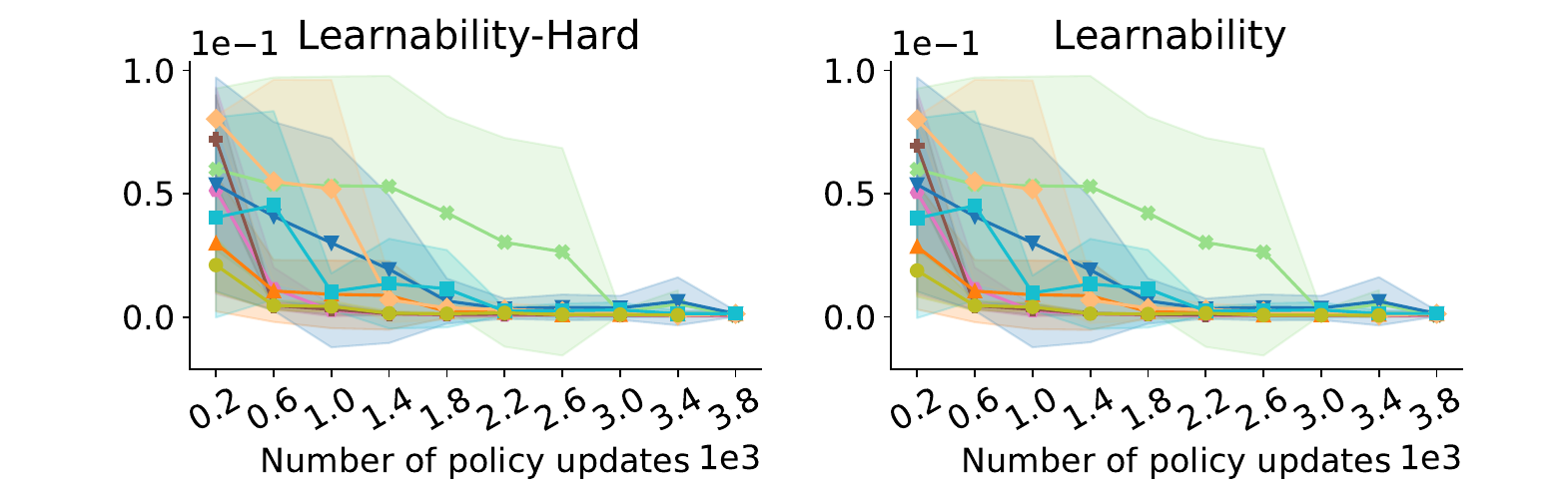}
    \vspace*{-3mm}
    \caption{Case 3: Regret ($\AMGAE$, $\PVL$, $\MaxMC$), realism ($\GCADE$), and learnability ($\Learnability$, $\LearnabilityHard$), progression during training with 80,000 scenarios from WOMD: We evaluate in 10,000 test scenarios. Bold markers indicate the mean, whereas the shaded area covers one standard deviation around it across two independent training runs.}
    \label{fig:experiment_step3_detailed_results}
\end{figure}

\begin{figure}[t]
    \centering
    \includegraphics[width=\textwidth,trim={0 240pt 0pt 0pt},clip]{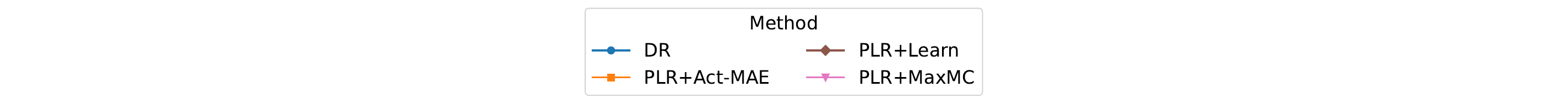}
    \begin{subfigure}[b]{\textwidth}
        \centering
        \includegraphics[width=\textwidth,trim={0 0pt 0pt 90pt},clip]{figures/performance_shadow_plots_train_experiments_computeab.pdf}
        \includegraphics[width=0.74\textwidth,trim={0 0pt 0pt 90pt},clip]{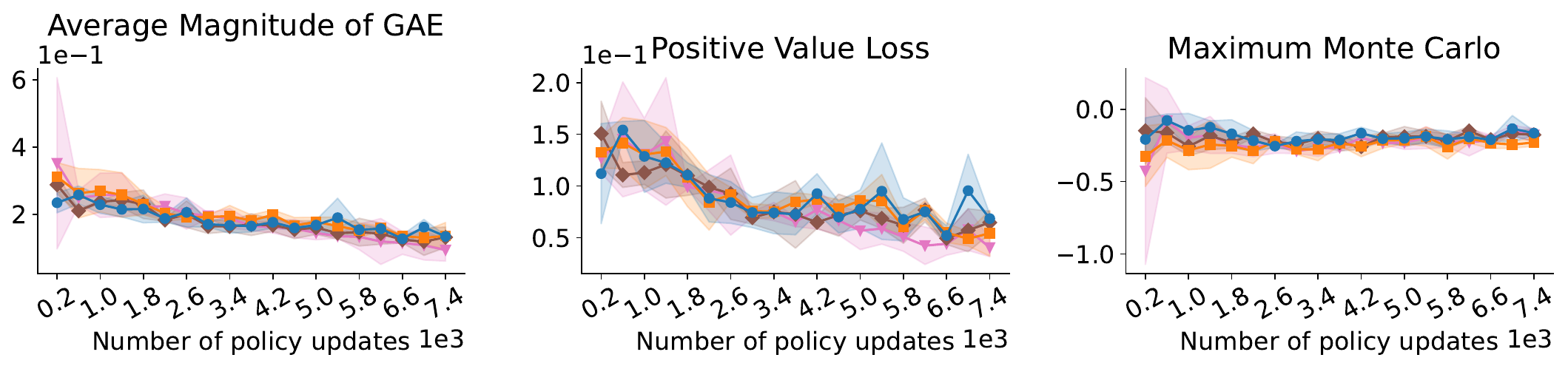}
        \includegraphics[width=0.25\textwidth,trim={0 0pt 0pt 90pt},clip]{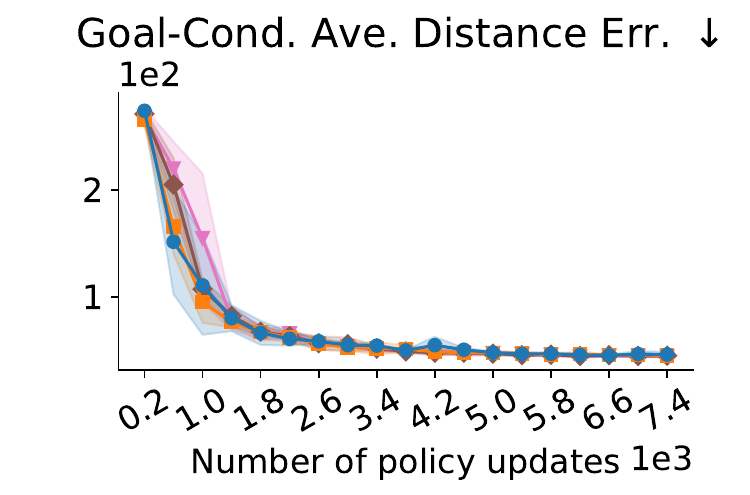}
        \includegraphics[width=0.5\textwidth,trim={0 0pt 0pt 90pt},clip]{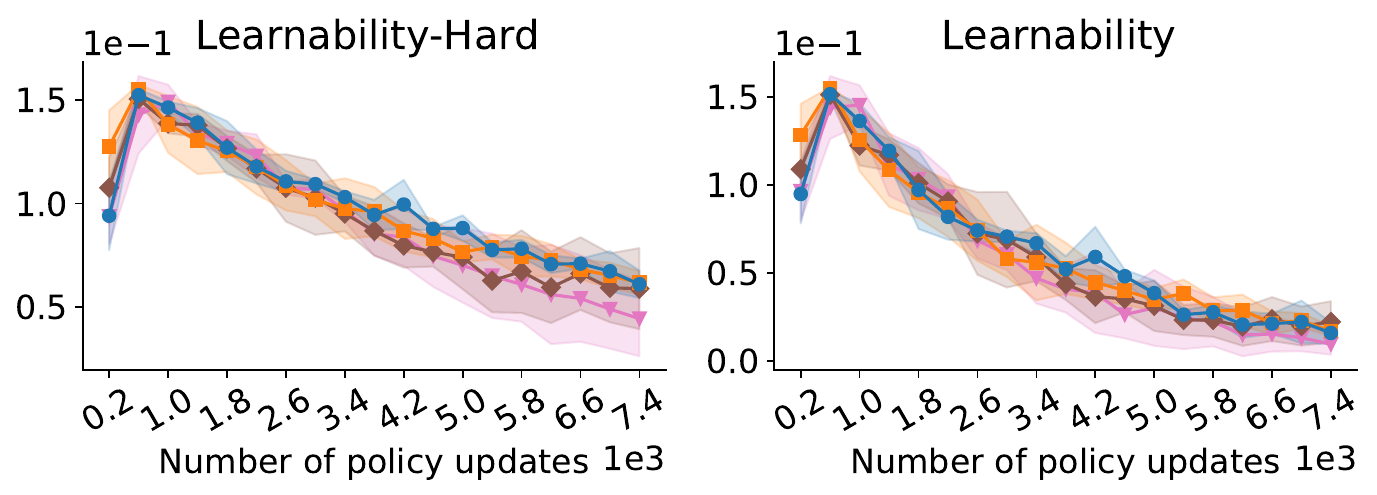}
        \caption{Evaluation on training partition}
    \end{subfigure}
    \begin{subfigure}[b]{\textwidth}
        \centering
        \includegraphics[width=\textwidth,trim={0 0pt 0pt 90pt},clip]{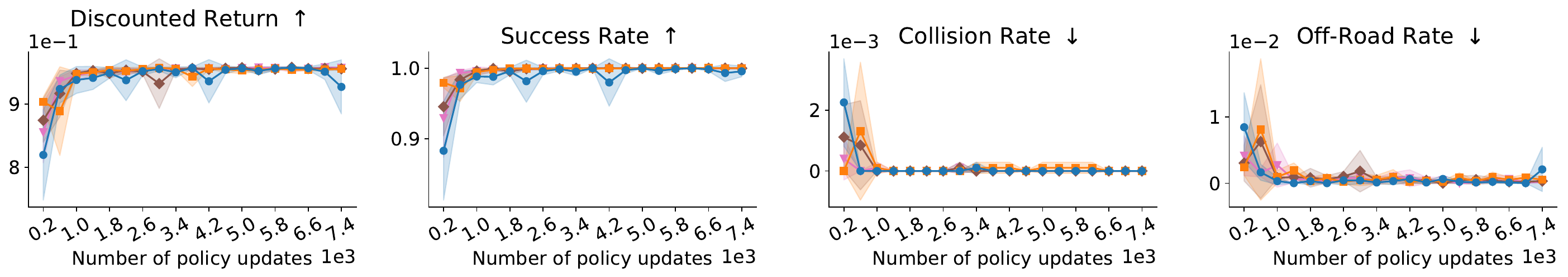}
        \includegraphics[width=0.74\textwidth,trim={0 0pt 0pt 90pt},clip]{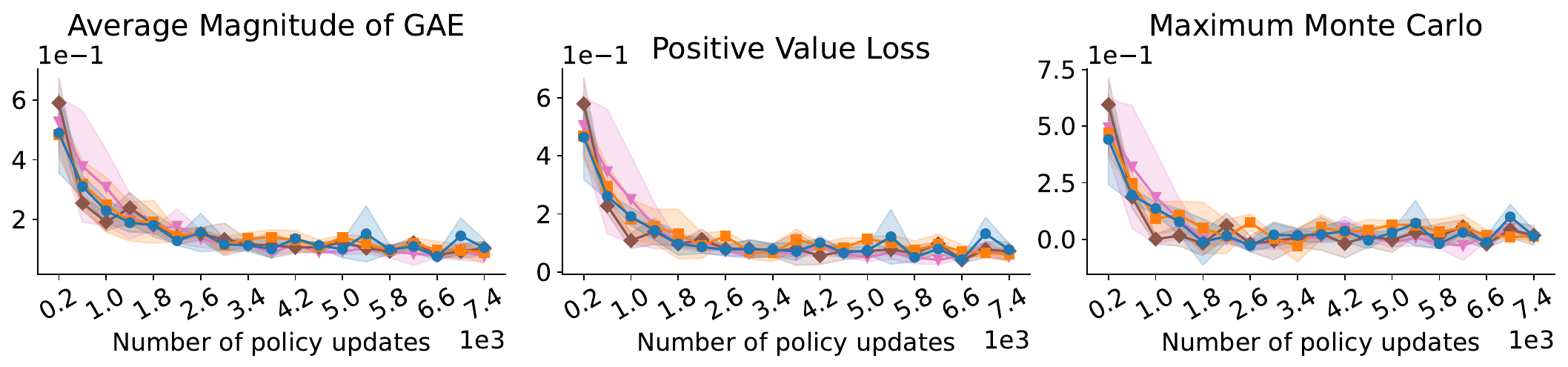}
        \includegraphics[width=0.25\textwidth,trim={0 0pt 0pt 90pt},clip]{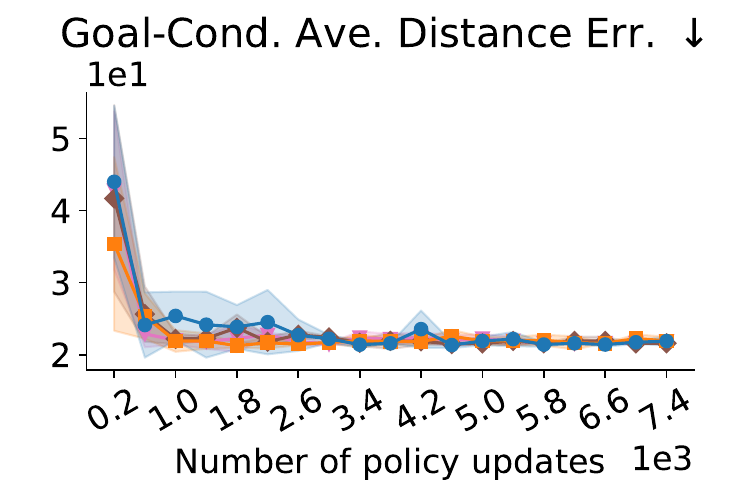}
        \includegraphics[width=0.5\textwidth,trim={0 0pt 0pt 90pt},clip]{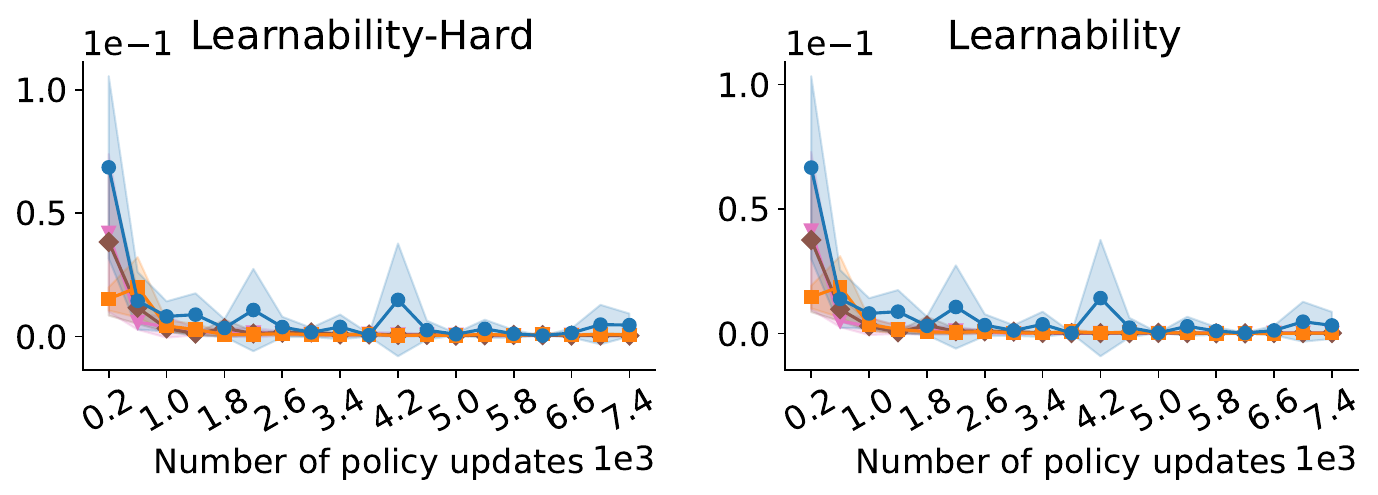}
        \caption{Evaluation on test partition}
    \end{subfigure}
    \vspace*{-5mm}
    \caption{Ablation: Performance, regret ($\AMGAE$, $\PVL$, $\MaxMC$), realism ($\GCADE$), and learnability ($\Learnability$, $\LearnabilityHard$), progression during training for our ablation study on compute resources: We evaluate in \textbf{(a)} training partition, and \textbf{(b)} 150 test scenarios. Bold markers indicate the mean, whereas the shaded area covers one standard deviation around it across three training runs.}
    \label{fig:ablation_detailed_results}
\end{figure}

Figures \ref{fig:experiment_step1_detailed_results}, \ref{fig:experiment_step2_detailed_results}, \ref{fig:experiment_step3_detailed_results}, and \ref{fig:ablation_detailed_results} demonstrate the progression of trained agents in cases 1, 2, and 3, as well as the compute ablation, respectively. These figures provide details on the progression of performance, regret, realism, and learnability when agents are evaluated in the training and test partitions of their respective experiments. Regret, learnability, and realism in the training partition highlight how automated curricula impact training. In most cases, we observe that PLR variants are significantly faster than DR at achieving low utility scores in these metrics, indicating that they obtain more performant and realistic policies more quickly. The performance progression, when evaluated on the training partition, leads to a similar observation as well. Progression in test scenarios demonstrates the generalization capabilities of these trained agents, as these scenarios were not encountered during training. Overall, we observe that PLR variants are again quickly becoming more capable at generalization or becoming robust and reliable faster than agents trained via DR.

\subsection{Qualitative Results}

\begin{figure}[t]
    \centering
    \includegraphics[width=.8\textwidth,trim={0 0pt 0pt 0pt},clip]{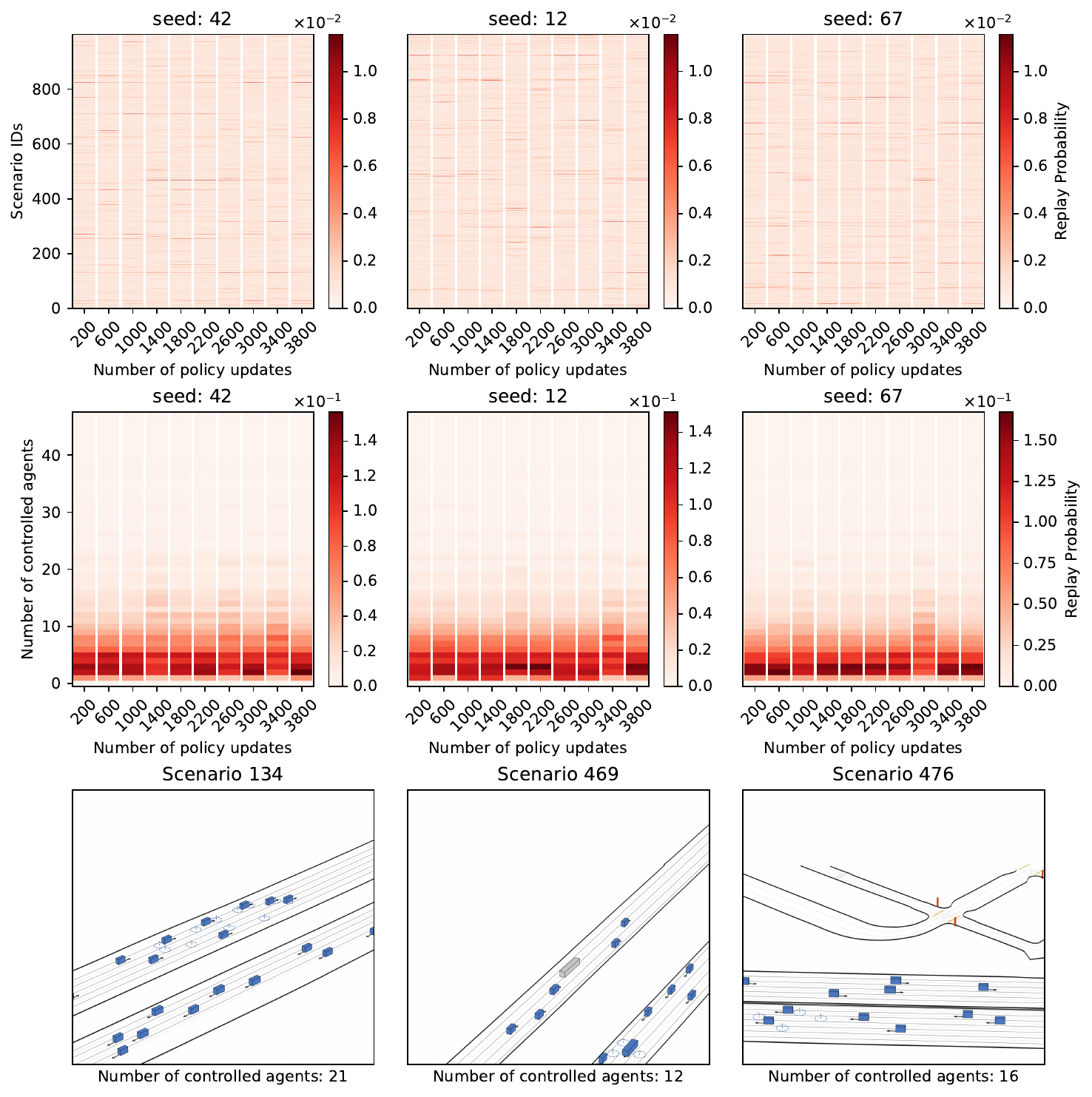}
    \caption{$\PLRReplayDistribution$ progression of PLR combined with $\ACTMAE$ in mini WOMD: We illustrate \textbf{(top)} the evolution of $\PLRReplayDistribution$, where darker line segments indicate scenarios with higher replay likelihood, \textbf{(middle)} a version of replay distribution under categorization with respect to the number of controlled agents in scenarios, and \textbf{(bottom)} we exemplify three scenarios that appear frequently.}
    \label{fig:experiment_step1_replay_actmae}
\end{figure}

\begin{figure}[t]
    \centering
    \includegraphics[width=.8\textwidth,trim={0 0pt 0pt 0pt},clip]{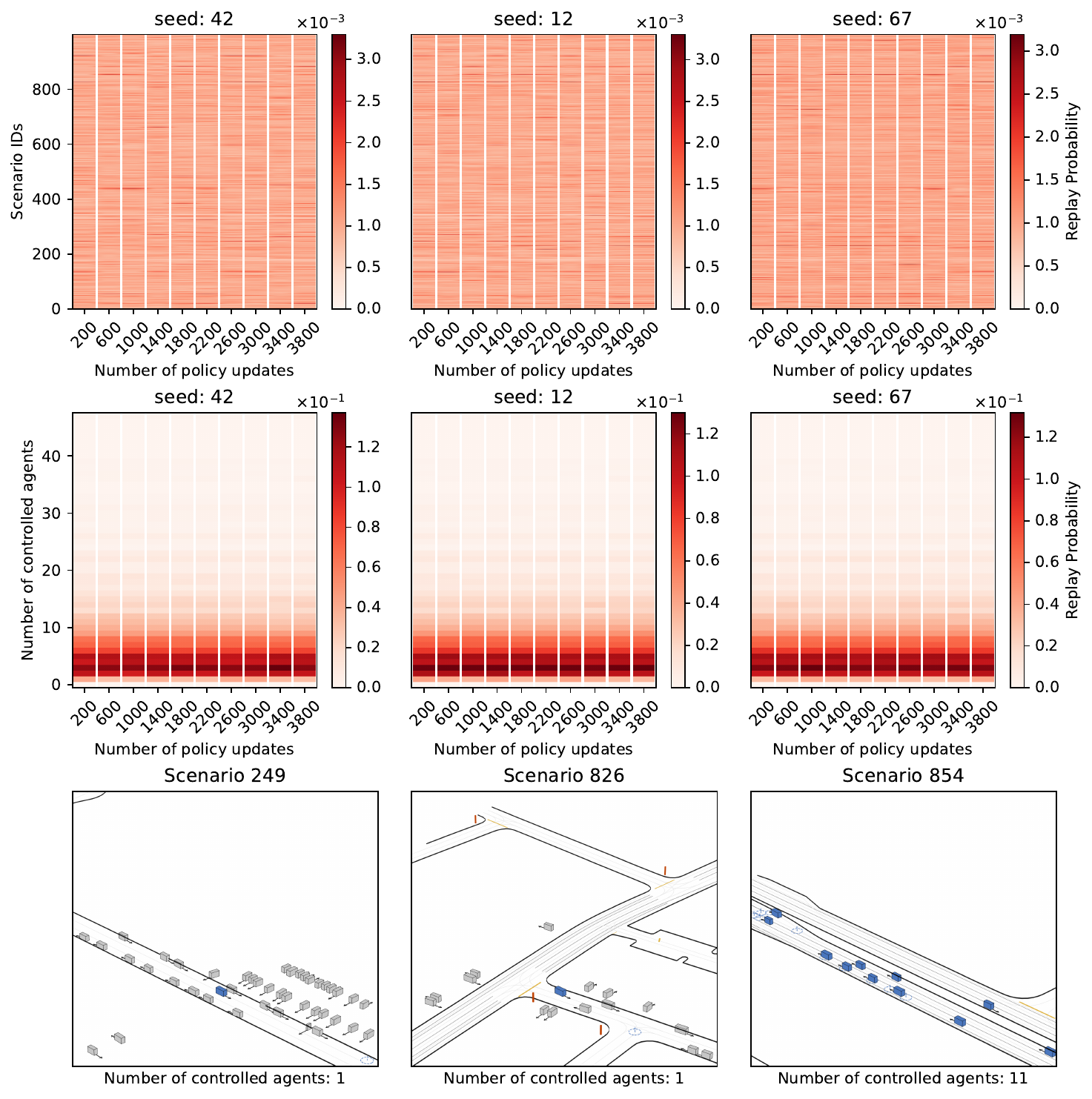}
    \caption{$\PLRReplayDistribution$ progression of PLR combined with $\AMGAE$ in mini WOMD: We illustrate \textbf{(top)} the evolution of $\PLRReplayDistribution$, where darker line segments indicate scenarios with higher replay likelihood, \textbf{(middle)} a version of replay distribution under categorization with respect to the number of controlled agents in scenarios, and \textbf{(bottom)} we exemplify three scenarios that appear frequently.}
    \label{fig:experiment_step1_replay_amgae}
\end{figure}

\begin{figure}[t]
    \centering
    \includegraphics[width=.8\textwidth,trim={0 0pt 0pt 0pt},clip]{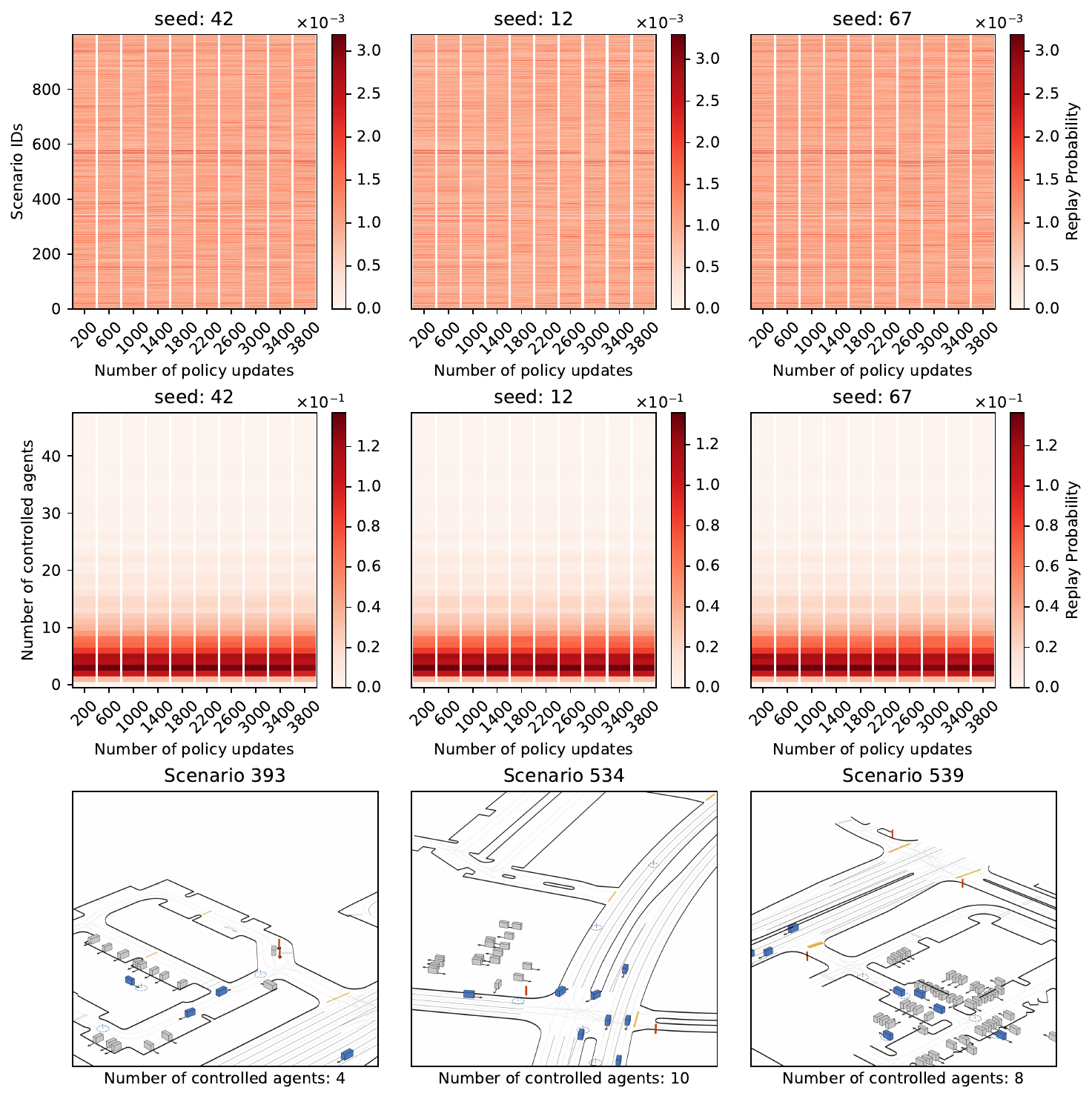}
    \caption{$\PLRReplayDistribution$ progression of PLR combined with $\GCADE$ in mini WOMD: We illustrate \textbf{(top)} the evolution of $\PLRReplayDistribution$, where darker line segments indicate scenarios with higher replay likelihood, \textbf{(middle)} a version of replay distribution under categorization with respect to the number of controlled agents in scenarios, and \textbf{(bottom)} we exemplify three scenarios that appear frequently.}
    \label{fig:experiment_step1_replay_gcade}
\end{figure}

\begin{figure}[t]
    \centering
    \includegraphics[width=.8\textwidth,trim={0 0pt 0pt 0pt},clip]{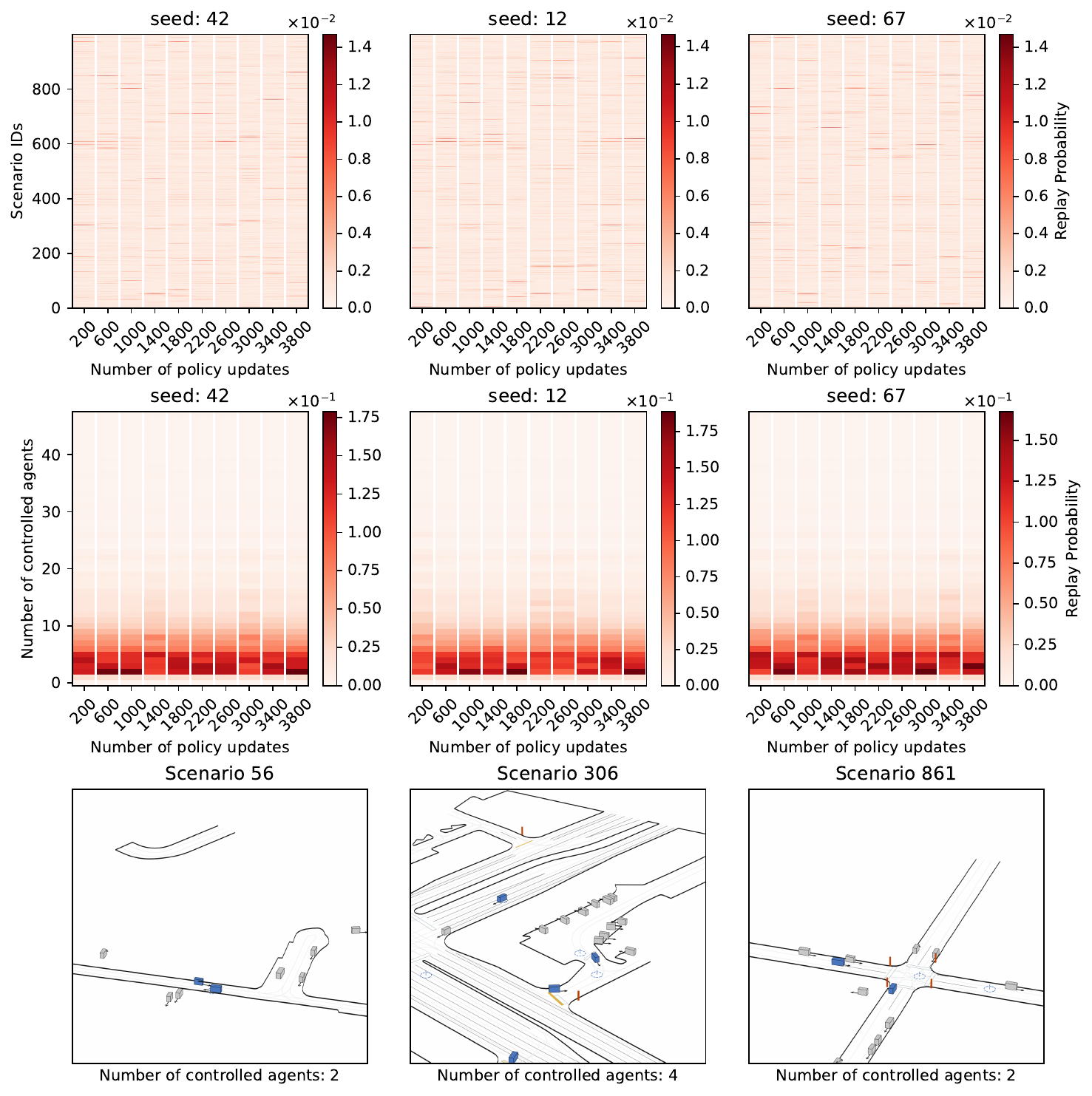}
    \caption{$\PLRReplayDistribution$ progression of PLR combined with $\LearnabilityHard$ in mini WOMD: We illustrate \textbf{(top)} the evolution of $\PLRReplayDistribution$, where darker line segments indicate scenarios with higher replay likelihood, \textbf{(middle)} a version of replay distribution under categorization with respect to the number of controlled agents in scenarios, and \textbf{(bottom)} we exemplify three scenarios that appear frequently.}
    \label{fig:experiment_step1_replay_learn}
\end{figure}

\begin{figure}[t]
    \centering
    \includegraphics[width=.8\textwidth,trim={0 0pt 0pt 0pt},clip]{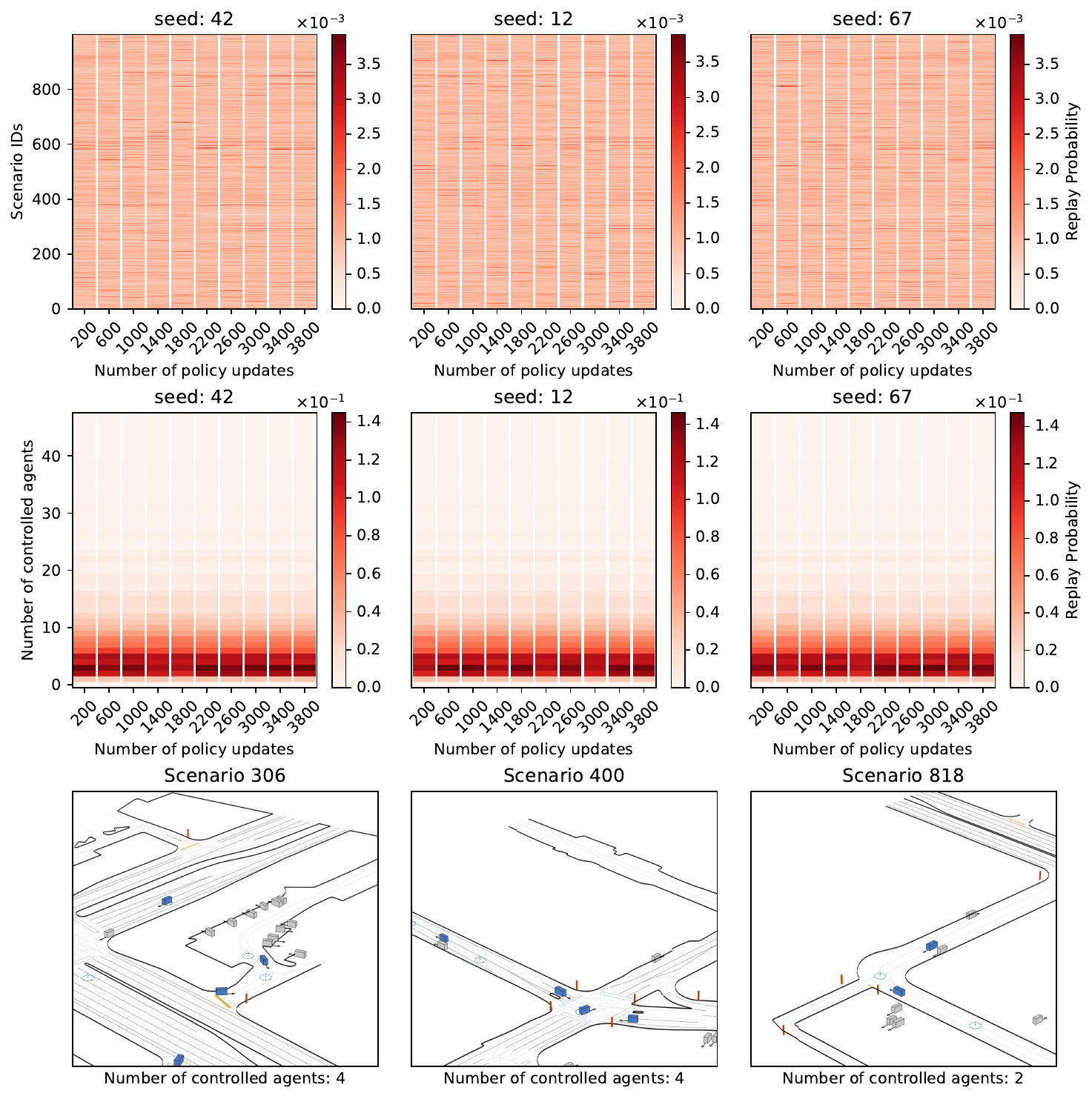}
    \caption{$\PLRReplayDistribution$ progression of PLR combined with $\Learnability$ in mini WOMD: We illustrate \textbf{(top)} the evolution of $\PLRReplayDistribution$, where darker line segments indicate scenarios with higher replay likelihood, \textbf{(middle)} a version of replay distribution under categorization with respect to the number of controlled agents in scenarios, and \textbf{(bottom)} we exemplify three scenarios that appear frequently.}
    \label{fig:experiment_step1_replay_learngoal}
\end{figure}

\begin{figure}[t]
    \centering
    \includegraphics[width=.8\textwidth,trim={0 0pt 0pt 0pt},clip]{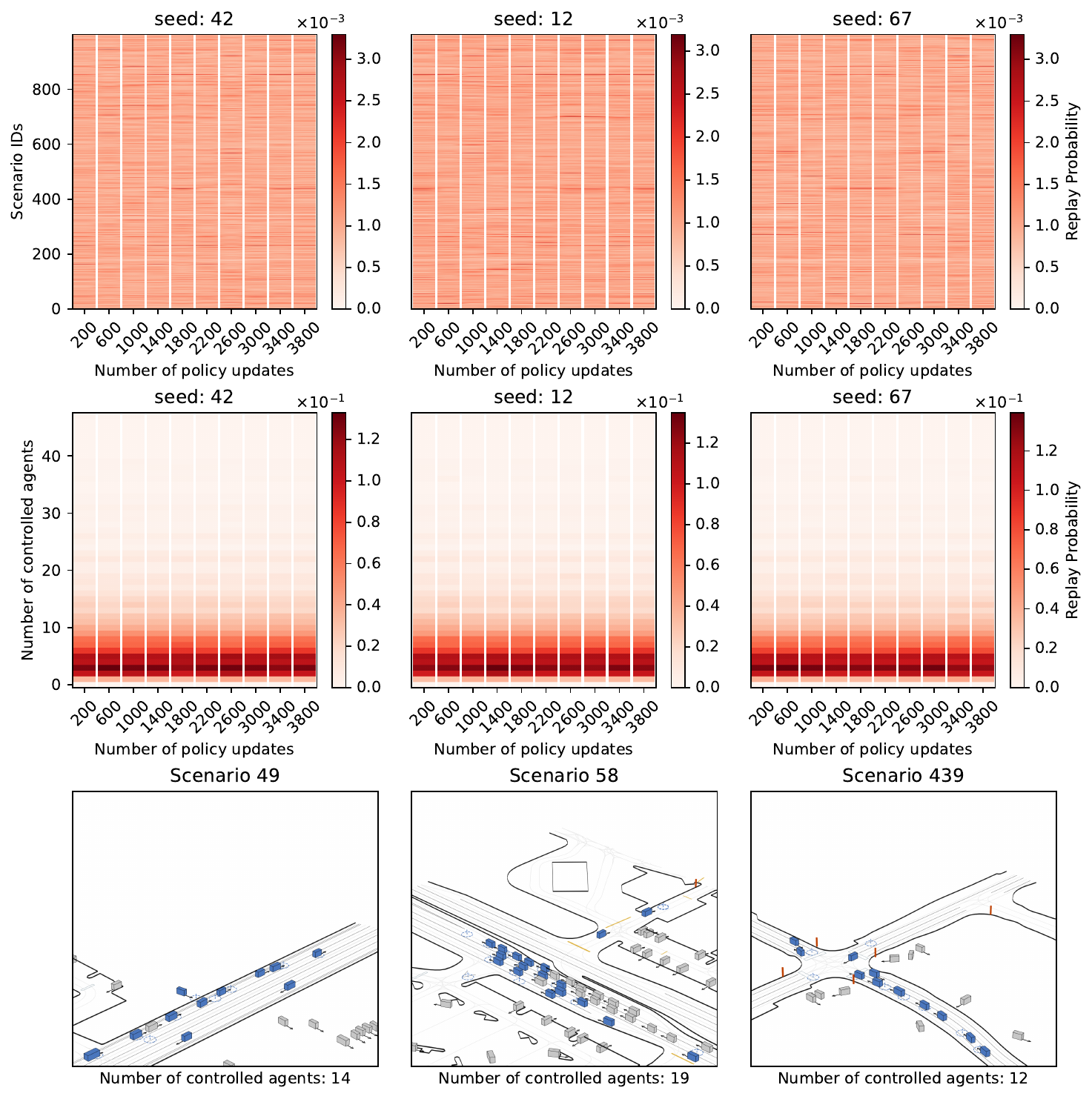}
    \caption{$\PLRReplayDistribution$ progression of PLR combined with $\PVL$ in mini WOMD: We illustrate \textbf{(top)} the evolution of $\PLRReplayDistribution$, where darker line segments indicate scenarios with higher replay likelihood, \textbf{(middle)} a version of replay distribution under categorization with respect to the number of controlled agents in scenarios, and \textbf{(bottom)} we exemplify three scenarios that appear frequently.}
    \label{fig:experiment_step1_replay_pvl}
\end{figure}

Figures \ref{fig:experiment_step1_replay_actmae}, \ref{fig:experiment_step1_replay_amgae}, \ref{fig:experiment_step1_replay_gcade}, \ref{fig:experiment_step1_replay_learn}, \ref{fig:experiment_step1_replay_learngoal}, and
\ref{fig:experiment_step1_replay_pvl} illustrate the $\PLRReplayDistribution$ progression of PLR in case 1. Here we omit $\MaxMC$, as we provide its illustration in the main document. The utility functions with a high score temperature, i.e., $\PLRScoreTemperature=4$, as opposed to $\PLRScoreTemperature=2$, lead to a more uniform replay distribution (see Figures \ref{fig:experiment_step1_replay_amgae}, \ref{fig:experiment_step1_replay_gcade}, \ref{fig:experiment_step1_replay_learngoal}, \ref{fig:experiment_step1_replay_pvl} for $\AMGAE$, $\GCADE$, $\Learnability$, and $\PVL$, respectively). As the score temperature decreases, the impact of the ranking on the replay distribution also decreases. Furthermore, we observe that certain utility functions result in significant changes in the replay distribution throughout training, specifically when visualized with respect to the number of controlled agents in scenarios (see Figures \ref{fig:experiment_step1_replay_actmae} and \ref{fig:experiment_step1_replay_learn} for $\ACTMAE$ and $\LearnabilityHard$, respectively). This change may be due to a lower score temperature, which allows the ranking to impact the replay distribution more strongly. 

\subsection{Sensitivity to sampling interval and buffer size}
\label{app:sensitivity}
We ablate the two PLR hyperparameters that govern how often scores are refreshed and how many scenarios the buffer retains. Our reported configuration uses a buffer equal to the training set and $\ScenarioSamplingInterval=2\times10^6$ interactions, which is the scenario sampling interval GPUDrive uses under domain randomization. We compare against three variations, all with $\MaxMC$ in case 1 across three seeds:
\begin{enumerate*}[label=\textbf{(\arabic*)}]
    \item Small Buffer: a buffer of 100 scenarios, i.e., $|\PLRBuffer=100$|, and $\ScenarioSamplingInterval=2\times10^6$,
    \item Faster Sampling: full buffer, i.e., $|\PLRBuffer|=1,000$, and $\ScenarioSamplingInterval=1\times10^6$, and
    \item Slower Sampling: full buffer, i.e., $|\PLRBuffer|=1,000$, and $\ScenarioSamplingInterval=4\times10^6$.
\end{enumerate*}
\begin{figure}[t]
    \centering
    \includegraphics[width=.8\textwidth,trim={0 0pt 0pt 0pt},clip]{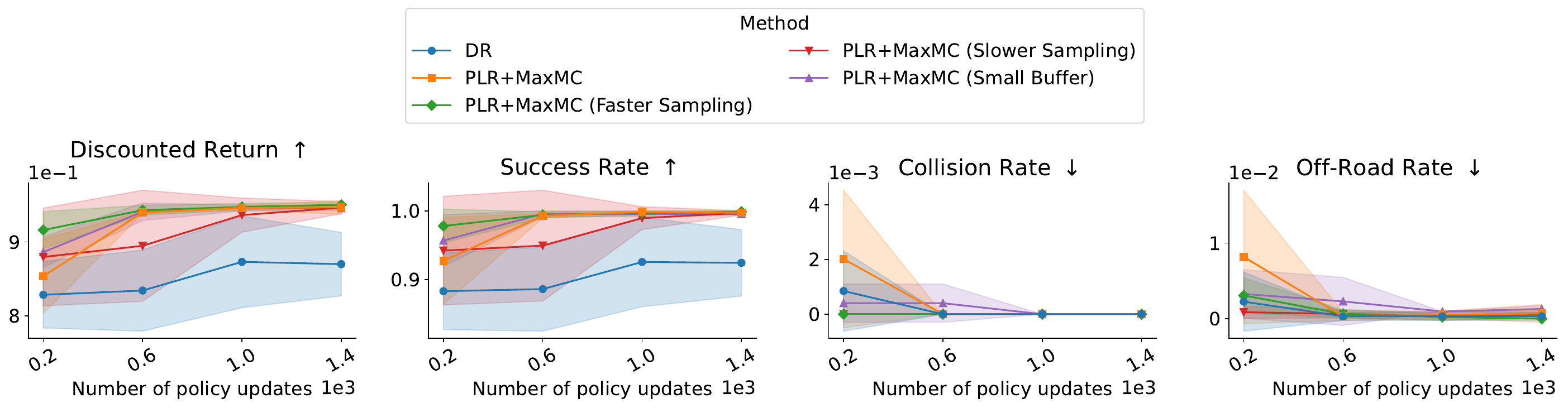}
    \caption{Sensitivity to sampling interval and buffer size: Performance progression of $\MaxMC$ in case 1 under three variations of our reported configuration, i.e., a buffer of 100 scenarios and scenario sampling intervals of $1\times10^6$ and $4\times10^6$ interactions. We evaluate on 150 test scenarios. Bold markers indicate the mean, whereas the shaded area covers one standard deviation around it across three independent training runs.}
    \label{fig:experiment_step1_plr_ablation}
\end{figure}
\cref{fig:experiment_step1_plr_ablation} reports the progression on the test split. All four configurations reach the same discounted return and success rate by 1,400 policy updates, with collision and off-road rates near zero from 1,000 updates onward, and all four exceed DR in return and success rate at every checkpoint, so CL4AD is insensitive to both hyperparameters over the ranges we test. The configurations differ only over the first few hundred updates, where faster sampling and the small buffer reach high success earliest and slower sampling trails the other three. This follows from the staleness mechanism in \cref{sec:multi_agent}: shorter sampling intervals keep buffer scores closer to on-policy, and a small buffer has a similar effect, as evicting all but the top-ranked 10\% of scenarios causes the retained ones to be revisited more often and prioritized more sharply. The small buffer carries elevated collision and off-road rates through 600 updates, since the curriculum spends less of its budget on the broader set of scenarios where the remaining safety failures occur.

\subsection{Combining utility functions across categories}
\label{app:combination}
\cref{sec:correlation} shows that utility functions correlate within a category, except for realism, but not across categories, suggesting that cross-category pairs carry complementary information. We combine two utility functions by ranking the scenarios independently under each and sampling from the averaged rank, preserving the scale invariance of $\PLRScoreDistribution$ in \cref{eq:p_utility}, since a weighted sum of raw scores does not consist of utilities expressed in different units. We evaluate three pairs in case 1 across three seeds, one from each pairwise combination of categories:
\begin{enumerate*}[label=\textbf{(\arabic*)}]
    \item $\MaxMC$ + $\LearnabilityHard$,
    \item $\MaxMC$ + $\ACTMAE$, and
    \item $\LearnabilityHard$ + $\ACTMAE$.
\end{enumerate*}
\begin{figure}[t]
    \centering
    \includegraphics[width=.8\textwidth,trim={0 0pt 0pt 0pt},clip]{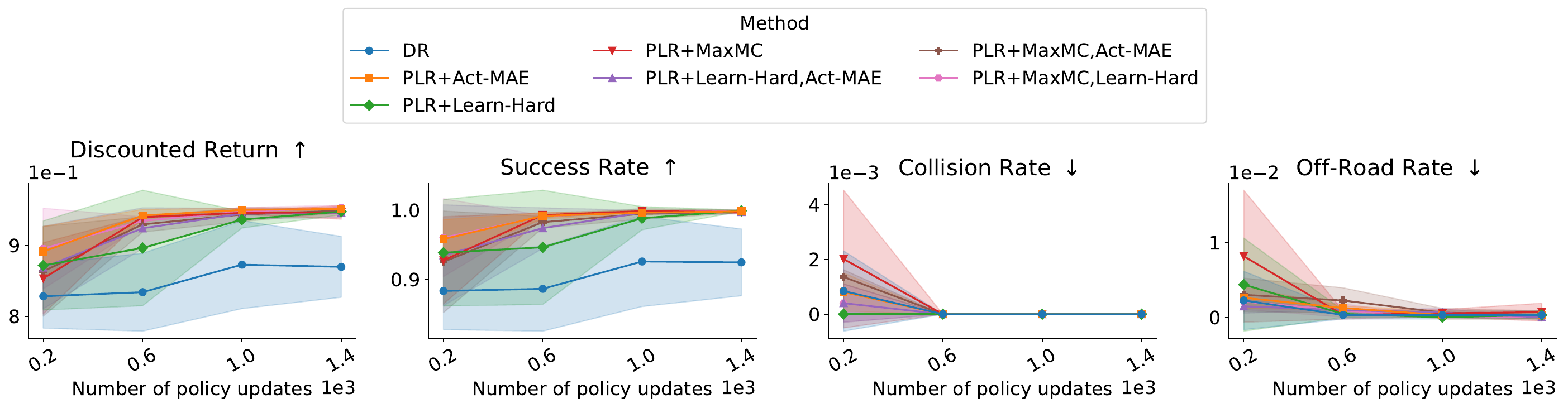}
    \caption{Combining utility functions across categories: Performance progression of three cross-category combinations in case 1, reported alongside their individual components and DR. We evaluate in 150 test scenarios. Bold markers indicate the mean, whereas the shaded area covers one standard deviation around it across three independent training runs.}
    \label{fig:experiment_step1_plr_combination}
\end{figure}
\cref{fig:experiment_step1_plr_combination} reports the progression on the test split. All three combinations converge with their individual components in discounted return and success rate by 1,400 policy updates, with collision and off-road rates near zero, and all exceed DR at every checkpoint. $\LearnabilityHard$ alone is the slowest PLR variant in this setting, whereas $\MaxMC$ + $\LearnabilityHard$ sits at or above both components at the early checkpoints, so combining a success-based function with a regret-based one does not average their behavior. One explanation, which we do not test here, is that the combined ranking draws from both density regimes, since Section 5.2 shows regret-based functions drifting above the dataset average traffic density while success-based functions shift below.
\end{document}